\documentclass{article}

\usepackage{microtype}
\usepackage{graphicx}
\usepackage{subcaption}
\usepackage{booktabs} 

\usepackage{hyperref}
\usepackage{xspace}

\PassOptionsToPackage{dvipsnames}{xcolor} 
\usepackage{xcolor}                        
\usepackage{amssymb}
\usepackage{booktabs}
\usepackage{subcaption}
\usepackage[export]{adjustbox}
\usepackage{bbm}
\usepackage{wrapfig}
\usepackage{multirow}

\usepackage{pgfplots}
\pgfplotsset{compat=1.14}
\usetikzlibrary{arrows,calc,intersections,fit,matrix,shapes.geometric}
\usepgfplotslibrary{fillbetween,colorbrewer,ternary,groupplots} 

\usepackage{tikzviolinplots}
\usepackage[noenumitem]{eda}
\usepackage{prettyplots}

\definecolor{forestgreen}{rgb}{0.13, 0.55, 0.13}

\newcommand{\attmap}{\mathcal{A}}
\DeclareMathOperator*{\argmax}{arg\,max}
\DeclareMathOperator*{\argmin}{arg\,min}
\newcommand{\varr}[1]{\textcolor{forestgreen}{$+$#1$\%$}}
\newcommand{\negvarr}[1]{\textcolor{forestgreen}{$-$#1$\%$}}

\pgfplotscreateplotcyclelist{prcl-iccv-line}{
	{dollarbill, mark=*, dotted},
	{darkblue, mark=*, dotted},
	{mediumpurple, mark=*, dotted},
	{mikadoyellow, mark=*, dotted},
    {orange, mark=*, dotted},
	{dollarbill, mark=*, solid},
	{darkblue, mark=*, solid},
	{mediumpurple, mark=*, solid},
	{mikadoyellow, mark=*, solid},
	{orange, mark=*, solid},
}

\newcommand{\ourmethod}{\textsc{AL}\xspace}
\newcommand{\gradcam}{GradCAM\xspace}
\newcommand{\gbp}{GBP\xspace}

\newcommand{\gbpfull}{Guided Backpropagation\xspace}

\newcommand{\ixg}{Input$\times$Gradient\xspace}

\newcommand{\ixgfull}{Input$\times$Gradient\xspace}

\newcommand{\igfull}{Integrated Gradient\xspace}

\makeatletter
\newcommand\notsotiny{\@setfontsize\notsotiny\@vipt\@viipt}
\makeatother

\usepackage[preprint]{icml2026}

\usepackage{fontawesome5}

\usepackage{amsmath}
\usepackage{amssymb}
\usepackage{mathtools}
\usepackage{amsthm}

\usepackage[capitalize,noabbrev]{cleveref}

\newcommand{\dist}{\attmap^{\text{soft}}}

\usetikzlibrary{positioning}

\usepackage{comment}
\usepackage{pifont}

\newsavebox{\teaserbox}

\begin{document}

\sbox{\teaserbox}{%
\begin{minipage}{.9\textwidth}
  \centering
\begin{tikzpicture}[
  img/.style={inner sep=0pt, outer sep=0pt},
  title/.style={inner sep=1pt, outer sep=0pt, font=\small},
  cls/.style={inner sep=1pt, outer sep=0pt, font=\small},
  rowlab/.style={inner sep=0pt, outer sep=0pt, font=\small, rotate=90}
]

\def\imgWidthOrig{.125\textwidth}
\def\imgWidthMini{.10\textwidth}
\def\miniGap{.01\textwidth} 

\pgfmathsetlengthmacro{\dx}{0.5*\imgWidthMini + 0.5*\miniGap }      
\pgfmathsetlengthmacro{\panelW}{2*\imgWidthMini + \miniGap }        
\pgfmathsetlengthmacro{\panelSep}{(\textwidth - 4*\panelW)/3}      

\pgfmathsetlengthmacro{\xA}{0.5*\panelW}
\pgfmathsetlengthmacro{\xB}{\xA + \panelW + \panelSep}
\pgfmathsetlengthmacro{\xC}{\xB + \panelW + \panelSep}
\pgfmathsetlengthmacro{\xD}{\xC + \panelW + \panelSep}

\pgfmathsetlengthmacro{\yOrig}{1.2cm}
\pgfmathsetlengthmacro{\yTitle}{2.45cm}     
\pgfmathsetlengthmacro{\yRowOne}{-1cm}   
\pgfmathsetlengthmacro{\yRowTwo}{-2.9cm}   
\pgfmathsetlengthmacro{\yClass}{-4cm}    

\newcommand{\Panel}[9]{%
  \begin{scope}[shift={(#1,0)}]
    \node[title] at (0,\yTitle) {#2};

    \node[img] at (0,\yOrig) {\includegraphics[width=\imgWidthOrig]{#3}};

    \node[img] at (-\dx,\yRowOne) {\includegraphics[width=\imgWidthMini]{#4}};
    \node[img] at ( \dx,\yRowOne) {\includegraphics[width=\imgWidthMini]{#5}};

    \node[img] at (-\dx,\yRowTwo) {\includegraphics[width=\imgWidthMini]{#6}};
    \node[img] at ( \dx,\yRowTwo) {\includegraphics[width=\imgWidthMini]{#7}};

    \node[cls] at (-\dx,\yClass) {#8};
    \node[cls] at ( \dx,\yClass) {#9};
  \end{scope}%
}

\Panel{\xA}{Object-Specific}
  {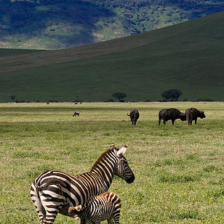}
  {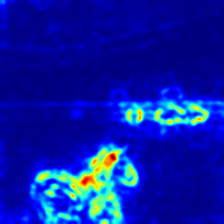}{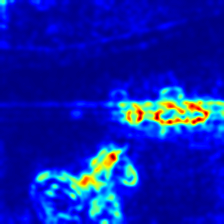}
  {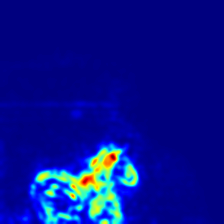}{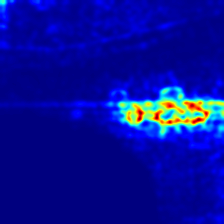}
  {Zebra}{Bison}

\Panel{\xB}{Concept-Specific}
  {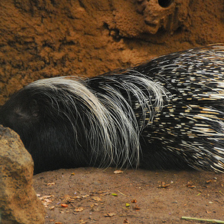}
  {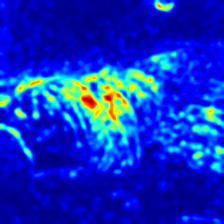}{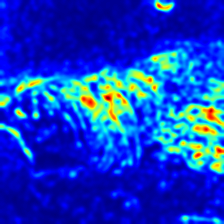}
  {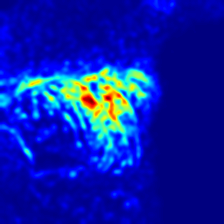}{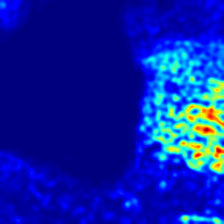}
  {Colobus}{Echidna}

\Panel{\xC}{Fine-grained}
  {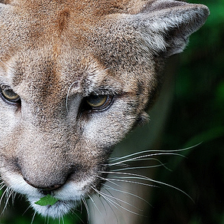}
  {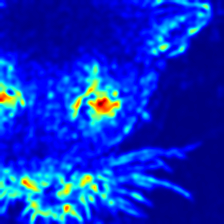}{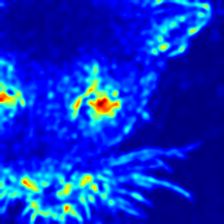}
  {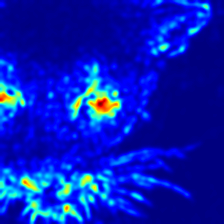}{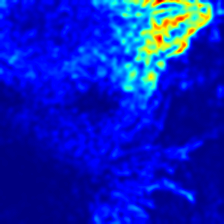}
  {Cougar}{Lynx}

\Panel{\xD}{Shared}
  {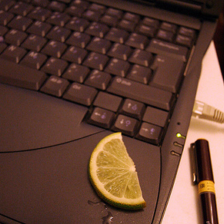}
  {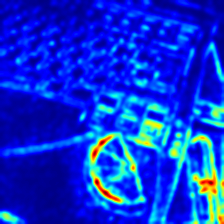}{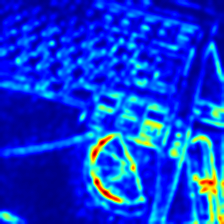}
  {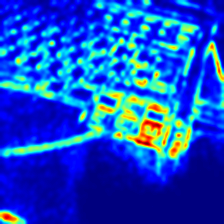}{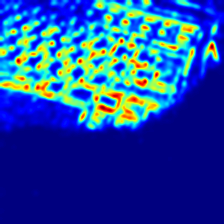}
  {Notebook}{Keyboard}

\pgfmathsetlengthmacro{\rowLabX}{\xA - 0.5*\panelW - 0.3cm}
\node[rowlab] at (\rowLabX,\yRowOne) {GBP};
\node[rowlab] at (\rowLabX,\yRowTwo) {GBP + \textbf{Ours}};

\end{tikzpicture}
\label{x}
\end{minipage}
}

\twocolumn[
  \icmltitle{Hidden in Plain Sight -- Class Competition Focuses Attribution Maps}



  \icmlsetsymbol{equal}{*}

  \begin{icmlauthorlist}
    \icmlauthor{Nils Philipp Walter}{yyy}
    \icmlauthor{Jilles Vreeken}{yyy}
    \icmlauthor{Jonas Fischer}{comp}
  \end{icmlauthorlist}

  \icmlaffiliation{yyy}{CISPA Helmholtz Center for Information Security, Saarbrücken, Deutschland}
  \icmlaffiliation{comp}{Max Planck Institute for Informatics, Saarbrücken, Deutschland}

  \icmlcorrespondingauthor{Nils Philipp Walter}{nils.walter@cispa.de}

  \icmlkeywords{Machine Learning, Interpretability, Attribution, ICML}
\vskip 0.25in
\begin{center}
  \usebox{\teaserbox} \vspace{-0.3cm}
    \captionof{figure}{\textit{Attributions on ImageNet.} Attributions computed as distributions across classes are: \textbf{object-specific} -- visually ground correct target objects, \textbf{concept-specific}, identifying features that are relevant on a by-part-basis, and  \textbf{fine-grained}, yielding features that distinguish closely related classes, while at the same time not obscuring features that are \textbf{shared} between closely related classes. In contrast, the standard approach of computing attributions on the logit of the predicted class does not reveal any of these properties.}
  \label{fig:crownjewel}
\end{center}
\vskip 0.2in
]

\printAffiliationsAndNotice{}  


\begin{abstract}

Attribution methods reveal which input features a neural network uses for a prediction, adding transparency to their decisions. A common problem is that these attributions seem unspecific, highlighting both important and irrelevant features. We revisit the common attribution pipeline and observe that using logits as attribution target is a main cause of this phenomenon. We show that the solution is in plain sight: considering distributions of attributions over multiple classes using existing attribution methods yields specific and fine-grained attributions. On common benchmarks, including the grid-pointing game and randomization-based sanity checks, this improves the ability of 18 attribution methods across 7 architectures up to $2\times$, agnostic to model architecture.

\end{abstract}


\section{Introduction}
\label{sec:intro}
Neural networks are widely used for decision-making but remain opaque. This is especially problematic in high-stakes settings such as medical imaging~\cite{borys:2023:saliencymed}, but also in a more general context, motivating the growing need for transparent explanations. A common approach  to explain and understand the prediction is to highlight which features in the input, such as regions in the input, drive a prediction; such methods are termed attribution methods as they attribute an importance score to input features. 

Such explanations, however, have shortcomings. While some of the attributed features appear sensible, the whole attribution seems overcomplete, making it difficult to determine which features are relevant for \textit{discriminating} between classes~\cite{rao2022towards}. 
As illustrated in the top row of Figure~\ref{fig:crownjewel}, standard methods (e.g., \textsc{GuidedBackprop}) show little difference when attributing features to different classes. They tend to highlight all salient objects (e.g., both Zebra and Bison) regardless of the target. Hence, these methods currently lack the ability to differentiate between features that are distinct and important for a class and those that are just loosely associated with it.

We argue that this lack of focus is not as much a problem of the attribution methods, but rather stems from \textit{how} we consider their output. That is, attribution is typically computed on the logit of a target class. However, looking at a logit in isolation discards the model-inherent discriminative mechanism of the subsequent softmax layer, which, for the final prediction, weighs the evidence for one class against others. Attributing to the softmax output is usually ineffective because the gradient vanishes for confident predictions, and the resulting maps become uninformative. 

\begin{figure*}
    \centering 
    \begin{tikzpicture}

      \node[anchor=south west, inner sep=0] (img) at (0,0) {
        \includegraphics[width=0.9\linewidth, trim=0.5cm 13cm 12cm 0.7cm, clip]{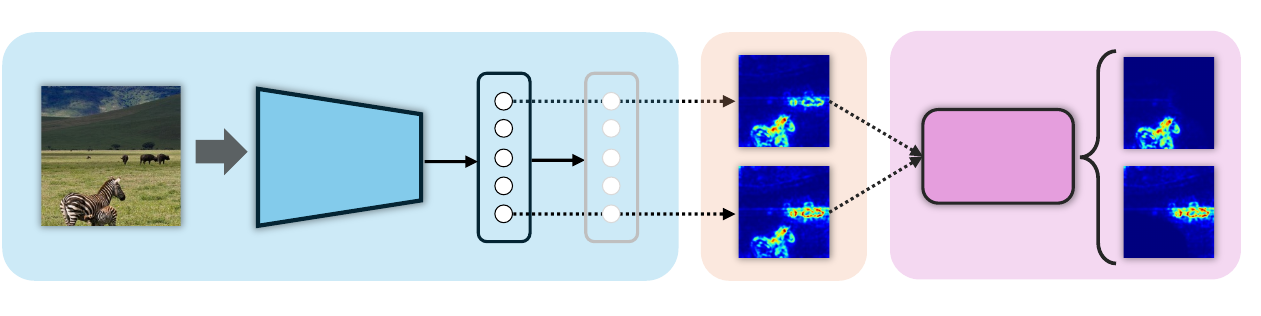}%
      };

      \def\topm{0.975}
      \begin{scope}[x={(img.south east)}, y={(img.north west)}]
       \node[] (text) at (0.27, \topm) {\textbf{Image Encoder + logits $=f$ }};
       \node[] (text) at (0.619, \topm) {\textbf{Attribution $\mathcal{A}$}};
       \node[] (text) at (0.84, \topm) {\textbf{Attribution + \ourmethod }};

       \node[] (text) at (0.2685, 0.52)  {\shortstack{Image\\Encoder}};

        \node[] (text) at (0.085, 0.78)  {\footnotesize Input Image };
       \node[] (text) at (0.4, 0.82)  {\scriptsize Logits};
       \node[] (text) at (0.485, 0.825)  {\scriptsize Softmax};

       \node[] (text) at (0.545, 0.725)  {\textbf{\emph{\scriptsize Zebra}}};
       \node[] (text) at (0.545, 0.374)  {\textbf{\emph{\scriptsize Bison}}};

       \node[] (text) at (0.79, 0.505)  {\scriptsize \shortstack{Pixel-wise\\
 attribution \\ distribution}
};

    \end{scope}
    \end{tikzpicture}
\caption{\textit{Reconsidering how to apply attributions}. The standard approach for softmax classifiers computes attribution maps with respect to a single class logit, effectively bypassing the softmax (because of vanishing gradients). This can yield diffuse and partially redundant attributions across classes (middle). We instead compute per-class attributions and convert them into pixel-wise attribution distributions by applying a softmax over classes at each pixel (right), producing explanations that better reflect the decision process of the model.}
 \vspace*{-0.2cm}
 \label{fig:overview}
\end{figure*}

Here, we propose a refinement that \emph{reintroduces} the competitive nature of the softmax without suffering from saturation to \textit{any} existing attribution method. Instead of considering the logits in isolation, we compute the \textit{distributions of attributions over multiple classes} (see Figure~\ref{fig:overview}). This can be seen as a lens that focuses attributions by turning single-class explanations into multi-class attribution distributions, hence we call the refinement \textbf{A}ttribution \textbf{L}ens (\textbf{\ourmethod}). 
By analyzing how the attribution for the target class relates to conflicting classes, we unlock properties that standard logit attribution miss. Specifically, as shown in Figure~\ref{fig:crownjewel}, the resulting attributions are: (i) \textbf{object-specific}, visually grounding the correct target object (e.g., separating Zebra from Bison) (ii) \textbf{concept-specific}, identifying features relevant on a by-part basis (e.g. fur vs. spikes); (iii) \textbf{fine-grained}, yielding features that distinguish closely related classes (e.g., the ears of a Lynx vs. Cougar); and (iv) \textbf{shared}, properly identifying features common between classes (e.g., between a notebook and a keyboard) without obscuring them.

These properties are also reflected in results on established attribution benchmarks, including the grid-pointing game and a part-annotated multi-object dataset \cite{rao2022towards,coco} (Table~\ref{table:exp:localization:resnet}, App. Table~\ref{table:exp-localization-all},\ref{table:exp-localization-all2}), 
insertion tests (Table~\ref{tab:insertion}, App. Table~\ref{table:exp-insertion-supp}), and randomization-based sanity checks \cite{adebayo2018sanity} (Figure \ref{fig:sanity-metrics-pearson-vit-resnet}, App. 
Figure~\ref{fig:sanity-metrics-full-resnet}-\ref{fig:sanity-metrics-full-vit32}).
On these benchmarks we show that \textbf{A}ttribution \textbf{L}ens refines existing attribution methods while remaining agnostic to model architecture, improving benchmark metrics across 18 attribution methods and 7 architectures by up to 
$\sim\!2\times$.

\section{Related Work}

In post-hoc explainability, there exist three main approaches for discovering prediction-relevant input features. Perturbation techniques probe model behavior by systematically modifying inputs, for instance by masking or deleting regions and measuring the resulting change in the model output~\cite{petsiuk2018rise, fong2019understanding, lundberg2017unified}, and are hence computationally expensive. Approximation techniques~\cite{ribeiro2016should,parekh2021framework} create interpretable surrogate models to mimic complex networks, but without guarantees that the surrogate reflects how the original model arrives at its decision.

Activation- and gradient-based attribution methods aim to balance efficiency and fidelity by considering the computation graph of the network. Well-known examples include \ixg~\cite{simonyan2013deep}, Integrated Gradients~\cite{sundararajan2017axiomatic,IG2}, and GBP~\cite{springenberg2014striving}, 
which are all based on gradients through the network, and 
\gradcam~\cite{selvaraju2017grad} and LayerCAM~\cite{layercam} which upsample feature maps while taking class information into account.

Layer-wise Relevance Propagation \citep[LRP, ][]{bach2015pixel}  considers the flow of activation values across the network under a conservation property, which requires architecture-specific adaptations~\cite{otsuki2024layer, chefer2021cheferlrp}. Similarly, DeepLift \cite{shrikumar2017learning} uses reference activations to determine neuron importance through custom backpropagation procedures.
For transformers, recent approaches propose modifications of attention roll-out, which reflects the propagation of information through the layers by multiplying each of their transition matrices, including Bi-attn~\cite{chen2023biattn}, T-attn~\cite{yuan2021tattn}, and InFlow~\cite{walker2025inflow}.

Because of their widespread use, benchmarking attribution methods in computer vision has been of growing interest.
\citet{ancona2018towards} study attribution sensitivity and formally proved equivalence between approaches under specific assumptions, whereas \citet{rao2022towards} systematically studied how faithful attributions are to an underlying prediction using the grid-pointing game. Insertion ablations~\cite{kapishnikov2019xrai} instead study the effect of insertion and deletion of attributed pixels on downstream performance as a proxy for attribution quality. \citet{adebayo2018sanity} evaluate attribution faithfulness based on stability of explanations with randomization of network components, which was later critically revisited~\cite{binder2023shortcomings}. 
We will use each of these metrics to study the impact of our suggested attribution approach.
Orthogonally, different learning objectives have been suggested to generally improve post-hoc explanation quality such as attributions \cite{gairola2025how}, which we later relate to our findings.

\section{Rethinking Attribution Computation}

Post-hoc attribution methods have been shown to perform poorly in recovering the classification-relevant information from the network~\cite{rao2022towards, bohle2022bcos} and arguably fail network perturbation based sanity checks~\cite{adebayo2018sanity}.
Commonly, the attributions for a target class--usually the predicted class--are computed using its logit as a target,%
which, however, means that the attribution will ignore the information from the other logits (see Fig.~\ref{fig:overview}).

Recent proposals for concept visualization in diffusion models, such as ConceptAttention \cite{conceptattention}, owe their success to considering \textit{multiple} concepts at once. Then, for each spatial location in the image, concept attention scores are normalized \textit{across all concepts}, thus determining which concept the image location is most associated with.

Similarly, in classification, the final prediction is based on all logits, with softmax contrasting the logits between classes, so it is far-fetched to expect attributions to recover prediction-relevant features from a single logit alone.
We propose to reconsider this common paradigm and propose
to compute attributions for logits of multiple classes, appropriately normalize them, and compute \textit{distributions over attributions} at each spatial location, similar to how a classification head computes an output distribution over multiple logits. 
This approach is grounded in a formal argument, which we discuss next, showing that existing attribution methods \textit{can retrieve class-specific information} when appropriately normalized across classes.

\subsection{Notation}
\label{sec:background}
We consider an input $x \in \mathcal{I}$, where here $\mathcal{I} = \mathbb{R}^{H \times W \times d}$ is typically an image of height $H$, width $W$, and $d$ channels. 
We describe a classification model as a function $f : \mathcal{I} \rightarrow \mathbb{R}^C$, predicting $C$ classes $\mathcal{C}=\{1,...,C\}$. 
The final class label is usually obtained as argmax over $f(x)$.
We consider attributions as functions $\attmap : \mathcal{I} \times f \times \mathcal{C} \rightarrow \mathcal{I}'$ that for an input, a model, and optionally a target class provide an explanation of a similar shape as the input. For images, we typically aggregate attributions across the channel dimension. The attribution method assigns each input feature $x_i$ a score indicating its contribution in the prediction of $f$ for that specific class.
This broad definition covers all attribution methods discussed in the related work section, examples are Input$\times$Gradient (IxG) as $\attmap_{\text{IxG}}(x, f, c) = x \odot \frac{\partial f_c}{\partial x}$, or
\gradcam as $\attmap_{\text{\gradcam}}(x, f, c) = \text{ReLU}(\sum_k \alpha_c^k a^k)$, where $\alpha_c^k = \frac{1}{Z} \sum_i \sum_j \frac{\partial f_c}{\partial a_{ij}^k}$ are the importance weights computed by global average pooling of the gradients. Although we focus on images here, the following generalizes without modification to other domains such as language.

\subsection{Focusing Attribution Methods}

Using a single logit $f_c$ is non-contrastive by construction, since it does not take competing classes into account. A more principled alternative is to consider the softmax probability $p_c = \text{softmax}(f(x))_c$, which inherently contrasts between class logits. However, the gradient of $p_c$, which would provide the attribution signal for most attribution methods, has an important drawback. Let $z_c$ denote the logit $f_c(x)$. The gradient of the softmax probability $p_c$ is
\begin{equation}
\nabla_x p_c = p_c \left( \nabla_x z_c - \sum_{c'=1}^\mathcal{C} p_{c'} \nabla_x z_{c'} \right)\,.
\label{eq:true-softmax-gradient}
\end{equation}
Although the subtractive term $\sum p_{c'} \nabla_x z_{c'}$ appears to provide the necessary contrast, it is negligible in practice. In the high-confidence regime where $p_c \to 1$, the weighted sum of gradients converges to the gradient of the class with the highest logit, $\nabla_x z_c$. This causes Eq.~\ref{eq:true-softmax-gradient} to approach zero. The resulting gradient thus fails to attribute importance to the very features that drive a confident prediction.

To build a robust contrastive attribution, we must preserve the model-inherent competition between logits while avoiding this self-canceling behavior. Our core idea is to move the contrastive mechanism from the output layer of the model, which operates on saturated probabilities, directly into the attribution maps themselves, computed for the logits.
We accomplish this by a softmax over \textit{attributions of} logits at each input location (pixel) instead of output logits.

Assume a gradient-based attribution $\mathcal{A}_c = \nabla_x z_c$ for class $c$. For a chosen set of classes $\mathcal{C}' \subseteq \mathcal{C}$ of size $C'=|\mathcal{C}'|$, we compute the base attribution map $\mathcal{A}_c$ for each class $c \in \mathcal{C}'$. To create contrast, we then apply a softmax to each input feature attribution, here at each pixel $(i,j)$in an image,
\begin{equation}
\dist_c[i,j] = \frac{\exp(\mathcal{A}_c[i,j] / t)}{\sum_{c' \in \mathcal{C}'} \exp(\mathcal{A}_{c'}[i,j] / t)}
\label{eq:local-softmaxq}
\end{equation}
where $t$ is a temperature to amplify the contrast.
This yields a \textit{distribution of attribution over classes} at each input feature $\sum_{c \in \mathcal{C}'} \dist_c[i,j] = 1$. These local class probabilities express how dominant each class is in each spatial location.
One might now attempt to directly mimic Eq.~\eqref{eq:true-softmax-gradient} by replacing the global softmax weights $p_c$ with the local $\dist_c(i,j)$, resulting in an attribution of the form:
\begin{equation}
\mathcal{A}_c[i,j] - \sum_{c'} \dist_{c'}[i,j] \mathcal{A}_{c'}[i,j].
\end{equation}
However, this naïve substitution reintroduces the vanishing behavior. The sum includes the self-term $\dist_{c'}[i,j] \attmap_{c'}[i,j]$, so when class $c'$ dominates a pixel, i.e., $\dist_{c'}[i,j] \approx 1$, the full expression again tends to zero. Summing only over $c'\neq c$ is also problematic, as $\dist_{c'}$ and $\mathcal{A}_{c'}$ are strongly correlated, which may lead to overshooting.

Instead, we reduce attributions in proportion to how strongly other classes (not the target class) are influenced by a pixel through $f$.  Rather than altering gradients for all classes, we discount the target attribution by the fraction of its evidence that is shared with non-target classes, which gives
\begin{align}
        & \mathcal{A}_c[i,j] - \sum_{c' \ne c} \dist_{c'}[i,j]\, \mathcal{A}_c[i,j] \\
        &= \mathcal{A}_c[i,j] \left(1 - \sum_{c' \ne c} \dist_{c'}[i,j]\right) = \mathcal{A}_c[i,j] \cdot \dist_c[i,j].
    \label{eq:contrastive-form}
\end{align}

Thus, each pixel’s attribution is the original attribution scaled, either up or down, by its class probability $\dist_c[i,j]$.

\subsection{From Gradients to General Attributions}
\label{subsec:method}
Most attribution methods can be understood as functions of the gradient. They either use it directly (Input$\times$Gradient), pool it spatially (Grad-CAM), or transform it through propagation rules (Guided Backpropagation, DeepLIFT, etc.). One way to extend our derivation would be to modify each method individually and insert the contrastive reweighting at the gradient level. However, such an approach would be cumbersome and method-specific.

We instead propose a plug-and-play refinement that operates directly on saliency maps, which we call \textbf{A}ttribution \textbf{L}ens ( \textbf{\ourmethod}, for short) as it functions as a lens that focuses on the evidence most important for the model (see Figure \ref{fig:overview}). For a subset $\mathcal{C'}$ of classes, we compute the class-wise attributions $\attmap(x,f,c)$ and normalize them at each pixel using the spatial softmax of Eq.~\eqref{eq:local-softmaxq}. We denote the resulting distribution by
\begin{equation}
\dist_c[i,j] :=\frac{\exp(\attmap(x,f,c)[i,j] / t)}{\sum_{c' \in \mathcal{C'}} \exp(\attmap(x,f,c')[i,j] / t)} \; ,
\end{equation}
which expresses the relative dominance of class $c$ at location $(i,j)$. To increase robustness, we average $\dist_c[i,j]$ over multiple temperatures $t$, producing smoother distributions that capture contrast at different granularities.
We then define \ourmethod to refine  $\mathcal{A}_c$ of the target class $c$ as
\begin{equation}\label{eq:attribution}
\ourmethod(x,f,c) = \attmap(x,f,c) \odot \dist_{c} \odot \mathbbm{1}_{\dist_{c} > \tfrac{1}{C'} }\; ,
\end{equation}
where $\odot$ denotes element-wise multiplication. We mask out pixels where the target class is no more than chance among the comparison classes \((\dist \leq 1/C')\), retaining only locations where \(c\) is above-chance \emph{within that set}. In practice, we average over \(1/t \in \{1,5,100\}\) to stabilize the class competition. Extensive ablations on $t$ in Appendix~\ref{appsec:scalingpars} show that results are comparable across varying $t$.

\begin{figure*}[t]
\centering
\begin{tikzpicture}
  \node[anchor=south west, inner sep=0] (img) at (0,0) {%
    \includegraphics[width=0.93\linewidth,trim={1.5cm 3cm 2cm 4.5cm}, clip]{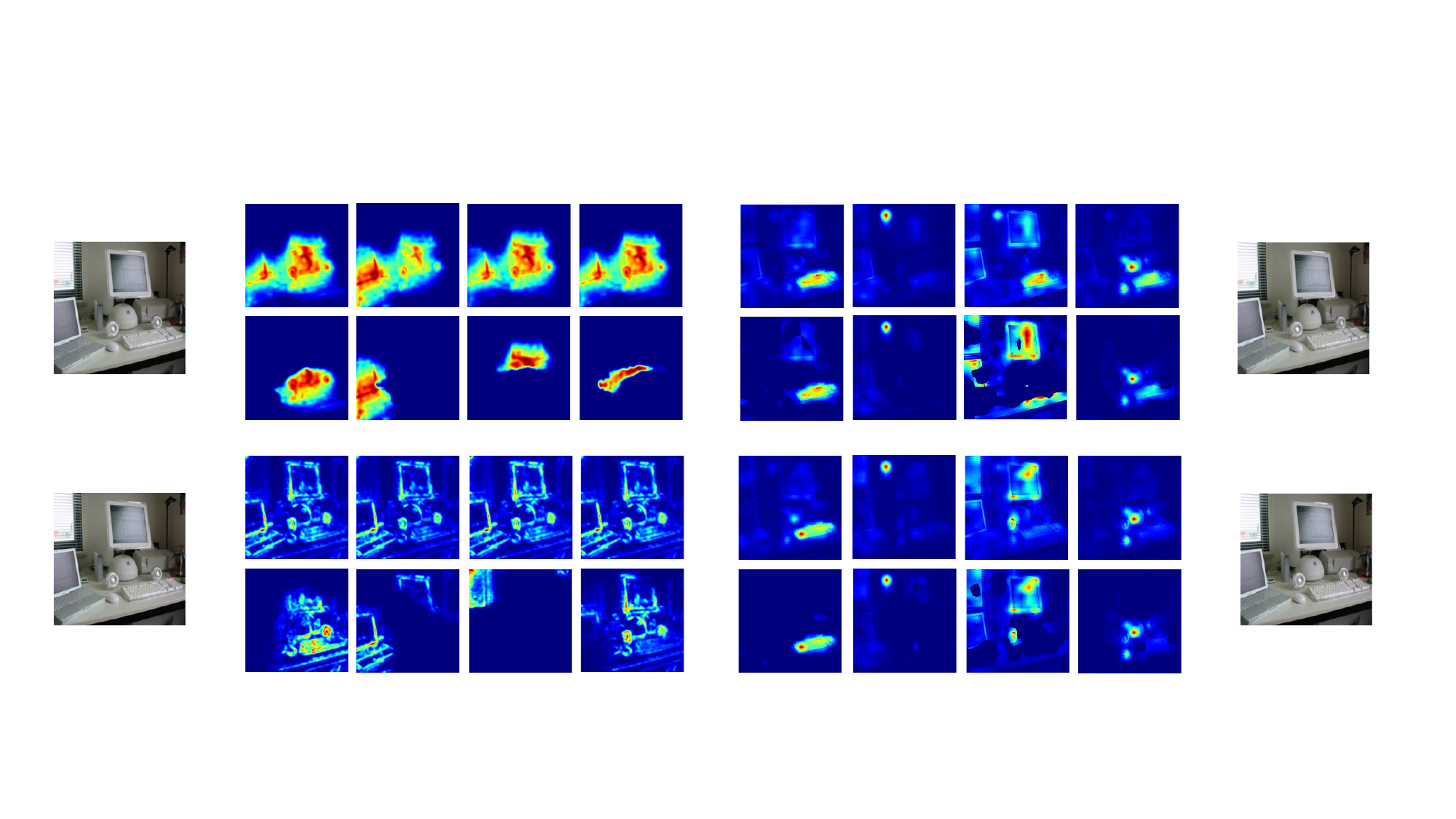}%
  };

  \begin{scope}[x={(img.south east)}, y={(img.north west)}]

    \node[font=\normalsize] at (0.31, 1.1) {Resnet 50};
    \node[font=\normalsize] at (0.69, 1.1) {ViT-base-16};

    \node[font=\scriptsize] at (0.175, 1.018) {\emph{Keyboard}};
    \node[font=\scriptsize] at (0.263, 1.018) {\emph{Laptop}};
    \node[font=\scriptsize] at (0.348, 1.018) {\emph{Monitor}};
    \node[font=\scriptsize] at (0.43, 1.018) {\emph{Mouse}};
    \node[font=\scriptsize] at (0.557, 1.018) {\emph{Keyboard}};
    \node[font=\scriptsize] at (0.644, 1.018) {\emph{Laptop}};
    \node[font=\scriptsize] at (0.73, 1.018) {\emph{Monitor}};
    \node[font=\scriptsize] at (0.812, 1.018) {\emph{Mouse}};

    \node[font=\small, rotate=90]  at (-0.025, 0.78) {LRP};
    \node[font=\small, rotate=90]  at (-0.025, 0.26) {Guided BackProp};
    \node[font=\small, rotate=-90] at ( 1.025, 0.78) {T\text{-}LRP};
    \node[font=\small, rotate=-90] at ( 1.025, 0.26) {InFlow};

    \node[font=\scriptsize, rotate=90] at ( 0.121, 0.865) {Vanilla};
    \node[font=\scriptsize, rotate=90] at ( 0.121, 0.65) {\textbf{+ \ourmethod}};

    \node[font=\scriptsize, rotate=90] at ( 0.121, 0.36) {Vanilla};
    \node[font=\scriptsize, rotate=90] at ( 0.121, 0.145) {\textbf{+ \ourmethod}};

    \node[font=\scriptsize, rotate=-90] at ( 0.87, 0.865) {Vanilla};
    \node[font=\scriptsize, rotate=-90] at ( 0.87, 0.65) {\textbf{+ \ourmethod}};

        \node[font=\scriptsize, rotate=-90] at ( 0.87, 0.36) {Vanilla};
    \node[font=\scriptsize, rotate=-90] at ( 0.87, 0.145) {\textbf{+ \ourmethod}};

  \end{scope}
\end{tikzpicture}

\caption{\emph{\ourmethod works across architectures and methods.} For each baseline (top), our refinement (bottom) sharpens class-specific regions (keyboard, laptop, monitor, mouse). In ResNet-50, the effect is strongest, revealing the class-specific information clearly. In ViT-base-16, attributions already cover relevant areas but remain diffuse; \ourmethod reduces this blur and highlights the important regions cleanly.}
\vspace*{-0.25cm}
\label{fig:attr-comparison-arch}
\end{figure*}

Since \ourmethod operates solely on the outputs of attribution methods, it is model-agnostic and can be applied to any model and attribution function that satisfy the signature defined above. In Figure~\ref{fig:attr-comparison-arch}, we demonstrate this plug-and-play behavior across different models and attribution pairings.

\subsection{Selecting the Set of Classes}
\label{sec:selectclasses}

One design choice of \ourmethod is the set of classes $\mathcal{C'}$ used to compute the refinement. We explore three approaches, each offering distinct advantages depending on the specific goals and application context.

\textbf{Predefined class sets.} When meaningful contrasts are dictated by the task or domain, one can fix $\mathcal{C'}$ a priori. Examples include quadrant classes in grid-pointing games or clinically relevant disease subtypes in medical imaging. This yields application-aligned attributions, but requires prior knowledge that may not be available.

\textbf{Top-$k$ most probable classes.} A model-driven option is to choose the $k$ highest-scoring classes for the input (and, optionally, also include the lowest-scoring class). Contrasting these classes require only mild computational overhead and emphasizes evidence that separates the most plausible alternatives, while also revealing features shared among closely related high-probability classes.

\textbf{Best--vs--worst classes.} The third approach compares the highest-probability  against lowest-probability class
$\mathcal{C'}=\{c_{\max},c_{\min}\}$ with $c_{\max}=\argmax_c f_c(x)$ and $c_{\min}=\argmin_c f_c(x)$.
Such extreme can reveal the most distinctive characteristics of the input as interpreted by the model, by showing which features are most critical for pushing the model toward or away from certain classifications and is very efficient to compute.

We use predefined sets for localization and deletion/insertion tests, and a top-$2$ strategy for randomization tests.

\subsection{Computational Complexity}
We need to compute \(C'\) attribution maps, each with cost \(A\); hence, the runtime is \(O(C'A)\). The refinement step applies softmax, weights by $\dist$, and thresholds, and is parallelizable making its cost negligible compared to the attribution maps. In practice, \(C'\) is small (e.g., only a few top-scoring classes), and for gradient-based methods much of the computation can be reused across classes by caching the backward path since only the final layer changes, further 

\section{Experiments}%

\begin{table*}
    \centering

\caption{\textit{Consistent improvement of attributions.} We measure the improvement of applying \ method across 13 different attribution methods, considering convolutional and transformer-based architectures, using Region Attribution (RA), Intersection over Union (IoU), and F1. We observe that \ourmethod consistently improves the base method. We provide results for more architectures in App. Tab.~\ref{table:exp:localization:resnet} showing similar trends. We show the value that the method achieves when augmented with \ourmethod and in percent the level of improvement. }
\resizebox{0.9\textwidth}{!}{\begin{tabular}{cl@{\hspace{2em}} r@{}lr@{}lr@{}l@{} c@{\hspace{2em}} r@{}lr@{}lr@{}l@{} c@{\hspace{2em}} r@{}lr@{}lr@{}l}
\toprule
& & \multicolumn{6}{c}{Quad-ImageNet} && \multicolumn{6}{c}{Part-Quad-ImageNet} && \multicolumn{6}{c}{COCO} \\
\cmidrule{3-8} 
\cmidrule{10-15}  
\cmidrule{17-22}
& Method & \multicolumn{2}{c}{RA} & \multicolumn{2}{c}{IoU} & \multicolumn{2}{c}{F1} && \multicolumn{2}{c}{RA} & \multicolumn{2}{c}{IoU} & \multicolumn{2}{c}{F1} && \multicolumn{2}{c}{RA} & \multicolumn{2}{c}{IoU} & \multicolumn{2}{c}{F1}\\
\midrule
\multirow{6}{*}{\rotatebox[origin=c]{90}{Resnet50}}& GradCam & 0.88&\varr{25} & 0.67&\varr{64} & 0.79&\varr{38} && 0.31&\varr{28} & 0.24&\varr{112} & 0.36&\varr{87} && 0.18&\varr{19} & 0.11&\varr{16} & 0.17&\varr{12} \\
& GBP & 0.86&\varr{144} & 0.26&\varr{32} & 0.41&\varr{25} && 0.44&\varr{146} & 0.08&\varr{43} & 0.14&\varr{38} && 0.19&\varr{30} & 0.09&\varr{3} & 0.15&\varr{2} \\
& Guide-GC & 0.91&\varr{21} & 0.34&\varr{31} & 0.50&\varr{23} && 0.50&\varr{24} & 0.12&\varr{49} & 0.21&\varr{42} && 0.23&\varr{16} & 0.10&\varr{8} & 0.16&\varr{8} \\
& IxG & 0.55&\varr{37} & 0.20&\textcolor{gray}{$+0\%$} & 0.33&\textcolor{gray}{$+0\%$} && 0.25&\varr{47} & 0.06&\textcolor{gray}{$+0\%$} & 0.11&\textcolor{gray}{$+0\%$} && 0.13&\varr{11} & 0.09&\textcolor{gray}{$+0\%$} & 0.15&\textcolor{gray}{$+0\%$} \\
& IG & 0.56&\varr{36} & 0.20&\textcolor{gray}{$+0\%$} & 0.34&\textcolor{gray}{$+0\%$} && 0.28&\varr{48} & 0.06&\textcolor{gray}{$+0\%$} & 0.12&\textcolor{gray}{$+0\%$} && 0.14&\varr{11} & 0.09&\textcolor{gray}{$+0\%$} & 0.15&\textcolor{gray}{$+0\%$} \\
& LRP & 0.88&\varr{56} & 0.69&\varr{97} & 0.79&\varr{55} && 0.37&\varr{49} & 0.22&\varr{117} & 0.34&\varr{90} && 0.21&\varr{20} & 0.13&\varr{8} & 0.20&\varr{7} \\
\cmidrule[0.1pt]{3-22}
& Avg. Improvement & \multicolumn{2}{r}{\varr{53.17}\hphantom{$1$}} & \multicolumn{2}{r}{\varr{37.33}\hphantom{$1$}} & \multicolumn{2}{r}{\varr{23.50}} && \multicolumn{2}{r}{\varr{57.00}\hphantom{$1$}} & \multicolumn{2}{r}{\varr{53.50}} & \multicolumn{2}{r}{\varr{42.83}} && \multicolumn{2}{r}{\varr{17.83}} & \multicolumn{2}{r}{\varr{5.83}\hphantom{$1$}} & \multicolumn{2}{r}{\varr{4.83}\hphantom{$1$}} \\
\midrule
\multirow{7}{*}{\rotatebox[origin=c]{90}{ViT-base-16}}& Bi-attn & 0.94&\varr{31} & 0.71&\varr{180} & 0.82&\varr{103} && 0.51&\varr{40} & 0.28&\varr{309} & 0.40&\varr{222} && 0.30&\varr{43} & 0.16&\varr{52} & 0.23&\varr{42} \\
& GradCam & 0.91&\varr{6} & 0.62&\varr{16} & 0.75&\varr{10} && 0.58&\varr{11} & 0.27&\varr{39} & 0.39&\varr{32} && 0.31&\varr{10} & 0.15&\varr{11} & 0.22&\varr{9} \\
& InFlow & 0.86&\varr{21} & 0.56&\varr{126} & 0.71&\varr{78} && 0.53&\varr{23} & 0.20&\varr{198} & 0.31&\varr{153} && 0.29&\varr{20} & 0.13&\varr{23} & 0.20&\varr{21} \\
& Grad-Rollout & 0.73&\varr{76} & 0.53&\varr{113} & 0.68&\varr{71} && 0.40&\varr{94} & 0.20&\varr{197} & 0.30&\varr{148} && 0.24&\varr{30} & 0.12&\varr{19} & 0.19&\varr{17} \\
& T-attn & 0.93&\varr{32} & 0.71&\varr{180} & 0.82&\varr{102} && 0.47&\varr{38} & 0.29&\varr{321} & 0.40&\varr{229} && 0.29&\varr{44} & 0.16&\varr{53} & 0.23&\varr{43} \\
& T-LRP & 0.77&\varr{35} & 0.51&\varr{105} & 0.66&\varr{65} && 0.47&\varr{36} & 0.20&\varr{201} & 0.31&\varr{152} && 0.27&\varr{17} & 0.12&\varr{20} & 0.19&\varr{18} \\
& Gradient & 0.93&\varr{4} & 0.57&\varr{3} & 0.70&\varr{2} && 0.50&\varr{8} & 0.34&\varr{11} & 0.47&\varr{9} && 0.30&\varr{10} & 0.17&\varr{2} & 0.25&\varr{2} \\
\cmidrule[0.1pt]{3-22}
& Avg. Improvement & \multicolumn{2}{r}{\varr{29.29}\hphantom{$11$}} & \multicolumn{2}{r}{\varr{103.29}\hphantom{$1$}} & \multicolumn{2}{r}{\varr{61.57}\hphantom{$1$}} && \multicolumn{2}{r}{\varr{35.71}\hphantom{$11$}} & \multicolumn{2}{r}{\varr{182.29}\hphantom{$1$}} & \multicolumn{2}{r}{\varr{135.00}} && \multicolumn{2}{r}{\varr{26.29}} & \multicolumn{2}{r}{\varr{28.57}\hphantom{$1$}} & \multicolumn{2}{r}{\varr{25.71}\hphantom{$1$}} \\

\bottomrule
\end{tabular}}
    \label{table:exp:localization:resnet}
\end{table*}
%
\begin{table*}
  \centering
  \caption{\textit{Improving attributions on insertion test.}  Augmenting the base method with \ourmethod improves the AUC for insertion tests for convolutional architectures by $8$-$11\%$ and modestly improves ransformer attribution methods by $\sim$$2\%$ across architectures.}%
\begin{subtable}{0.55\linewidth}
  \centering
    \subcaption{CNN-based architectures}%
  \resizebox{0.96\linewidth}{!}{\begin{tabular}{l r@{}r@{}l r@{}r@{}l r@{}r@{}l r@{}r@{}l}
    \toprule
    Method & \multicolumn{3}{c}{ResNet50}  & \multicolumn{3}{r}{WideResNet50-2} & \multicolumn{3}{r}{DenseNet121} & \multicolumn{3}{r}{ConvNeXT} \\
    \midrule
        IG       &\hphantom{i}& 0.53&\varr{13} &\hphantom{WideRe}& 0.60&\varr{11} &\hphantom{Den}& 0.50&\varr{9} &\hphantom{Co}& 0.30&\varr{11} \\
        GBP      && 0.65&\varr{25} && 0.72&\varr{24} && 0.64&\varr{23} && 0.26&\varr{13} \\
        IxG      && 0.50&\varr{14} && 0.58&\varr{14} && 0.45&\varr{10} && 0.25&\varr{14} \\
        Guide-GC && 0.65&\varr{3}  && 0.72&\varr{3} && 0.61&\varr{7} && 0.34&\varr{3} \\
        GradCam  && 0.61&\textcolor{gray}{${+0\%}$} && 0.67&\textcolor{gray}{${+0\%}$} && 0.51&\textcolor{darkred}{${-2\%}$} && 0.31&\textcolor{darkred}{${-6\%}$} \\
        LRP      && 0.61&\varr{11} && 0.69&\varr{5} && 0.44&\textcolor{gray}{${+0\%}$} && 0.00&\textcolor{gray}{$+0\%$} \\
        \cmidrule[0.1pt](lr){2-13}
        Avg. Improvement & \multicolumn{3}{r}{\varr{11.00}} & \multicolumn{3}{r}{\varr{9.50}\hphantom{$1$}} & \multicolumn{3}{r}{\varr{7.83}\hphantom{$1$}} & \multicolumn{3}{r}{\varr{5.83}\hphantom{$1$}} \\
    \bottomrule
  \end{tabular}}
\end{subtable}\hfill%
\begin{subtable}{0.4\linewidth}
  \centering
  \small
    \subcaption{Transformer-based architectures}%
  \resizebox{\linewidth}{!}{\begin{tabular}{lrrr}
    \toprule
    Method & ViT-base-8 & ViT-base-16 & ViT-base-32 \\
    \midrule
    Bi-attn & 0.53\varr{4} & 0.46\varr{2} & 0.52\varr{2} \\
    T-attn & 0.53\varr{4} & 0.46\varr{2} & 0.52\varr{2} \\
    InFlow & 0.51\varr{4} & 0.45\varr{2} & 0.51\textcolor{gray}{${+0\%}$} \\
    Gradient & 0.52\textcolor{gray}{${+0\%}$} & 0.44\textcolor{gray}{${+0\%}$} & 0.49\textcolor{darkred}{${-2\%}$} \\
    Grad-Rollout & 0.51\varr{9} & 0.43\varr{7} & 0.50\varr{6} \\
    TLRP & 0.00\textcolor{gray}{$+0\%$} & 0.00\textcolor{gray}{$+0\%$} & 0.00\textcolor{gray}{$+0\%$} \\
    \cmidrule[0.1pt](lr){2-4}
    Avg. Improvement & \varr{3.50} & \varr{2.17} & \varr{1.33} \\
    \bottomrule
  \end{tabular}}
\end{subtable}

  \label{tab:insertion}
\end{table*}           
\input{tex_figs/quads_main}%
We empirically evaluate our attribution refinement \ourmethod across 13 gradient-based and 5 perturbation-based attribution methods on three benchmark settings. We assess (i) localization performance, (ii) insertion \& deletion tests, and (iii) randomization-based sanity checks to verify robustness.

To assess localization ability, we consider the validation set of ImageNet~\cite{imagenet} and PartImageNet \cite{he2022partimagenet} to generate images for the Grid Pointing Game introduced by ~\citet{rao2022towards}, as well on the MS-COCO dataset ~\cite{coco}. We assess the quality of attributions by measuring how well these match annotated bounding boxes and segmentation masks. 

For insertion \& deletion tests, we quantitatively evaluate attributions using perturbation-based metrics \cite{petsiuk2018rise}, which assess whether highly attributed pixels are truly relevant to the model’s prediction. Specifically, we employ the insertion (deletion) protocol, which progressively adds (removes) the most-attributed pixels and measures the resulting change in model confidence.

To check whether attribution methods reflect what the model has learned, we run cascading randomization tests \citep{adebayo2018sanity}. We progressively randomize model parameters (from later to earlier layers) and verify that the resulting attributions degrade accordingly.

We consider different architectures, including ResNet-50~\cite{he2016deep}, Vision Transformer B/16 (ViT)~\cite{dosovitskiy2020image}, and provide further results for  DenseNet-121~\cite{huang2017densely}, WideResNet-50-2~\cite{zagoruyko2016wide}, ConvNeXt~\cite{liu2022convnet}, Vision Transformer B/8 and Vision Transformer B/16 in the Appendix. All models are pretrained on ImageNet and downloaded from PyTorch \cite{paszke2019pytorch}. 

For attribution methods, we adopt widely used approaches for CNNs Grad-CAM, Guided Backprop, IG, IxG, Guided Grad-CAM~\cite{selvaraju2017grad,springenberg2014striving,sundararajan2017axiomatic,shrikumar2017learning,kokhlikyan2020captum} and for ViTs Grad-CAM, InFlow, Grad-Rollout, Bi-Attn, T-attn, T-LRP, gradient saliency ~\cite{walker2025inflow,abnar2020quantifying,chen2023biattn,yuan2021tattn,chefer2021cheferlrp}. Because transformer saliency maps are blurry, we multiply them with the input (similar to IxG) for illustration.

As perturbation-based explainability methods, we consider Shapley Values and their kernel-based approximation (Kernel SHAP) \cite{lundberg2017unified}, LIME \cite{ribeiro2016should}, as well as simple perturbation schemes such as feature ablation and occlusion \cite{zeiler2014visualizing}. These approaches quantify feature importance by selectively modifying parts of the input and observing the resulting change in the model output. We report the results in Appendix \ref{app:perturbation}.

\input{tex_figs/ablations}
\input{tex_figs/sanity_checks}
\subsection{Localization ability}
\paragraph*{Metrics} We assess attribution quality by measuring how well attributions align with the
actual object regions. The Region Attribution (RA) metric quantifies what portion of the total attribution weight falls within the target region, providing insight into attribution focus.
The Intersection over Union (IoU) measures the spatial overlap between the
attribution map and the ground truth region, and the F1-score score is computed between attributed and true pixels of the target object. To prevent methods from being unduly
rewarded for producing diffuse attributions, we apply a Gaussian blur to the attribution maps and ensure a fair comparison across different approaches following~\citet{rao2022towards}. For both setups, we use the ground truths for $\mathcal{C}'$.

\paragraph*{Grid Pointing} \label{para:gridpointing}
For the grid-pointing game, we compile a $2 \times 2$ grid of random images from ImageNet validation set, which we call
Quad-ImageNet. We also generate such grids using the PartImagenet dataset \cite{he2022partimagenet}, which is a subset of ImageNet but annotated with segmentation masks.
Across attribution scores and architectures, we observe that \ourmethod never degrades performance, at minimum we see that for specific methods and benchmark setups it is on par with the standard pipeline. Most of the time, we see that the \textbf{existing attribution methods refined through \ourmethod show much better localization ability} (see Tab.~\ref{table:exp:localization:resnet}).

For ResNet50, we observe substantial gains in RA, IoU, and F1, with an average improvement of upto \varr{53}, \varr{37} and \varr{42} across different attribution scores. For specific methods, such as \gbpfull, the RA score even doubles on the Quad-Imagenet and QuadPart-ImageNet benchmark.
For the ViT model, IoU and F1 scores more than double. This strong improvement can be attributed to \ourmethod filtering out uniformly unimportant regions in the noisy attribution maps produced for ViT.

Qualitatively, we also observe these improvements, now capturing both the \textit{distinguishing} as well as \textit{common} features of closely related classes (cf. Fig. \ref{fig:attr-comparison}).
These results show that the \ourmethod pipeline enhances attribution methods in precisely localizing features most relevant to the classification. 

\paragraph*{MS-COCO}
For MS-COCO, we use the whole validation set. We filter objects that are smaller than 1\% of the image and objects for which the model's confidence is less than $10^{-4}$.
We observe similar trends, albeit more modest than on Quad-Imagenet, achieving an average improvement of \varr{17.83}, \varr{5.83}, and \varr{4.83} on RA, IoU, and F1, respectively. Again we observe that through \ourmethod, the attributions focus on more distinctive features rather than entire object regions. COCO's natural images contain multiple objects with complex backgrounds, making precise localization more challenging, yet with \ourmethod we do improve F1 scores across regardless of attribution scoring approach on both ResNet50 and ViT, indicating better overall localization despite the more challenging context. We provide results for different convolutional and transformer-based architectures in App. Tab.~\ref{table:exp:localization:resnet}, showing similar improvements.

\subsection{Insertion \& Deletion ablations}

We evaluate the actionability of refined saliency maps using standard \emph{insertion} and \emph{deletion} tests. 
For \emph{insertion}, we start from a blurred baseline and progressively reveal the top-$k$ pixels ranked by the attribution map, tracking the target-class probability; better attributions yield faster increases. 
For \emph{deletion}, we start from the original image and progressively remove the top-$k$ pixels, where better attributions yield faster decreases. We summarize both tests by the area under the corresponding probability curves (AUC).

To obtain a controlled multi-object setting without overly diluting probability mass, as compated to  four-image grids, we construct two-image composites. 
Using PartImageNet segmentation masks, we retain images where at least $80\%$ of object pixels lie within the central $50\%$ of the image width, crop this region, and concatenate two such crops horizontally to form a $224\times224$ input. We show an example in Appendix Figure~\ref{fig:insdel-ex}.
We present the results for the insertion test in Table \ref{tab:insertion} and for deletion in Appendix \ref{app:insdel}.

We observe that adding \ourmethod substantially improves insertion AUC for CNNs, often by around 8–11 \% on average, indicating much more actionable pixel rankings. Grad-CAM is the main exception; its large, blurry regions are well-suited to insertion since a big part of the image is introduced at once, so sharpening/refining them can slightly reduce insertion AUC even if the maps look cleaner.
For ViTs the gains are smaller because their baseline attributions are already relatively focused, so \ourmethod mainly denoises/cleans them up rather than dramatically re-localizing evidence. 

To demonstrate that \ourmethod can uncover concept-specific information encoded in attribution methods, we apply it to an image of a porcupine. Visually, the porcupine resembles a “Colobus × Echidna” hybrid (Figure \ref{fig:ablation-app}). We compute class-specific saliency maps for the two target classes and then mask the corresponding explanation regions. Figure \ref{fig:ablation} illustrates this procedure for \ourmethod with \gbp on ResNet-50 (additional examples are provided in Figure \ref{fig:ablation-app}).

We observe that \ourmethod highlights class-relevant concepts more clearly than the base method (Figure \ref{fig:crownjewel}). Importantly, this also indicates that the base method already contains the relevant concept information--however, it is not directly exposed. With \ourmethod, we can extract and isolate this latent signal. As expected, removing the Echidna evidence increases the Colobus probability, and removing the Colobus evidence increases the Echidna probability. 
This confirms that the \textbf{attributions refined with \ourmethod identify genuinely class-discriminative cues}.

\subsection{Sanity Checks}

To check whether attribution methods reflect what the model has learned, we run cascading randomization tests \citep{adebayo2018sanity}. We progressively randomize parameters from output to input layers and track how attribution maps change as model information is destroyed.
We follow the original protocol and report Spearman correlation between attributions before and after randomization (Fig.~\ref{fig:sanity-metrics-pearson-vit-resnet}). We also include cosine similarity and Pearson correlation across architectures in App. Fig.~\ref{fig:sanity-metrics-full-resnet}--\ref{fig:sanity-metrics-full-vit32}, which show similar trends. Ideally, once later layers are randomized, attribution maps should contain little target-relevant information.
As noted by \citet{binder2023shortcomings}, these tests have limitations because they can “preserve scales of forward pass activations with high probability.” We therefore focus on the \textit{relative change} in attributions with and without \ourmethod.
 
We find that attribution maps using the \ourmethod pipeline yield better results for all baseline methods and across randomization percentages. 
For Guided-Backprop and Input $\times$ Gradient, the improvement is most pronounced, as well as for randomizing the latest layers, which carry most of the conceptual meaning for the classification. 
Intriguingly, for ViT models, we observe that after randomization at any point in the network the similarity score is virtually zero, meaning that specific \textbf{attribution methods taking class contrast into account can pass the sanity check}.

\section{Discussion \& Conclusion}

Attribution methods are widely used but often criticized for failing to highlight decision-relevant features. We argue that attributing a single target logit misrepresents how classifiers make decisions, and instead proposed attribution distributions over multiple classes.
Through this change, we enable existing attribution methods to capture object- and concept-specific structure, revealing both class-discriminative and shared features that single-logit attributions overlook. Crucially, we show that even standard CNN attributions already encode rich class-specific signals--\emph{hidden in plain sight}.

To quantitatively substantiate these claims, we evaluated across 18 attribution methods, 7 CNN and transformer architectures, and common attribution benchmarks, including the grid pointing game~\cite{rao2022towards}, sanity checks~\cite{adebayo2018sanity}, and insertion/deletion tests. We acknowledge the interpretability literature offers many additional protocols, yet these widely adopted benchmarks provide a thorough quantitative assessment.

The reconsideration of where and how to apply attributions is method and model agnostic, training-free, and faithful to the target model in that we do not use surrogates or other, eg. generative, models that could introduce new biases.
Interestingly, \citet{gairola2025how} recently found that training with binary cross-entropy loss significantly improves attributions in terms of downstream benchmarks, arguing for BCE for improved post-hoc explanations.
Our findings provide a reason why this is the case, as BCE incentivizes the network to learn class-specific features, which will consequently appear in attributions even in the standard attribution pipeline looking at a logit in isolation.

Here, we compute attributions as distributions over classes using pre-computed attribution maps , which is training-free, thus maintaining the accuracy of the model, and not only reveals class-specific but also features shared across classes, which BCE discourages in training, and 
reveals object-specific attributions in multi-object settings.

The proposed refinement improves existing attribution methods by reintroducing the competitive mechanism of the softmax into the attribution process, thereby helping to \textit{understand the distinguishing features a model uses for prediction}. It is training-free and can be \textit{combined with any attribution method}, independent of architecture or attribution type.

\section*{Impact Statement}
In this work, we show that common attribution methods are better than commonly believed, but only if one uses them correctly.
By introducing a lightweight refinement, which we call Attribution Lens (\ourmethod), we are able to extract this information. Hence, our work
positively contributes to a better understanding the decision-making of neural networks and thus to ultimately a safer deployment of such approaches. Hence, we do not expect any potentially negative consequences caused by our work.

\small
\bibliographystyle{icml2026}
\bibliography{main}

\appendix
\onecolumn
\clearpage
\setcounter{page}{1}

\section{Method}

\subsection{Selecting the Set of Classes}
\label{sec:selectclasses}

Having defined our class-relevant attribution operator $\mathcal{C}_{\mathcal{H}}$, an important consideration is the selection of the set of classes $K$ used for calculation. 
We explore three approaches for class selection, each offering distinct advantages depending on the specific analysis goals and application context.

\textbf{Predefined Class Sets.} The canonical approach is to use a predefined set of classes $K$ that are of particular interest. This is especially useful in contexts where specific class comparisons have natural interpretations. For example, in a grid-pointing game where users must identify the quadrant containing a particular object, the four quadrant classes directly correspond to the task structure. Similarly, in medical applications, contrasting disease subtypes can highlight discriminative features that aid differential diagnosis. This approach ensures that the resulting attributions focus on distinctions that are meaningful to the particular application domain. However, this approach requires specific knowledge about the task, which is often not available.
The following approaches are data- and model-driven and, hence, do not require prior knowledge to select classes.

\textbf{Top-$k$ Most Probable Classes.} A model dependent approach to class selection involves choosing the $k$ classes with highest predicted probabilities and the class with the lowest probability for a given input. 
This approach is particularly effective for highlighting the features that distinguish between the most plausible classifications for a given input, but also reveal information that is shared between highly related classes that are likely among the highest probabilities. As these classes represent the top candidates for the final classification, contrasting their attribution maps reveals the most decision-relevant features.

\textbf{Best--vs--Worst Classes.} The third approach compares the highest-probability class against the lowest-probability class: $K = \{c_{\text{max}}, c_{\text{min}}\}$ where $c_{\text{max}} = \argmax_c S_c(x)$ and $c_{\text{min}} = \argmin_c S_c(x)$. 
Such extreme can surprisingly reveal the most distinctive characteristics of the input as interpreted by the model, by showing which features are most critical for pushing the model toward or away from certain classifications. 

\section{Evaluation Metrics}

In our experimental setup, we evaluate attribution methods across several metrics to assess their efficacy in highlighting relevant features for model predictions. We define an input as a vector $x \in \mathbb{R}^d$, and a model as a function $f : \mathbb{R}^d \rightarrow \mathbb{R}^C$, where $C$ is the number of classes in the classification problem. The final classification is performed via an argmax over $f(x)$. An explanation method provides an explanation map $\mathcal{A} : \mathbb{R}^d \times S \times \{1,...,C\} \rightarrow \mathbb{R}^d$ that maps an input, a model, and optionally a target class to an attribution map of the same shape as the input.

\subsection{Localization metrics}
\label{supp:metrics}
We evaluate attribution methods using two datasets: a Grid Pointing Game based on ImageNet and COCO dataset with segmentation masks. For both evaluations, we apply the same set of metrics, treating both bounding boxes and segmentation masks as regions of interest $R$ in the image. We match the region of interest with the correct attribution map  $\attmap_c$ i.e. for the first quadrant we also 
take the first attribution map. We only take the positive part of $\attmap_c$.
Before evaluation, we apply a Gaussian blur with a kernel size of $11 \times 11$ to the attribution maps

$$\tilde{\mathcal{A}}_c = \mathcal{G}_\sigma * \mathcal{A}_c\,,$$ 

where $\mathcal{G}_\sigma$ is a Gaussian kernel with standard deviation $\sigma$ and $*$ denotes the convolution operation. This preprocessing is common to prevent methods from being unduly rewarded for producing diffuse attribution maps.
We then compute the following metrics:

\subsubsection{Region Attribution}
We quantify what fraction of the total positive attribution falls within the region of interest by

$$\text{RA} = \frac{\sum_{i \in R} \tilde{\mathcal{A}}_c(i)}{\sum_i  \tilde{\mathcal{A}}_c(i)} \,.$$

\subsubsection{Intersection over Union (IoU)}
We compute the overlap between the attribution map and the region of interest as

$$\text{IoU} = \frac{|(\tilde{\mathcal{A}}_c \cap R|}{|\tilde{\mathcal{A}}_c \cup R|} \,.$$

\subsubsection{Precision and Recall}

To calculate precision and recall, we use the commen intersection-based formulas
$$\text{Precision} = \frac{|\tilde{\mathcal{A}}_c \cap R|}{|\tilde{\mathcal{A}}_c|}, ~~~~~\text{Recall} = \frac{|\tilde{\mathcal{A}}_c \cap R|}{|R|} \,.$$

\subsubsection{F1 Score}

To calculate F1, we make use of the previously defined precision and recall metrics, caluclating
$$\text{F1} = \frac{2 \cdot \text{Precision} \cdot \text{Recall}}{\text{Precision} + \text{Recall}} \,.$$

\section{LLM Use}
In this work, we used GPT-5 for both writing and coding support. On the writing side, it assisted with editing and condensing text to improve clarity. For coding, GPT-5 was used for debugging, providing autocomplete suggestions in VS Code, and generating code for LaTeX figures.

\section{Additional Resulst}
\subsection{Localization}\label{app:addres}

\begin{figure*}
	\begin{tikzpicture}
		\path (0,0) -- (16,0);
		\def\igX{4.4}
		\def\igY{0}
		\node (label_ig) at (\igX, \igY) {\small Integrated Gradients};
		
		\def\gbpX{9.3}
		\def\gbpY{0}
		\node (label_gbp) at (\gbpX, \gbpY) {\small Guided Backprop};
		
		\def\ixgX{14.25}
		\def\ixgY{0}
		\node (label_ixg) at (\ixgX, \ixgY) {\small \ixg};
	\end{tikzpicture}
    \centering
    \includegraphics[width=\linewidth, trim=0 0 0 0, clip]{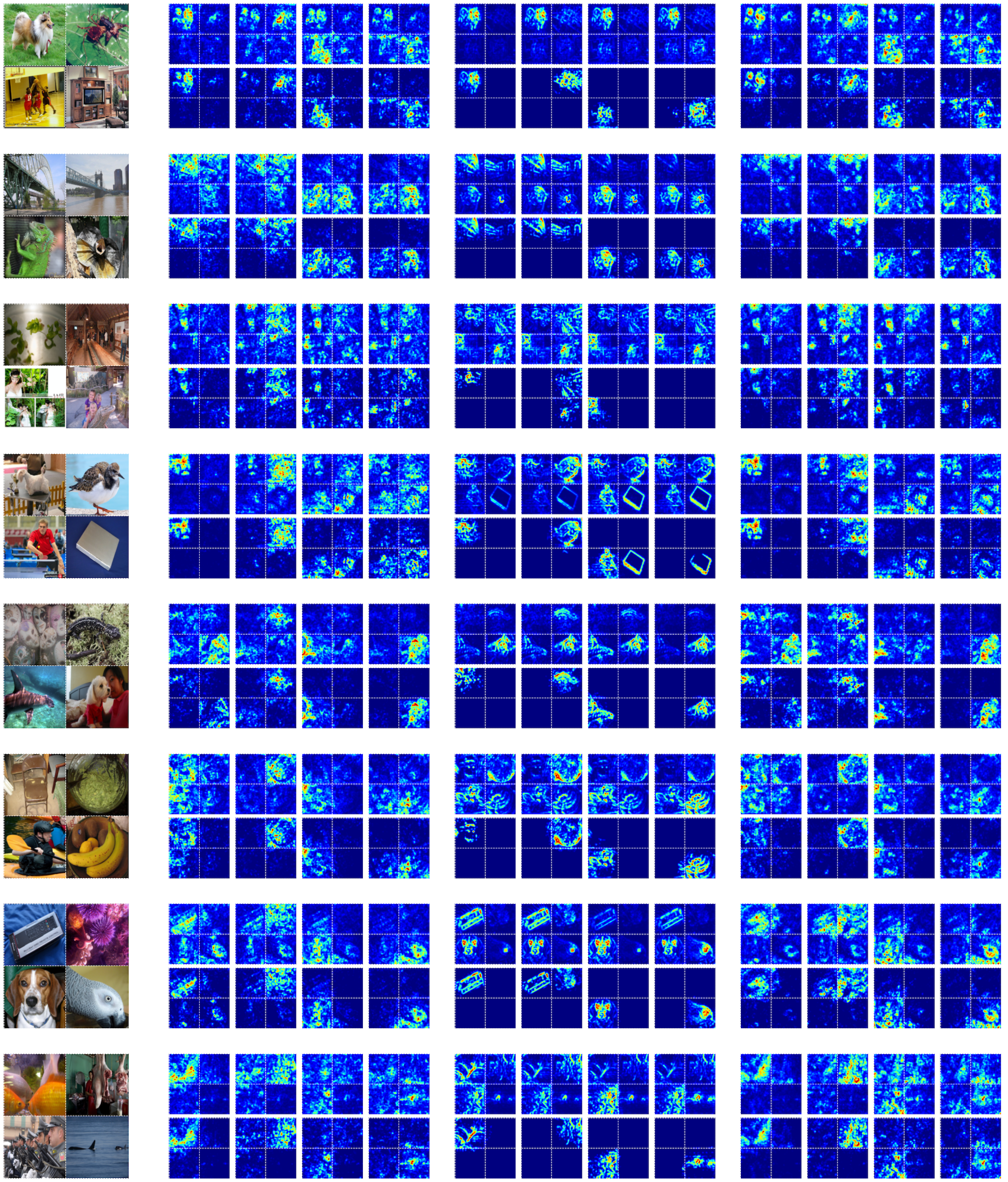} 
    \caption{\textbf{ResNet50}: \textit{\ourmethod on the Grid Pointing Game.} We show examples from the grid pointing game for methods most affected by our framework (as columns: \igfull, \gbpfull, \ixgfull). Input Images are given on the left, for each we provide vanilla attribution methods (top row) and augmented with \ourmethod (bottom row).
    For each, we show the attribution for the four different classes in the grid as columns.}
        \label{app:resnet}
\end{figure*}

We provide additional results for all the architectures mentioned in the Experiments in 
Table \ref{table:exp-localization-all}. The trend remains the same across architectures and methods;
if they are augmented using \ourmethod they improve the localization metrics and trade-off recall.
Additionally we provide plots similar to Figure \ref{fig:attr-comparison} for all these architectures in Figure \ref{app:resnet}-\ref{app:vit}.

\begin{table*}[]
    \centering

\resizebox{\textwidth}{!}{\begin{tabular}{cl@{\hspace{3em}}ccc@{\hspace{2em}}ccc@{\hspace{2em}}ccc}
\toprule
& & \multicolumn{3}{c}{Quad-ImageNet} & \multicolumn{3}{c}{Part-Quad-ImageNet} & \multicolumn{3}{c}{COCO} \\
\cmidrule(lr){3-5} \cmidrule(lr){6-8}  \cmidrule(lr){9-11}
& Method & RA & IoU & F1 & RA & IoU & F1 & RA & IoU & F1\\
\midrule
\multirow{6}{*}{\rotatebox[origin=c]{90}{Resnet50}}& GradCam & 0.88\varr{25} & 0.67\varr{64} & 0.79\varr{38} & 0.31\varr{28} & 0.24\varr{112} & 0.36\varr{87} & 0.18\varr{19} & 0.11\varr{16} & 0.17\varr{12} \\
& GBP & 0.86\varr{144} & 0.26\varr{32} & 0.41\varr{25} & 0.44\varr{146} & 0.08\varr{43} & 0.14\varr{38} & 0.19\varr{30} & 0.09\varr{3} & 0.15\varr{2} \\
& Guide-GC & 0.91\varr{21} & 0.34\varr{31} & 0.50\varr{23} & 0.50\varr{24} & 0.12\varr{49} & 0.21\varr{42} & 0.23\varr{16} & 0.10\varr{8} & 0.16\varr{8} \\
& IxG & 0.55\varr{37} & 0.20\textcolor{gray}{$+0\%$} & 0.33\textcolor{gray}{$+0\%$} & 0.25\varr{47} & 0.06\textcolor{gray}{$+0\%$} & 0.11\textcolor{gray}{$+0\%$} & 0.13\varr{11} & 0.09\textcolor{gray}{$+0\%$} & 0.15\textcolor{gray}{$+0\%$} \\
& IG & 0.56\varr{36} & 0.20\textcolor{gray}{$+0\%$} & 0.34\textcolor{gray}{$+0\%$} & 0.28\varr{48} & 0.06\textcolor{gray}{$+0\%$} & 0.12\textcolor{gray}{$+0\%$} & 0.14\varr{11} & 0.09\textcolor{gray}{$+0\%$} & 0.15\textcolor{gray}{$+0\%$} \\
& LRP & 0.88\varr{56} & 0.69\varr{97} & 0.79\varr{55} & 0.37\varr{49} & 0.22\varr{117} & 0.34\varr{90} & 0.21\varr{20} & 0.13\varr{8} & 0.20\varr{7} \\
\cmidrule[0.1pt](lr){3-11}
 & Avg. Improvement  &\varr{53.17} & \varr{44.8} & \varr{28.2} & \varr{57.0} & \varr{80.25} & \varr{64.25} & \varr{18.33} & \varr{6.17} & \varr{5.33} \\
\midrule
\multirow{7}{*}{\rotatebox[origin=c]{90}{Wide-Resnet502}}& GradCam & 0.88\varr{22} & 0.66\varr{62} & 0.78\varr{37} & 0.30\varr{23} & 0.24\varr{108} & 0.36\varr{84} & 0.18\varr{14} & 0.11\varr{11} & 0.17\varr{9} \\
& GBP & 0.89\varr{109} & 0.28\varr{42} & 0.43\varr{32} & 0.47\varr{108} & 0.09\varr{57} & 0.16\varr{51} & 0.20\varr{24} & 0.10\varr{2} & 0.16\varr{2} \\
& Guide-GC & 0.92\varr{17} & 0.35\varr{32} & 0.51\varr{24} & 0.51\varr{20} & 0.13\varr{53} & 0.22\varr{46} & 0.24\varr{13} & 0.10\varr{6} & 0.17\varr{6} \\
& IxG & 0.62\varr{38} & 0.20\textcolor{gray}{$+0\%$} & 0.33\textcolor{gray}{$+0\%$} & 0.27\varr{48} & 0.06\textcolor{gray}{$+0\%$} & 0.11\textcolor{gray}{$+0\%$} & 0.15\varr{12} & 0.10\textcolor{gray}{$+0\%$} & 0.15\textcolor{gray}{$+0\%$} \\
& IG & 0.62\varr{37} & 0.20\textcolor{gray}{$+0\%$} & 0.34\textcolor{gray}{$+0\%$} & 0.31\varr{48} & 0.06\textcolor{gray}{$+0\%$} & 0.12\textcolor{gray}{$+0\%$} & 0.15\varr{12} & 0.10\textcolor{gray}{$+0\%$} & 0.15\textcolor{gray}{$+0\%$} \\
& LRP & 0.89\varr{49} & 0.72\varr{115} & 0.82\varr{65} & 0.37\varr{46} & 0.22\varr{132} & 0.34\varr{102} & 0.22\varr{22} & 0.13\varr{11} & 0.20\varr{9} \\
\cmidrule[0.1pt](lr){3-11}
 & Avg. Improvement & \varr{45.33} & \varr{41.83} & \varr{26.33} & \varr{48.83} & \varr{58.33} & \varr{47.17} & \varr{16.17} & \varr{5.0} & \varr{4.33}\\
\midrule
\multirow{7}{*}{\rotatebox[origin=c]{90}{Densenet121}}& GradCam & 0.60\varr{17} & 0.37\varr{6} & 0.48\textcolor{darkred}{$-2\%$} & 0.22\varr{30} & 0.15\varr{51} & 0.23\varr{39} & 0.11\textcolor{darkred}{$-11\%$} & 0.07\textcolor{darkred}{$-25\%$} & 0.11\textcolor{darkred}{$-25\%$} \\
& GBP & 0.85\varr{158} & 0.25\varr{27} & 0.40\varr{21} & 0.41\varr{159} & 0.08\varr{35} & 0.14\varr{32} & 0.19\varr{36} & 0.10\varr{3} & 0.15\varr{3} \\
& Guide-GC & 0.71\varr{28} & 0.26\varr{12} & 0.40\varr{9} & 0.37\varr{34} & 0.10\varr{30} & 0.17\varr{26} & 0.17\varr{3} & 0.08\textcolor{darkred}{$-5\%$} & 0.14\textcolor{darkred}{$-5\%$} \\
& IxG & 0.46\varr{31} & 0.20\textcolor{gray}{$+0\%$} & 0.33\textcolor{gray}{$+0\%$} & 0.20\varr{42} & 0.06\textcolor{gray}{$+0\%$} & 0.11\textcolor{gray}{$+0\%$} & 0.13\varr{10} & 0.09\textcolor{gray}{$+0\%$} & 0.15\textcolor{gray}{$+0\%$} \\
& IG & 0.50\varr{34} & 0.20\textcolor{gray}{$+0\%$} & 0.34\textcolor{gray}{$+0\%$} & 0.24\varr{46} & 0.06\textcolor{gray}{$+0\%$} & 0.12\textcolor{gray}{$+0\%$} & 0.14\varr{11} & 0.09\textcolor{gray}{$+0\%$} & 0.15\textcolor{gray}{$+0\%$} \\
& LRP & 0.44\varr{24} & 0.25\textcolor{gray}{$+0\%$} & 0.40\textcolor{gray}{$+0\%$} & 0.19\varr{35} & 0.07\textcolor{gray}{$+0\%$} & 0.12\textcolor{gray}{$+0\%$} & 0.15\varr{5} & 0.12\textcolor{gray}{$+0\%$} & 0.18\textcolor{gray}{$+0\%$} \\
\cmidrule[0.1pt](lr){3-11}
 & Avg. Improvement & \varr{48.67} & \varr{7.5}  & \varr{4.67}  & \varr{57.67}  & \varr{19.33}  & \varr{16.17}  & \varr{9.0}  & \textcolor{darkred}{$-4.5\%$}  & \textcolor{darkred}{$-4.5\%$} \\
\midrule
\multirow{6}{*}{\rotatebox[origin=c]{90}{Convnext}}& GradCam & 0.96\varr{2} & 0.55\textcolor{darkred}{$-7\%$} & 0.70\textcolor{darkred}{$-6\%$} & 0.48\varr{8} & 0.29\varr{31} & 0.42\varr{24} & 0.28\varr{8} & 0.15\varr{2} & 0.23\varr{2} \\
& GBP & 0.52\varr{26} & 0.20\textcolor{gray}{$+0\%$} & 0.33\textcolor{gray}{$+0\%$} & 0.19\varr{33} & 0.06\textcolor{gray}{$+0\%$} & 0.11\textcolor{gray}{$+0\%$} & 0.15\varr{13} & 0.09\textcolor{gray}{$+0\%$} & 0.15\textcolor{gray}{$+0\%$} \\
& Guide-GC & 0.96\varr{1} & 0.35\varr{1} & 0.52\varr{1} & 0.58\varr{5} & 0.16\varr{2} & 0.26\varr{2} & 0.31\varr{5} & 0.14\varr{1} & 0.22\varr{1} \\
& IxG & 0.51\varr{27} & 0.20\textcolor{gray}{$+0\%$} & 0.33\textcolor{gray}{$+0\%$} & 0.19\varr{33} & 0.06\textcolor{gray}{$+0\%$} & 0.11\textcolor{gray}{$+0\%$} & 0.15\varr{13} & 0.09\textcolor{gray}{$+0\%$} & 0.15\textcolor{gray}{$+0\%$} \\
& IG & 0.64\varr{35} & 0.21\textcolor{gray}{$+0\%$} & 0.34\textcolor{gray}{$+0\%$} & 0.26\varr{49} & 0.06\varr{1} & 0.12\varr{1} & 0.15\varr{16} & 0.09\textcolor{gray}{$+0\%$} & 0.15\textcolor{gray}{$+0\%$} \\
\cmidrule[0.1pt](lr){3-11}
& Avg. Improvement & \varr{18.20} & \textcolor{darkred}{$-1.20\%$} & \textcolor{darkred}{$-1\%$} & \varr{25.60} & \varr{6.80} & \varr{5.40} & \varr{11.00} & \varr{0.60} & \varr{0.60} \\
\midrule
\midrule
\multirow{7}{*}{\rotatebox[origin=c]{90}{ViT-base-8}}& Bi-attn & 0.91\varr{48} & 0.62\varr{149} & 0.76\varr{89} & 0.56\varr{61} & 0.25\varr{272} & 0.36\varr{199} & 0.29\varr{45} & 0.14\varr{32} & 0.21\varr{27} \\
& GradCam & 0.83\varr{8} & 0.49\varr{18} & 0.64\varr{12} & 0.61\varr{11} & 0.28\varr{46} & 0.40\varr{36} & 0.30\varr{13} & 0.14\varr{9} & 0.21\varr{7} \\
& InFlow & 0.82\varr{18} & 0.47\varr{89} & 0.63\varr{58} & 0.59\varr{19} & 0.18\varr{165} & 0.28\varr{131} & 0.32\varr{18} & 0.12\varr{14} & 0.19\varr{13} \\
& Grad-Rollout & 0.71\varr{51} & 0.45\varr{80} & 0.61\varr{53} & 0.48\varr{60} & 0.20\varr{197} & 0.30\varr{147} & 0.26\varr{27} & 0.12\varr{14} & 0.19\varr{12} \\
& T-attn & 0.90\varr{53} & 0.63\varr{152} & 0.76\varr{90} & 0.51\varr{76} & 0.28\varr{322} & 0.40\varr{230} & 0.28\varr{56} & 0.14\varr{34} & 0.22\varr{29} \\
& LRP & 0.76\varr{25} & 0.42\varr{69} & 0.58\varr{46} & 0.54\varr{24} & 0.20\varr{195} & 0.30\varr{148} & 0.28\varr{16} & 0.12\varr{14} & 0.19\varr{13} \\
& Gradient & 0.90\varr{7} & 0.49\varr{7} & 0.64\varr{5} & 0.57\varr{11} & 0.35\varr{20} & 0.48\varr{16} & 0.31\varr{17} & 0.16\varr{2} & 0.23\varr{1} \\
\cmidrule[0.1pt](lr){3-11}
 & Avg. Improvement  & \varr{30.0} & \varr{80.57} & \varr{50.43} & \varr{37.43} & \varr{173.86} & \varr{129.57} & \varr{27.43} & \varr{17.0} & \varr{14.57}\\
\midrule
\multirow{7}{*}{\rotatebox[origin=c]{90}{ViT-base-16}}& Bi-attn & 0.94\varr{31} & 0.71\varr{180} & 0.82\varr{103} & 0.51\varr{40} & 0.28\varr{309} & 0.40\varr{222} & 0.30\varr{43} & 0.16\varr{52} & 0.23\varr{42} \\
& GradCam & 0.91\varr{6} & 0.62\varr{16} & 0.75\varr{10} & 0.58\varr{11} & 0.27\varr{39} & 0.39\varr{32} & 0.31\varr{10} & 0.15\varr{11} & 0.22\varr{9} \\
& InFlow & 0.86\varr{21} & 0.56\varr{126} & 0.71\varr{78} & 0.53\varr{23} & 0.20\varr{198} & 0.31\varr{153} & 0.29\varr{20} & 0.13\varr{23} & 0.20\varr{21} \\
& Grad-Rollout & 0.73\varr{76} & 0.53\varr{113} & 0.68\varr{71} & 0.40\varr{94} & 0.20\varr{197} & 0.30\varr{148} & 0.24\varr{30} & 0.12\varr{19} & 0.19\varr{17} \\
& T-attn & 0.93\varr{32} & 0.71\varr{180} & 0.82\varr{102} & 0.47\varr{38} & 0.29\varr{321} & 0.40\varr{229} & 0.29\varr{44} & 0.16\varr{53} & 0.23\varr{43} \\
& LRP & 0.77\varr{35} & 0.51\varr{105} & 0.66\varr{65} & 0.47\varr{36} & 0.20\varr{201} & 0.31\varr{152} & 0.27\varr{17} & 0.12\varr{20} & 0.19\varr{18} \\
& Gradient & 0.93\varr{4} & 0.57\varr{3} & 0.70\varr{2} & 0.50\varr{8} & 0.34\varr{11} & 0.47\varr{9} & 0.30\varr{10} & 0.17\varr{2} & 0.25\varr{2} \\
\cmidrule[0.1pt](lr){3-11}
 & Avg. Improvement  & \varr{29.29} & \varr{103.29} & \varr{61.57} & \varr{35.71} & \varr{182.29} & \varr{135.0} & \varr{24.86} & \varr{25.71} & \varr{21.71} \\
\midrule
\multirow{7}{*}{\rotatebox[origin=c]{90}{ViT-base-32}}& Bi-attn & 0.86\varr{71} & 0.62\varr{149} & 0.75\varr{87} & 0.36\varr{80} & 0.24\varr{263} & 0.36\varr{195} & 0.21\varr{37} & 0.13\varr{33} & 0.20\varr{27} \\
& GradCam & 0.78\varr{18} & 0.51\varr{50} & 0.65\varr{30} & 0.41\varr{28} & 0.22\varr{119} & 0.32\varr{91} & 0.22\varr{25} & 0.13\varr{15} & 0.20\varr{13} \\
& InFlow & 0.78\varr{21} & 0.56\varr{124} & 0.70\varr{75} & 0.38\varr{24} & 0.19\varr{176} & 0.29\varr{136} & 0.22\varr{19} & 0.12\varr{21} & 0.19\varr{17} \\
& Grad-Rollout & 0.66\varr{91} & 0.51\varr{106} & 0.66\varr{66} & 0.27\varr{112} & 0.17\varr{151} & 0.27\varr{119} & 0.17\varr{28} & 0.11\varr{13} & 0.18\varr{12} \\
& T-attn & 0.84\varr{70} & 0.62\varr{146} & 0.74\varr{86} & 0.35\varr{77} & 0.25\varr{267} & 0.36\varr{197} & 0.20\varr{35} & 0.14\varr{34} & 0.20\varr{27} \\
& LRP & 0.66\varr{49} & 0.46\varr{85} & 0.61\varr{53} & 0.31\varr{51} & 0.17\varr{147} & 0.26\varr{115} & 0.20\varr{16} & 0.11\varr{11} & 0.18\varr{9} \\
& Gradient & 0.79\varr{19} & 0.51\varr{19} & 0.65\varr{12} & 0.36\varr{27} & 0.23\varr{53} & 0.34\varr{42} & 0.21\varr{17} & 0.13\varr{11} & 0.20\varr{8} \\
\cmidrule[0.1pt](lr){3-11}
 & Avg. Improvement  &\varr{48.43} & \varr{97.0} & \varr{58.43} & \varr{57.0} & \varr{168.0} & \varr{127.86} & \varr{25.29} & \varr{19.71} & \varr{16.14} \\
\bottomrule

\end{tabular}}
    \caption{\textit{Consistent improvement of attributions.} Across 11 different  attribution methods considering convolutional and transformer based architectures, quantitative metrics measured using Region Attribution (RA), Intersection over Union (IoU), and F1 get consistently improved by a wide margin.\vspace{0.4cm}}
    \label{table:exp-localization-all}
\end{table*}

\subsection{Perturbation based methods}
\label{app:perturbation}
We compute perturbation-based attribution maps with and  provide additional results for all architectures used in the Experiments in Table \ref{table:exp-localization-all}. For Shapley Value Sampling, KernelSHAP, Feature Ablation, and LIME we define “features” as SLIC superpixels (fixed to 100 segments with compactness 10) and attribute the target class by perturbing superpixels against a common baseline. Occlusion instead perturbs fixed square patches: we occlude $15×15$ regions (jointly over all channels) placed on a regular grid with stride 8 pixels; overlapping occlusions are aggregated to obtain a dense heatmap. Unless stated otherwise, we keep method hyperparameters fixed across models (e.g., 
$n_{samples}=64$ for Shapley/KernelSHAP and $n_{samples}=1000$ for LIME).

Across both Quad-ImageNet  and Part-Quad-ImageNet we observe a consistent and strict improvement when enabling VAR, indicating better localization quality under the same evaluation protocol. On COCO, VAR increases region accuracy (RA), but degrades F1 and IoU; this discrepancy is primarily driven by the superpixel-based perturbation setup. In particular, the superpixels (and resulting patches) often fail to align with full object extents in cluttered multi-object scenes, which reduces coverage of the ground-truth regions and leads to a pronounced drop in recall--ultimately hurting F1 and IoU despite improved RA.

\begin{table*}[]
    \centering

\resizebox{\textwidth}{!}{\begin{tabular}{cl@{\hspace{3em}}ccc@{\hspace{2em}}ccc@{\hspace{2em}}ccc}
\toprule
& & \multicolumn{3}{c}{Quad-ImageNet} & \multicolumn{3}{c}{Part-Quad-ImageNet} & \multicolumn{3}{c}{COCO} \\
\cmidrule(lr){3-5} \cmidrule(lr){6-8}  \cmidrule(lr){9-11}
& Method & RA & IoU & F1 & RA & IoU & F1 & RA & IoU & F1\\
\midrule
\multirow{5}{*}{\rotatebox[origin=c]{90}{ResNet50}}& Feature Ablation & 0.71\varr{31} & 0.35\varr{33} & 0.51\varr{22} & 0.40\varr{42} & 0.14\varr{69} & 0.23\varr{57} & 0.20\varr{13} & 0.11\textcolor{darkred}{$-1\%$} & 0.17\varr{1} \\
& Kernel SV & 0.51\varr{23} & 0.30\varr{12} & 0.46\varr{7} & 0.26\varr{36} & 0.12\varr{44} & 0.20\varr{38} & 0.16\varr{3} & 0.11\textcolor{darkred}{$-7\%$} & 0.17\textcolor{darkred}{$-4\%$} \\
& LIME & 0.85\varr{24} & 0.48\varr{43} & 0.64\varr{28} & 0.50\varr{32} & 0.17\varr{71} & 0.27\varr{58} & 0.27\varr{14} & 0.13\varr{5} & 0.21\varr{6} \\
& Occlusion & 0.69\varr{30} & 0.32\varr{24} & 0.48\varr{17} & 0.41\varr{45} & 0.11\varr{50} & 0.20\varr{43} & 0.22\varr{14} & 0.11\textcolor{gray}{$+0\%$} & 0.18\varr{2} \\
& Shapley Values & 0.83\varr{27} & 0.47\varr{52} & 0.64\varr{34} & 0.49\varr{35} & 0.17\varr{83} & 0.27\varr{67} & 0.26\varr{17} & 0.13\varr{6} & 0.20\varr{8} \\
\cmidrule[0.1pt](lr){3-11}
 & Avg. Improvement & \varr{27.00} & \varr{32.80} & \varr{21.60} & \varr{38.00} & \varr{63.40} & \varr{52.60} & \varr{12.20} & \varr{0.60} & \varr{2.60} \\
\midrule
\multirow{5}{*}{\rotatebox[origin=c]{90}{WRN50-2}}& Feature Ablation & 0.74\varr{28} & 0.36\varr{37} & 0.52\varr{25} & 0.40\varr{35} & 0.14\varr{68} & 0.23\varr{57} & 0.19\varr{12} & 0.10\textcolor{darkred}{$-1\%$} & 0.16\varr{0} \\
& Kernel SV & 0.53\varr{24} & 0.31\varr{13} & 0.46\varr{9} & 0.26\varr{36} & 0.12\varr{44} & 0.21\varr{38} & 0.15\varr{4} & 0.11\textcolor{darkred}{$-4\%$} & 0.17\textcolor{darkred}{$-2\%$} \\
& LIME & 0.87\varr{20} & 0.51\varr{45} & 0.66\varr{29} & 0.50\varr{28} & 0.18\varr{71} & 0.28\varr{58} & 0.27\varr{14} & 0.12\varr{4} & 0.20\varr{5} \\
& Occlusion & 0.72\varr{28} & 0.33\varr{28} & 0.49\varr{20} & 0.41\varr{39} & 0.12\varr{49} & 0.20\varr{42} & 0.19\varr{11} & 0.10\textcolor{darkred}{$-2\%$} & 0.16\textcolor{gray}{$+0\%$} \\
& Shapley Values & 0.86\varr{23} & 0.51\varr{57} & 0.66\varr{37} & 0.50\varr{29} & 0.18\varr{84} & 0.28\varr{68} & 0.26\varr{18} & 0.12\varr{8} & 0.19\varr{9} \\
\cmidrule[0.1pt](lr){3-11}
 & Avg. Improvement & \varr{24.60} & \varr{36.00} & \varr{24.00} & \varr{33.40} & \varr{63.20} & \varr{52.60} & \varr{11.80} & \varr{1.00} & \varr{2.40} \\
\midrule
\multirow{5}{*}{\rotatebox[origin=c]{90}{DenseNet121}}& Feature Ablation & 0.63\varr{30} & 0.31\varr{22} & 0.46\varr{14} & 0.37\varr{38} & 0.13\varr{60} & 0.22\varr{50} & 0.20\varr{12} & 0.11\textcolor{darkred}{$-3\%$} & 0.17\textcolor{darkred}{$-1\%$} \\
& Kernel SV & 0.50\varr{22} & 0.29\varr{9} & 0.44\varr{6} & 0.25\varr{35} & 0.12\varr{41} & 0.20\varr{35} & 0.17\varr{3} & 0.11\textcolor{darkred}{$-5\%$} & 0.18\textcolor{darkred}{$-3\%$} \\
& LIME & 0.82\varr{25} & 0.45\varr{37} & 0.61\varr{24} & 0.49\varr{31} & 0.16\varr{67} & 0.27\varr{55} & 0.28\varr{14} & 0.14\varr{5} & 0.21\varr{6} \\
& Occlusion & 0.65\varr{28} & 0.32\varr{20} & 0.48\varr{14} & 0.38\varr{39} & 0.12\varr{45} & 0.20\varr{38} & 0.21\varr{13} & 0.11\textcolor{darkred}{$-1\%$} & 0.18\varr{1} \\
& Shapley Values & 0.80\varr{28} & 0.45\varr{45} & 0.61\varr{30} & 0.48\varr{34} & 0.16\varr{78} & 0.27\varr{63} & 0.27\varr{16} & 0.13\varr{6} & 0.21\varr{8} \\
\cmidrule[0.1pt](lr){3-11}
 & Avg. Improvement & \varr{26.60} & \varr{26.60} & \varr{17.60} & \varr{35.40} & \varr{58.20} & \varr{48.20} & \varr{11.60} & \varr{0.40} & \varr{2.20} \\
\midrule
\multirow{5}{*}{\rotatebox[origin=c]{90}{ConvNeXT}}& Feature Ablation & 0.67\varr{29} & 0.33\varr{28} & 0.48\varr{19} & 0.30\varr{35} & 0.10\varr{38} & 0.18\varr{33} & 0.18\varr{16} & 0.10\textcolor{darkred}{$-4\%$} & 0.16\textcolor{darkred}{$-1\%$} \\
& Kernel SV & 0.57\varr{27} & 0.32\varr{16} & 0.48\varr{11} & 0.25\varr{42} & 0.12\varr{46} & 0.20\varr{41} & 0.16\varr{6} & 0.10\textcolor{darkred}{$-4\%$} & 0.17\textcolor{darkred}{$-1\%$} \\
& LIME & 0.90\varr{15} & 0.52\varr{43} & 0.68\varr{28} & 0.49\varr{21} & 0.19\varr{58} & 0.30\varr{46} & 0.30\varr{9} & 0.12\varr{4} & 0.20\varr{4} \\
& Occlusion & 0.57\varr{26} & 0.28\varr{14} & 0.43\varr{11} & 0.21\varr{32} & 0.07\varr{14} & 0.13\varr{12} & 0.18\varr{12} & 0.10\textcolor{darkred}{$-4\%$} & 0.15\textcolor{darkred}{$-2\%$} \\
& Shapley Values & 0.92\varr{17} & 0.58\varr{79} & 0.73\varr{49} & 0.49\varr{26} & 0.19\varr{111} & 0.30\varr{88} & 0.27\varr{18} & 0.12\varr{10} & 0.20\varr{11} \\
\cmidrule[0.1pt](lr){3-11}
 & Avg. Improvement & \varr{22.80} & \varr{36.00} & \varr{23.60} & \varr{31.20} & \varr{53.40} & \varr{44.00} & \varr{12.20} & \varr{0.40} & \varr{2.20} \\
\midrule
\midrule
\multirow{5}{*}{\rotatebox[origin=c]{90}{ViT-base-16}}& Feature Ablation & 0.59\varr{32} & 0.29\varr{18} & 0.44\varr{11} & 0.30\varr{46} & 0.11\varr{50} & 0.19\varr{42} & 0.18\varr{15} & 0.10\textcolor{darkred}{$-3\%$} & 0.17\textcolor{darkred}{$-1\%$} \\
& Kernel SV & 0.53\varr{26} & 0.30\varr{14} & 0.46\varr{9} & 0.25\varr{39} & 0.12\varr{45} & 0.20\varr{39} & 0.16\varr{5} & 0.11\textcolor{darkred}{$-4\%$} & 0.18\textcolor{darkred}{$-2\%$} \\
& LIME & 0.88\varr{18} & 0.50\varr{48} & 0.65\varr{31} & 0.49\varr{24} & 0.17\varr{72} & 0.28\varr{58} & 0.31\varr{12} & 0.14\varr{5} & 0.22\varr{5} \\
& Occlusion & 0.53\varr{31} & 0.27\varr{12} & 0.41\varr{8} & 0.27\varr{45} & 0.09\varr{27} & 0.15\varr{24} & 0.19\varr{14} & 0.10\textcolor{darkred}{$-3\%$} & 0.17\textcolor{darkred}{$-1\%$} \\
& Shapley Values & 0.86\varr{24} & 0.49\varr{63} & 0.65\varr{41} & 0.47\varr{32} & 0.17\varr{95} & 0.27\varr{77} & 0.28\varr{19} & 0.13\varr{9} & 0.21\varr{10} \\
\cmidrule[0.1pt](lr){3-11}
 & Avg. Improvement & \varr{26.20} & \varr{31.00} & \varr{20.00} & \varr{37.20} & \varr{57.80} & \varr{48.00} & \varr{13.00} & \varr{0.80} & \varr{2.20} \\
\midrule
\multirow{5}{*}{\rotatebox[origin=c]{90}{ViT-base-8}}& Feature Ablation & 0.57\varr{28} & 0.26\varr{12} & 0.40\varr{7} & 0.30\varr{36} & 0.10\varr{39} & 0.18\varr{33} & 0.18\varr{9} & 0.10\textcolor{darkred}{$-4\%$} & 0.16\textcolor{darkred}{$-3\%$} \\
& Kernel SV & 0.39\varr{13} & 0.25\textcolor{darkred}{$-2\%$} & 0.39\textcolor{darkred}{$-3\%$} & 0.16\varr{24} & 0.10\varr{24} & 0.17\varr{21} & 0.14\varr{4} & 0.11\textcolor{darkred}{$-5\%$} & 0.17\textcolor{darkred}{$-3\%$} \\
& LIME & 0.90\varr{16} & 0.49\varr{44} & 0.65\varr{28} & 0.51\varr{19} & 0.18\varr{63} & 0.28\varr{52} & 0.30\varr{10} & 0.13\varr{2} & 0.21\varr{3} \\
& Occlusion & 0.42\varr{28} & 0.21\textcolor{gray}{$+0\%$} & 0.33\textcolor{darkred}{$-1\%$} & 0.19\varr{42} & 0.06\varr{12} & 0.12\varr{10} & 0.16\varr{8} & 0.09\textcolor{darkred}{$-8\%$} & 0.15\textcolor{darkred}{$-6\%$} \\
& Shapley Values & 0.88\varr{20} & 0.49\varr{60} & 0.65\varr{39} & 0.49\varr{26} & 0.17\varr{87} & 0.27\varr{70} & 0.27\varr{16} & 0.13\varr{6} & 0.20\varr{7} \\
\cmidrule[0.1pt](lr){3-11}
 & Avg. Improvement & \varr{21.00} & \varr{22.80} & \varr{14.00} & \varr{29.40} & \varr{45.00} & \varr{37.20} & \varr{9.40} & \textcolor{darkred}{$-1.80\%$} & \textcolor{darkred}{$-0.40\%$} \\
\midrule
\multirow{5}{*}{\rotatebox[origin=c]{90}{ViT-base-32}}& Feature Ablation & 0.63\varr{26} & 0.34\varr{22} & 0.49\varr{14} & 0.29\varr{32} & 0.13\varr{48} & 0.21\varr{40} & 0.19\varr{9} & 0.11\textcolor{darkred}{$-4\%$} & 0.17\textcolor{darkred}{$-2\%$} \\
& Kernel SV & 0.42\varr{17} & 0.27\varr{2} & 0.41\varr{0} & 0.17\varr{26} & 0.10\varr{26} & 0.17\varr{23} & 0.15\varr{1} & 0.11\textcolor{darkred}{$-7\%$} & 0.17\textcolor{darkred}{$-4\%$} \\
& LIME & 0.72\varr{26} & 0.37\varr{24} & 0.53\varr{16} & 0.37\varr{40} & 0.14\varr{58} & 0.23\varr{48} & 0.26\varr{14} & 0.13\varr{5} & 0.20\varr{6} \\
& Occlusion & 0.62\varr{26} & 0.31\varr{17} & 0.47\varr{12} & 0.29\varr{35} & 0.11\varr{35} & 0.18\varr{30} & 0.20\varr{10} & 0.11\textcolor{darkred}{$-2\%$} & 0.18\textcolor{gray}{$+0\%$} \\
& Shapley Values & 0.72\varr{28} & 0.40\varr{35} & 0.56\varr{23} & 0.37\varr{42} & 0.14\varr{69} & 0.24\varr{57} & 0.25\varr{17} & 0.13\varr{6} & 0.20\varr{7} \\
\cmidrule[0.1pt](lr){3-11}
 & Avg. Improvement & \varr{24.60} & \varr{20.00} & \varr{13.00} & \varr{35.00} & \varr{47.20} & \varr{39.60} & \varr{10.20} & \textcolor{darkred}{$-0.40\%$} & \varr{1.40} \\

\bottomrule
\end{tabular}}
    \caption{Across 5 different perturbation based attribution methods considering convolutional and transformer based architectures, quantitative metrics measured using Region Attribution (RA), Intersection over Union (IoU), and F1 get consistently improved.}
    \label{table:exp-localization-all2}
\end{table*}

\subsection{Sensitivity of \ourmethod regarding scaling parameters}
\label{appsec:scalingpars}
In this section, we investigate the sensitivity with respect to the hyperparameter of the scaling as described in Section \ref{subsec:method}. In all our experiments, we use the parameters $t = \{ 1, t_1, t_2\}$ with $t_1=5$ and $t_2=100$. We now vary $t_1 \in \{2,5,10\}$ and $t_2 \in \{50,100,500\}$. For this experiment we focus on the standard gridpointing game as described in Section \ref{para:gridpointing}. We present the results for all architectures and attribution methods as in the original setup in Table \ref{table:resnet50-abl} - \ref{table:vit_timm_32-abl}.

\begin{table*}[]
\centering

\resizebox{\textwidth}{!}{\begin{tabular}{cl@{\hspace{3em}}ccc@{\hspace{2em}}ccc@{\hspace{2em}}ccc}
& & \multicolumn{3}{c}{} & \multicolumn{3}{c}{ResNet50} & \multicolumn{3}{c}{} \\
\toprule
& & \multicolumn{3}{c}{$t_1=2$} & \multicolumn{3}{c}{$t_1=5$} & \multicolumn{3}{c}{$t_1=10$} \\
\midrule
& Method & RA & IoU & F1 & RA & IoU & F1 & RA & IoU & F1\\
\cmidrule(lr){3-5} \cmidrule(lr){6-8}  \cmidrule(lr){9-11}

\multirow{6}{*}{\rotatebox[origin=c]{90}{$t_2 = 50$}}& GradCam & 0.88\varr{25} & 0.67\varr{64} & 0.79\varr{39} & 0.88\varr{25} & 0.67\varr{64} & 0.79\varr{39} & 0.88\varr{25} & 0.67\varr{64} & 0.79\varr{39} \\
& GBP & 0.85\varr{140} & 0.26\varr{32} & 0.40\varr{25} & 0.85\varr{140} & 0.26\varr{32} & 0.40\varr{25} & 0.85\varr{140} & 0.26\varr{32} & 0.40\varr{25} \\
& Guide-GC & 0.91\varr{21} & 0.34\varr{31} & 0.50\varr{23} & 0.91\varr{21} & 0.34\varr{30} & 0.50\varr{23} & 0.91\varr{21} & 0.34\varr{30} & 0.50\varr{23} \\
& IxG & 0.54\varr{33} & 0.20\varr{0} & 0.33\varr{0} & 0.54\varr{34} & 0.20\varr{0} & 0.33\varr{0} & 0.54\varr{35} & 0.20\varr{0} & 0.33\varr{0} \\
& IG & 0.55\varr{33} & 0.20\textcolor{gray}{$+0\%$} & 0.34\textcolor{gray}{$+0\%$} & 0.56\varr{34} & 0.20\textcolor{gray}{$+0\%$} & 0.34\textcolor{gray}{$+0\%$} & 0.56\varr{35} & 0.20\textcolor{gray}{$+0\%$} & 0.34\textcolor{gray}{$+0\%$} \\
& LRP & 0.87\varr{54} & 0.68\varr{95} & 0.79\varr{54} & 0.87\varr{54} & 0.68\varr{94} & 0.79\varr{54} & 0.87\varr{54} & 0.68\varr{94} & 0.79\varr{54} \\
\midrule
\multirow{6}{*}{\rotatebox[origin=c]{90}{$t_2 = 100$}}& GradCam & 0.88\varr{25} & 0.67\varr{64} & 0.79\varr{38} & 0.88\varr{25} & 0.67\varr{64} & 0.79\varr{38} & 0.88\varr{25} & 0.67\varr{64} & 0.79\varr{38} \\
& GBP & 0.86\varr{144} & 0.26\varr{32} & 0.41\varr{25} & 0.86\varr{144} & 0.26\varr{32} & 0.41\varr{25} & 0.87\varr{145} & 0.26\varr{32} & 0.40\varr{25} \\
& Guide-GC & 0.91\varr{21} & 0.34\varr{31} & 0.50\varr{23} & 0.91\varr{21} & 0.34\varr{31} & 0.50\varr{23} & 0.92\varr{22} & 0.34\varr{30} & 0.50\varr{23} \\
& IxG & 0.55\varr{36} & 0.20\varr{0} & 0.33\varr{0} & 0.55\varr{37} & 0.20\varr{0} & 0.33\varr{0} & 0.55\varr{38} & 0.20\varr{0} & 0.33\varr{0} \\
& IG & 0.56\varr{35} & 0.20\textcolor{gray}{$+0\%$} & 0.34\textcolor{gray}{$+0\%$} & 0.56\varr{36} & 0.20\textcolor{gray}{$+0\%$} & 0.34\textcolor{gray}{$+0\%$} & 0.57\varr{37} & 0.20\textcolor{gray}{$+0\%$} & 0.34\textcolor{gray}{$+0\%$} \\
& LRP & 0.88\varr{56} & 0.69\varr{97} & 0.79\varr{55} & 0.88\varr{56} & 0.69\varr{97} & 0.79\varr{55} & 0.88\varr{56} & 0.69\varr{96} & 0.79\varr{55} \\
\midrule
\multirow{6}{*}{\rotatebox[origin=c]{90}{$t_2 = 500$}}& GradCam & 0.88\varr{25} & 0.66\varr{63} & 0.78\varr{38} & 0.88\varr{25} & 0.66\varr{63} & 0.78\varr{38} & 0.88\varr{25} & 0.67\varr{63} & 0.78\varr{38} \\
& GBP & 0.89\varr{153} & 0.26\varr{33} & 0.41\varr{26} & 0.89\varr{153} & 0.26\varr{33} & 0.41\varr{26} & 0.90\varr{153} & 0.26\varr{33} & 0.41\varr{26} \\
& Guide-GC & 0.92\varr{23} & 0.34\varr{31} & 0.50\varr{23} & 0.92\varr{23} & 0.34\varr{31} & 0.50\varr{23} & 0.92\varr{23} & 0.34\varr{31} & 0.50\varr{23} \\
& IxG & 0.55\varr{37} & 0.20\varr{0} & 0.33\varr{0} & 0.55\varr{38} & 0.20\varr{0} & 0.33\varr{0} & 0.56\varr{39} & 0.20\varr{0} & 0.33\varr{0} \\
& IG & 0.56\varr{35} & 0.20\textcolor{gray}{$+0\%$} & 0.34\textcolor{gray}{$+0\%$} & 0.57\varr{36} & 0.20\textcolor{gray}{$+0\%$} & 0.34\textcolor{gray}{$+0\%$} & 0.57\varr{37} & 0.20\textcolor{gray}{$+0\%$} & 0.34\textcolor{gray}{$+0\%$} \\
& LRP & 0.89\varr{58} & 0.70\varr{99} & 0.80\varr{56} & 0.89\varr{58} & 0.70\varr{99} & 0.80\varr{56} & 0.89\varr{58} & 0.70\varr{99} & 0.80\varr{56} \\

\bottomrule

\end{tabular}}
    \caption{\textit{Low variation between scaling parameters.} Across 6 different attribution methods considering convolutional and transformer based architectures, quantitative metrics measured using Region Attribution (RA), Intersection over Union (IoU), and F1 vary very slightly across various hyperparameter selections.}
    \label{table:resnet50-abl}
\end{table*}

\begin{table*}[]
\centering

\resizebox{\textwidth}{!}{\begin{tabular}{cl@{\hspace{3em}}ccc@{\hspace{2em}}ccc@{\hspace{2em}}ccc}
& & \multicolumn{3}{c}{} & \multicolumn{3}{c}{WideResNet50-2} & \multicolumn{3}{c}{} \\
\toprule
& & \multicolumn{3}{c}{$t_1=2$} & \multicolumn{3}{c}{$t_1=5$} & \multicolumn{3}{c}{$t_1=10$} \\
\midrule
& Method & RA & IoU & F1 & RA & IoU & F1 & RA & IoU & F1\\
\cmidrule(lr){3-5} \cmidrule(lr){6-8}  \cmidrule(lr){9-11}

\multirow{6}{*}{\rotatebox[origin=c]{90}{$t_2 = 50$}}& GradCam & 0.88\varr{21} & 0.66\varr{62} & 0.79\varr{37} & 0.88\varr{22} & 0.66\varr{62} & 0.79\varr{37} & 0.88\varr{22} & 0.66\varr{62} & 0.79\varr{37} \\
& GBP & 0.87\varr{106} & 0.28\varr{41} & 0.43\varr{32} & 0.88\varr{106} & 0.28\varr{41} & 0.43\varr{32} & 0.88\varr{107} & 0.28\varr{41} & 0.43\varr{32} \\
& Guide-GC & 0.92\varr{16} & 0.35\varr{32} & 0.51\varr{24} & 0.92\varr{16} & 0.35\varr{32} & 0.51\varr{24} & 0.92\varr{16} & 0.35\varr{33} & 0.51\varr{24} \\
& IxG & 0.61\varr{35} & 0.20\varr{0} & 0.33\varr{0} & 0.61\varr{36} & 0.20\varr{0} & 0.33\varr{0} & 0.61\varr{36} & 0.20\varr{0} & 0.33\varr{0} \\
& IG & 0.60\varr{34} & 0.20\textcolor{gray}{$+0\%$} & 0.34\textcolor{gray}{$+0\%$} & 0.61\varr{35} & 0.20\textcolor{gray}{$+0\%$} & 0.34\textcolor{gray}{$+0\%$} & 0.61\varr{36} & 0.20\textcolor{gray}{$+0\%$} & 0.34\textcolor{gray}{$+0\%$} \\
& LRP & 0.88\varr{48} & 0.72\varr{113} & 0.82\varr{64} & 0.88\varr{48} & 0.71\varr{112} & 0.82\varr{63} & 0.89\varr{48} & 0.71\varr{112} & 0.82\varr{63} \\
\midrule
\multirow{6}{*}{\rotatebox[origin=c]{90}{$t_2 = 100$}}& GradCam & 0.88\varr{22} & 0.66\varr{61} & 0.78\varr{37} & 0.88\varr{22} & 0.66\varr{62} & 0.78\varr{37} & 0.88\varr{22} & 0.66\varr{62} & 0.79\varr{37} \\
& GBP & 0.89\varr{109} & 0.28\varr{42} & 0.43\varr{32} & 0.89\varr{109} & 0.28\varr{42} & 0.43\varr{32} & 0.89\varr{109} & 0.28\varr{42} & 0.43\varr{32} \\
& Guide-GC & 0.92\varr{17} & 0.35\varr{32} & 0.51\varr{24} & 0.92\varr{17} & 0.35\varr{32} & 0.51\varr{24} & 0.92\varr{17} & 0.35\varr{33} & 0.51\varr{24} \\
& IxG & 0.62\varr{38} & 0.20\varr{0} & 0.33\varr{0} & 0.62\varr{38} & 0.20\varr{0} & 0.33\varr{0} & 0.62\varr{39} & 0.20\varr{0} & 0.33\varr{0} \\
& IG & 0.61\varr{36} & 0.20\textcolor{gray}{$+0\%$} & 0.34\textcolor{gray}{$+0\%$} & 0.62\varr{37} & 0.20\textcolor{gray}{$+0\%$} & 0.34\textcolor{gray}{$+0\%$} & 0.62\varr{38} & 0.20\textcolor{gray}{$+0\%$} & 0.34\textcolor{gray}{$+0\%$} \\
& LRP & 0.89\varr{49} & 0.72\varr{115} & 0.83\varr{65} & 0.89\varr{49} & 0.72\varr{115} & 0.82\varr{65} & 0.89\varr{50} & 0.72\varr{115} & 0.82\varr{65} \\
\midrule
\multirow{6}{*}{\rotatebox[origin=c]{90}{$t_2 = 500$}}& GradCam & 0.88\varr{21} & 0.66\varr{61} & 0.78\varr{37} & 0.88\varr{22} & 0.66\varr{61} & 0.78\varr{37} & 0.88\varr{22} & 0.66\varr{61} & 0.78\varr{37} \\
& GBP & 0.91\varr{114} & 0.28\varr{42} & 0.43\varr{33} & 0.91\varr{114} & 0.28\varr{42} & 0.43\varr{33} & 0.91\varr{114} & 0.28\varr{43} & 0.43\varr{33} \\
& Guide-GC & 0.93\varr{17} & 0.35\varr{32} & 0.51\varr{24} & 0.93\varr{17} & 0.35\varr{32} & 0.51\varr{24} & 0.93\varr{18} & 0.35\varr{33} & 0.51\varr{24} \\
& IxG & 0.62\varr{39} & 0.20\varr{0} & 0.34\varr{0} & 0.62\varr{39} & 0.20\varr{0} & 0.34\varr{0} & 0.63\varr{40} & 0.20\varr{0} & 0.34\varr{0} \\
& IG & 0.61\varr{36} & 0.21\varr{0} & 0.34\varr{0} & 0.62\varr{37} & 0.21\varr{0} & 0.34\varr{0} & 0.62\varr{38} & 0.21\varr{0} & 0.34\varr{0} \\
& LRP & 0.90\varr{50} & 0.73\varr{117} & 0.83\varr{65} & 0.90\varr{50} & 0.73\varr{117} & 0.83\varr{65} & 0.90\varr{51} & 0.73\varr{117} & 0.83\varr{65} \\

\bottomrule

\end{tabular}}
    \caption{\textit{Low variation between scaling parameters.} Across 11 different attribution methods considering convolutional and transformer based architectures, quantitative metrics measured using Region Attribution (RA), Intersection over Union (IoU), and F1 vary very slightly across various hyperparameter selections.}
    \label{table:wide_resnet502-abl}
\end{table*}

\begin{table*}[]
\centering

\resizebox{\textwidth}{!}{\begin{tabular}{cl@{\hspace{3em}}ccc@{\hspace{2em}}ccc@{\hspace{2em}}ccc}
& & \multicolumn{3}{c}{} & \multicolumn{3}{c}{DenseNet121} & \multicolumn{3}{c}{} \\
\toprule
& & \multicolumn{3}{c}{$t_1=2$} & \multicolumn{3}{c}{$t_1=5$} & \multicolumn{3}{c}{$t_1=10$} \\
\midrule
& Method & RA & IoU & F1 & RA & IoU & F1 & RA & IoU & F1\\
\cmidrule(lr){3-5} \cmidrule(lr){6-8}  \cmidrule(lr){9-11}

\multirow{6}{*}{\rotatebox[origin=c]{90}{$t_2 = 50$}}& GradCam & 0.59\varr{17} & 0.37\varr{7} & 0.49\textcolor{darkred}{$-1\%$} & 0.59\varr{17} & 0.37\varr{7} & 0.49\textcolor{darkred}{$-1\%$} & 0.59\varr{17} & 0.37\varr{8} & 0.49\textcolor{darkred}{$-1\%$} \\
& GBP & 0.83\varr{152} & 0.25\varr{27} & 0.39\varr{21} & 0.83\varr{152} & 0.25\varr{26} & 0.39\varr{21} & 0.83\varr{153} & 0.25\varr{26} & 0.39\varr{21} \\
& Guide-GC & 0.71\varr{27} & 0.26\varr{12} & 0.40\varr{9} & 0.71\varr{27} & 0.26\varr{12} & 0.40\varr{9} & 0.71\varr{27} & 0.26\varr{12} & 0.40\varr{9} \\
& IxG & 0.45\varr{28} & 0.20\varr{0} & 0.33\varr{0} & 0.46\varr{28} & 0.20\varr{0} & 0.33\varr{0} & 0.46\varr{29} & 0.20\varr{0} & 0.33\varr{0} \\
& IG & 0.49\varr{31} & 0.20\textcolor{gray}{$+0\%$} & 0.34\textcolor{gray}{$+0\%$} & 0.49\varr{32} & 0.20\textcolor{gray}{$+0\%$} & 0.34\textcolor{gray}{$+0\%$} & 0.49\varr{33} & 0.20\textcolor{gray}{$+0\%$} & 0.34\textcolor{gray}{$+0\%$} \\
& LRP & 0.44\varr{23} & 0.25\varr{0} & 0.40\varr{0} & 0.44\varr{24} & 0.25\varr{0} & 0.40\varr{0} & 0.45\varr{25} & 0.25\varr{0} & 0.40\varr{0} \\
\midrule
\multirow{6}{*}{\rotatebox[origin=c]{90}{$t_2 = 100$}}& GradCam & 0.60\varr{17} & 0.37\varr{6} & 0.48\textcolor{darkred}{$-2\%$} & 0.60\varr{17} & 0.37\varr{6} & 0.48\textcolor{darkred}{$-2\%$} & 0.60\varr{17} & 0.37\varr{6} & 0.49\textcolor{darkred}{$-2\%$} \\
& GBP & 0.84\varr{157} & 0.25\varr{27} & 0.40\varr{21} & 0.85\varr{158} & 0.25\varr{27} & 0.40\varr{21} & 0.85\varr{158} & 0.25\varr{27} & 0.40\varr{21} \\
& Guide-GC & 0.71\varr{28} & 0.26\varr{12} & 0.40\varr{9} & 0.71\varr{28} & 0.26\varr{12} & 0.40\varr{9} & 0.71\varr{28} & 0.26\varr{12} & 0.40\varr{9} \\
& IxG & 0.46\varr{31} & 0.20\varr{0} & 0.33\varr{0} & 0.46\varr{31} & 0.20\varr{0} & 0.33\varr{0} & 0.47\varr{32} & 0.20\varr{0} & 0.33\varr{0} \\
& IG & 0.50\varr{33} & 0.20\textcolor{gray}{$+0\%$} & 0.34\textcolor{gray}{$+0\%$} & 0.50\varr{34} & 0.20\textcolor{gray}{$+0\%$} & 0.34\textcolor{gray}{$+0\%$} & 0.50\varr{35} & 0.20\textcolor{gray}{$+0\%$} & 0.34\textcolor{gray}{$+0\%$} \\
& LRP & 0.44\varr{24} & 0.25\varr{0} & 0.40\varr{0} & 0.44\varr{24} & 0.25\varr{0} & 0.40\varr{0} & 0.45\varr{25} & 0.25\varr{0} & 0.40\varr{0} \\
\midrule
\multirow{6}{*}{\rotatebox[origin=c]{90}{$t_2 = 500$}}& GradCam & 0.58\varr{14} & 0.34\textcolor{gray}{$+0\%$} & 0.45\textcolor{darkred}{$-8\%$} & 0.58\varr{14} & 0.34\textcolor{gray}{$+0\%$} & 0.46\textcolor{darkred}{$-8\%$} & 0.58\varr{14} & 0.34\textcolor{gray}{$+0\%$} & 0.46\textcolor{darkred}{$-8\%$} \\
& GBP & 0.87\varr{165} & 0.25\varr{30} & 0.40\varr{23} & 0.87\varr{166} & 0.25\varr{30} & 0.40\varr{23} & 0.87\varr{166} & 0.25\varr{30} & 0.40\varr{23} \\
& Guide-GC & 0.74\varr{32} & 0.26\varr{12} & 0.40\varr{9} & 0.74\varr{32} & 0.26\varr{12} & 0.40\varr{9} & 0.74\varr{32} & 0.26\varr{12} & 0.40\varr{9} \\
& IxG & 0.47\varr{32} & 0.20\varr{0} & 0.33\varr{0} & 0.47\varr{33} & 0.20\varr{0} & 0.33\varr{0} & 0.47\varr{33} & 0.20\varr{0} & 0.33\varr{0} \\
& IG & 0.50\varr{34} & 0.20\textcolor{gray}{$+0\%$} & 0.34\textcolor{gray}{$+0\%$} & 0.50\varr{34} & 0.20\textcolor{gray}{$+0\%$} & 0.34\textcolor{gray}{$+0\%$} & 0.50\varr{35} & 0.20\textcolor{gray}{$+0\%$} & 0.34\textcolor{gray}{$+0\%$} \\
& LRP & 0.44\varr{23} & 0.25\varr{0} & 0.40\varr{0} & 0.44\varr{24} & 0.25\varr{0} & 0.40\varr{0} & 0.45\varr{25} & 0.25\varr{0} & 0.40\varr{0} \\

\bottomrule

\end{tabular}}
    \caption{\textit{Low variation between scaling parameters.} Across 6 different attribution methods considering convolutional and transformer based architectures, quantitative metrics measured using Region Attribution (RA), Intersection over Union (IoU), and F1 vary very slightly across various hyperparameter selections.}
    \label{table:densenet121-abl}
\end{table*}

\begin{table*}[]
\centering

\resizebox{\textwidth}{!}{\begin{tabular}{cl@{\hspace{3em}}ccc@{\hspace{2em}}ccc@{\hspace{2em}}ccc}
& & \multicolumn{3}{c}{} & \multicolumn{3}{c}{ConvNeXT} & \multicolumn{3}{c}{} \\
\toprule
& & \multicolumn{3}{c}{$t_1=2$} & \multicolumn{3}{c}{$t_1=5$} & \multicolumn{3}{c}{$t_1=10$} \\
\midrule
& Method & RA & IoU & F1 & RA & IoU & F1 & RA & IoU & F1\\
\cmidrule(lr){3-5} \cmidrule(lr){6-8}  \cmidrule(lr){9-11}

\multirow{5}{*}{\rotatebox[origin=c]{90}{$t_2 = 50$}}& GradCam & 0.96\varr{2} & 0.56\textcolor{darkred}{$-6\%$} & 0.70\textcolor{darkred}{$-5\%$} & 0.96\varr{2} & 0.56\textcolor{darkred}{$-6\%$} & 0.70\textcolor{darkred}{$-5\%$} & 0.96\varr{2} & 0.56\textcolor{darkred}{$-6\%$} & 0.70\textcolor{darkred}{$-5\%$} \\
& GBP & 0.50\varr{23} & 0.20\varr{0} & 0.33\varr{0} & 0.50\varr{24} & 0.20\varr{0} & 0.33\varr{0} & 0.51\varr{24} & 0.20\varr{0} & 0.33\varr{0} \\
& Guide-GC & 0.96\varr{1} & 0.35\varr{1} & 0.52\varr{1} & 0.96\varr{1} & 0.35\varr{1} & 0.52\varr{1} & 0.96\varr{1} & 0.35\varr{1} & 0.52\varr{1} \\
& IxG & 0.50\varr{24} & 0.20\varr{0} & 0.33\varr{0} & 0.50\varr{24} & 0.20\varr{0} & 0.33\varr{0} & 0.50\varr{24} & 0.20\varr{0} & 0.33\varr{0} \\
& IG & 0.62\varr{32} & 0.21\varr{0} & 0.34\varr{0} & 0.63\varr{33} & 0.21\varr{0} & 0.34\varr{0} & 0.63\varr{33} & 0.21\varr{0} & 0.34\varr{0} \\
\midrule
\multirow{5}{*}{\rotatebox[origin=c]{90}{$t_2 = 100$}}& GradCam & 0.96\varr{2} & 0.55\textcolor{darkred}{$-7\%$} & 0.70\textcolor{darkred}{$-6\%$} & 0.96\varr{2} & 0.55\textcolor{darkred}{$-7\%$} & 0.70\textcolor{darkred}{$-6\%$} & 0.96\varr{2} & 0.55\textcolor{darkred}{$-7\%$} & 0.70\textcolor{darkred}{$-6\%$} \\
& GBP & 0.51\varr{26} & 0.20\varr{0} & 0.33\varr{0} & 0.52\varr{26} & 0.20\varr{0} & 0.33\varr{0} & 0.52\varr{27} & 0.20\varr{0} & 0.33\varr{0} \\
& Guide-GC & 0.96\varr{1} & 0.35\varr{1} & 0.52\varr{1} & 0.96\varr{1} & 0.35\varr{1} & 0.52\varr{1} & 0.96\varr{1} & 0.35\varr{1} & 0.52\varr{1} \\
& IxG & 0.51\varr{26} & 0.20\varr{0} & 0.33\varr{0} & 0.51\varr{27} & 0.20\varr{0} & 0.33\varr{0} & 0.51\varr{27} & 0.20\varr{0} & 0.33\varr{0} \\
& IG & 0.64\varr{35} & 0.21\varr{0} & 0.34\varr{0} & 0.64\varr{35} & 0.21\varr{0} & 0.34\varr{0} & 0.64\varr{36} & 0.21\varr{0} & 0.34\varr{0} \\
\midrule
\multirow{5}{*}{\rotatebox[origin=c]{90}{$t_2 = 500$}}& GradCam & 0.96\varr{2} & 0.55\textcolor{darkred}{$-8\%$} & 0.69\textcolor{darkred}{$-7\%$} & 0.96\varr{2} & 0.55\textcolor{darkred}{$-8\%$} & 0.69\textcolor{darkred}{$-7\%$} & 0.96\varr{2} & 0.55\textcolor{darkred}{$-8\%$} & 0.69\textcolor{darkred}{$-7\%$} \\
& GBP & 0.53\varr{31} & 0.20\varr{0} & 0.33\varr{0} & 0.54\varr{31} & 0.20\varr{0} & 0.33\varr{0} & 0.54\varr{32} & 0.20\varr{0} & 0.33\varr{0} \\
& Guide-GC & 0.97\varr{1} & 0.35\varr{1} & 0.52\varr{1} & 0.97\varr{1} & 0.35\varr{1} & 0.52\varr{1} & 0.97\varr{1} & 0.35\varr{1} & 0.52\varr{1} \\
& IxG & 0.53\varr{31} & 0.20\varr{0} & 0.33\varr{0} & 0.53\varr{31} & 0.20\varr{0} & 0.33\varr{0} & 0.53\varr{31} & 0.20\varr{0} & 0.33\varr{0} \\
& IG & 0.64\varr{36} & 0.21\varr{0} & 0.34\varr{0} & 0.65\varr{37} & 0.21\varr{0} & 0.34\varr{0} & 0.65\varr{38} & 0.21\varr{0} & 0.34\varr{0} \\

\bottomrule

\end{tabular}}
    \caption{\textit{Low variation between scaling parameters.} Across 5 different attribution methods considering convolutional and transformer based architectures, quantitative metrics measured using Region Attribution (RA), Intersection over Union (IoU), and F1 vary very slightly across various hyperparameter selections.}
    \label{table:convnext-abl}
\end{table*}

\begin{table*}[]
\centering

\resizebox{\textwidth}{!}{\begin{tabular}{cl@{\hspace{3em}}ccc@{\hspace{2em}}ccc@{\hspace{2em}}ccc}
& & \multicolumn{3}{c}{} & \multicolumn{3}{c}{ViT-base-16} & \multicolumn{3}{c}{} \\
\toprule
& & \multicolumn{3}{c}{$t_1=2$} & \multicolumn{3}{c}{$t_1=5$} & \multicolumn{3}{c}{$t_1=10$} \\
\midrule
& Method & RA & IoU & F1 & RA & IoU & F1 & RA & IoU & F1\\
\cmidrule(lr){3-5} \cmidrule(lr){6-8}  \cmidrule(lr){9-11}

\multirow{6}{*}{\rotatebox[origin=c]{90}{$t_2 = 50$}}& Bi-attn & 0.94\varr{31} & 0.71\varr{180} & 0.82\varr{103} & 0.94\varr{31} & 0.71\varr{180} & 0.82\varr{103} & 0.94\varr{31} & 0.71\varr{180} & 0.82\varr{103} \\
& InFlow & 0.86\varr{21} & 0.56\varr{126} & 0.71\varr{78} & 0.86\varr{21} & 0.56\varr{126} & 0.71\varr{78} & 0.86\varr{21} & 0.56\varr{126} & 0.71\varr{78} \\
& Grad-Rollout & 0.72\varr{73} & 0.53\varr{112} & 0.68\varr{70} & 0.72\varr{73} & 0.53\varr{112} & 0.68\varr{70} & 0.72\varr{73} & 0.53\varr{112} & 0.68\varr{70} \\
& T-attn & 0.94\varr{32} & 0.71\varr{180} & 0.82\varr{103} & 0.93\varr{32} & 0.71\varr{180} & 0.82\varr{102} & 0.93\varr{32} & 0.71\varr{179} & 0.82\varr{102} \\
& LRP & 0.77\varr{35} & 0.51\varr{105} & 0.66\varr{65} & 0.77\varr{35} & 0.51\varr{105} & 0.66\varr{65} & 0.77\varr{35} & 0.51\varr{105} & 0.66\varr{65} \\
& Gradient & 0.93\varr{4} & 0.57\varr{3} & 0.70\varr{2} & 0.93\varr{4} & 0.57\varr{3} & 0.70\varr{2} & 0.93\varr{4} & 0.57\varr{3} & 0.70\varr{2} \\
\midrule
\multirow{6}{*}{\rotatebox[origin=c]{90}{$t_2 = 100$}}& Bi-attn & 0.94\varr{31} & 0.71\varr{180} & 0.82\varr{103} & 0.94\varr{31} & 0.71\varr{180} & 0.82\varr{103} & 0.94\varr{31} & 0.71\varr{180} & 0.82\varr{103} \\
& InFlow & 0.86\varr{21} & 0.56\varr{126} & 0.71\varr{78} & 0.86\varr{21} & 0.56\varr{126} & 0.71\varr{78} & 0.86\varr{21} & 0.56\varr{126} & 0.71\varr{78} \\
& Grad-Rollout & 0.73\varr{76} & 0.53\varr{113} & 0.68\varr{71} & 0.73\varr{76} & 0.53\varr{113} & 0.68\varr{71} & 0.73\varr{76} & 0.53\varr{113} & 0.68\varr{71} \\
& T-attn & 0.94\varr{32} & 0.71\varr{180} & 0.82\varr{103} & 0.93\varr{32} & 0.71\varr{180} & 0.82\varr{102} & 0.93\varr{32} & 0.71\varr{179} & 0.82\varr{102} \\
& LRP & 0.77\varr{35} & 0.51\varr{105} & 0.66\varr{65} & 0.77\varr{35} & 0.51\varr{105} & 0.66\varr{65} & 0.77\varr{35} & 0.51\varr{105} & 0.66\varr{65} \\
& Gradient & 0.93\varr{4} & 0.57\varr{3} & 0.70\varr{2} & 0.93\varr{4} & 0.57\varr{3} & 0.70\varr{2} & 0.93\varr{4} & 0.57\varr{3} & 0.70\varr{2} \\
\midrule
\multirow{6}{*}{\rotatebox[origin=c]{90}{$t_2 = 500$}}& Bi-attn & 0.94\varr{31} & 0.71\varr{181} & 0.82\varr{103} & 0.94\varr{32} & 0.71\varr{181} & 0.82\varr{103} & 0.94\varr{31} & 0.71\varr{181} & 0.82\varr{103} \\
& InFlow & 0.88\varr{23} & 0.57\varr{126} & 0.71\varr{78} & 0.88\varr{23} & 0.57\varr{126} & 0.71\varr{78} & 0.88\varr{23} & 0.57\varr{126} & 0.71\varr{78} \\
& Grad-Rollout & 0.78\varr{88} & 0.55\varr{118} & 0.69\varr{73} & 0.78\varr{88} & 0.54\varr{118} & 0.69\varr{73} & 0.78\varr{88} & 0.54\varr{118} & 0.69\varr{73} \\
& T-attn & 0.94\varr{32} & 0.71\varr{180} & 0.82\varr{103} & 0.93\varr{32} & 0.71\varr{180} & 0.82\varr{102} & 0.93\varr{32} & 0.71\varr{180} & 0.82\varr{102} \\
& LRP & 0.78\varr{36} & 0.51\varr{105} & 0.66\varr{66} & 0.78\varr{36} & 0.51\varr{105} & 0.66\varr{66} & 0.78\varr{36} & 0.51\varr{105} & 0.66\varr{66} \\
& Gradient & 0.93\varr{4} & 0.57\varr{3} & 0.70\varr{2} & 0.93\varr{4} & 0.57\varr{3} & 0.70\varr{2} & 0.93\varr{4} & 0.57\varr{3} & 0.70\varr{2} \\

\bottomrule

\end{tabular}}
    \caption{\textit{Low variation between scaling parameters.} Across 6 different attribution methods considering convolutional and transformer based architectures, quantitative metrics measured using Region Attribution (RA), Intersection over Union (IoU), and F1 vary very slightly across various hyperparameter selections.}
    \label{table:vit_timm-abl}
\end{table*}

\begin{table*}[]
\centering

\resizebox{\textwidth}{!}{\begin{tabular}{cl@{\hspace{3em}}ccc@{\hspace{2em}}ccc@{\hspace{2em}}ccc}
& & \multicolumn{3}{c}{} & \multicolumn{3}{c}{ViT-base-8} & \multicolumn{3}{c}{} \\
\toprule
& & \multicolumn{3}{c}{$t_1=2$} & \multicolumn{3}{c}{$t_1=5$} & \multicolumn{3}{c}{$t_1=10$} \\
\midrule
& Method & RA & IoU & F1 & RA & IoU & F1 & RA & IoU & F1\\
\cmidrule(lr){3-5} \cmidrule(lr){6-8}  \cmidrule(lr){9-11}

\multirow{6}{*}{\rotatebox[origin=c]{90}{$t_2 = 50$}}& Bi-attn & 0.90\varr{47} & 0.62\varr{147} & 0.75\varr{88} & 0.91\varr{47} & 0.62\varr{146} & 0.75\varr{88} & 0.91\varr{48} & 0.62\varr{146} & 0.75\varr{88} \\
& InFlow & 0.82\varr{18} & 0.47\varr{89} & 0.63\varr{58} & 0.82\varr{18} & 0.47\varr{89} & 0.63\varr{58} & 0.82\varr{18} & 0.47\varr{89} & 0.63\varr{58} \\
& Grad-Rollout & 0.71\varr{50} & 0.45\varr{80} & 0.61\varr{53} & 0.71\varr{50} & 0.45\varr{80} & 0.61\varr{53} & 0.71\varr{50} & 0.45\varr{80} & 0.61\varr{53} \\
& T-attn & 0.90\varr{53} & 0.63\varr{151} & 0.76\varr{90} & 0.90\varr{53} & 0.63\varr{151} & 0.76\varr{90} & 0.90\varr{53} & 0.63\varr{151} & 0.76\varr{90} \\
& LRP & 0.76\varr{25} & 0.42\varr{70} & 0.58\varr{46} & 0.76\varr{25} & 0.42\varr{70} & 0.58\varr{46} & 0.76\varr{25} & 0.42\varr{70} & 0.58\varr{46} \\
& Gradient & 0.90\varr{7} & 0.49\varr{8} & 0.64\varr{5} & 0.90\varr{7} & 0.49\varr{8} & 0.64\varr{5} & 0.90\varr{7} & 0.49\varr{8} & 0.64\varr{5} \\
\midrule
\multirow{6}{*}{\rotatebox[origin=c]{90}{$t_2 = 100$}}& Bi-attn & 0.91\varr{47} & 0.62\varr{149} & 0.76\varr{89} & 0.91\varr{48} & 0.62\varr{149} & 0.76\varr{89} & 0.91\varr{48} & 0.62\varr{149} & 0.76\varr{89} \\
& InFlow & 0.82\varr{18} & 0.47\varr{89} & 0.63\varr{58} & 0.82\varr{18} & 0.47\varr{89} & 0.63\varr{58} & 0.82\varr{18} & 0.47\varr{89} & 0.63\varr{58} \\
& Grad-Rollout & 0.71\varr{51} & 0.45\varr{80} & 0.61\varr{53} & 0.71\varr{51} & 0.45\varr{80} & 0.61\varr{53} & 0.71\varr{51} & 0.45\varr{80} & 0.61\varr{53} \\
& T-attn & 0.90\varr{54} & 0.63\varr{152} & 0.76\varr{90} & 0.90\varr{53} & 0.63\varr{152} & 0.76\varr{90} & 0.90\varr{53} & 0.63\varr{152} & 0.76\varr{90} \\
& LRP & 0.76\varr{25} & 0.42\varr{69} & 0.58\varr{46} & 0.76\varr{25} & 0.42\varr{69} & 0.58\varr{46} & 0.76\varr{25} & 0.42\varr{69} & 0.58\varr{46} \\
& Gradient & 0.90\varr{7} & 0.49\varr{7} & 0.64\varr{5} & 0.90\varr{7} & 0.49\varr{7} & 0.64\varr{5} & 0.90\varr{7} & 0.49\varr{7} & 0.64\varr{5} \\
\midrule
\multirow{6}{*}{\rotatebox[origin=c]{90}{$t_2 = 500$}}& Bi-attn & 0.91\varr{47} & 0.63\varr{152} & 0.76\varr{90} & 0.91\varr{48} & 0.63\varr{152} & 0.76\varr{90} & 0.91\varr{48} & 0.63\varr{152} & 0.76\varr{90} \\
& InFlow & 0.83\varr{20} & 0.47\varr{89} & 0.63\varr{58} & 0.83\varr{20} & 0.47\varr{89} & 0.63\varr{58} & 0.83\varr{20} & 0.47\varr{89} & 0.63\varr{58} \\
& Grad-Rollout & 0.74\varr{57} & 0.45\varr{81} & 0.61\varr{53} & 0.74\varr{57} & 0.45\varr{81} & 0.61\varr{53} & 0.74\varr{57} & 0.45\varr{81} & 0.61\varr{53} \\
& T-attn & 0.91\varr{54} & 0.63\varr{152} & 0.76\varr{90} & 0.90\varr{54} & 0.63\varr{152} & 0.76\varr{90} & 0.90\varr{53} & 0.63\varr{152} & 0.76\varr{90} \\
& LRP & 0.76\varr{25} & 0.42\varr{69} & 0.58\varr{46} & 0.76\varr{25} & 0.42\varr{69} & 0.58\varr{46} & 0.76\varr{25} & 0.42\varr{69} & 0.58\varr{46} \\
& Gradient & 0.90\varr{7} & 0.49\varr{7} & 0.64\varr{4} & 0.90\varr{7} & 0.49\varr{7} & 0.64\varr{5} & 0.90\varr{7} & 0.49\varr{7} & 0.64\varr{5} \\

\bottomrule

\end{tabular}}
    \caption{\textit{Low variation between scaling parameters.} Across 6 different attribution methods considering convolutional and transformer based architectures, quantitative metrics measured using Region Attribution (RA), Intersection over Union (IoU), and F1 vary very slightly across various hyperparameter selections.}
    \label{table:vit_timm_8-abl}
\end{table*}

\begin{table*}[]
\centering

\resizebox{\textwidth}{!}{\begin{tabular}{cl@{\hspace{3em}}ccc@{\hspace{2em}}ccc@{\hspace{2em}}ccc}
& & \multicolumn{3}{c}{} & \multicolumn{3}{c}{ViT-base-32} & \multicolumn{3}{c}{} \\
\toprule
& & \multicolumn{3}{c}{$t_1=2$} & \multicolumn{3}{c}{$t_1=5$} & \multicolumn{3}{c}{$t_1=10$} \\
\midrule
& Method & RA & IoU & F1 & RA & IoU & F1 & RA & IoU & F1\\
\cmidrule(lr){3-5} \cmidrule(lr){6-8}  \cmidrule(lr){9-11}

\multirow{6}{*}{\rotatebox[origin=c]{90}{$t_2 = 50$}}& Bi-attn & 0.86\varr{71} & 0.62\varr{149} & 0.75\varr{87} & 0.85\varr{71} & 0.62\varr{149} & 0.75\varr{87} & 0.85\varr{70} & 0.62\varr{148} & 0.75\varr{87} \\
& InFlow & 0.77\varr{20} & 0.56\varr{124} & 0.70\varr{75} & 0.77\varr{20} & 0.56\varr{124} & 0.70\varr{75} & 0.77\varr{20} & 0.56\varr{124} & 0.70\varr{75} \\
& Grad-Rollout & 0.64\varr{84} & 0.51\varr{103} & 0.66\varr{64} & 0.64\varr{84} & 0.51\varr{102} & 0.66\varr{64} & 0.64\varr{84} & 0.51\varr{102} & 0.66\varr{64} \\
& T-attn & 0.84\varr{70} & 0.62\varr{146} & 0.74\varr{86} & 0.84\varr{70} & 0.62\varr{146} & 0.74\varr{86} & 0.84\varr{70} & 0.62\varr{146} & 0.74\varr{86} \\
& LRP & 0.66\varr{49} & 0.46\varr{85} & 0.61\varr{53} & 0.84\varr{70} & 0.62\varr{146} & 0.74\varr{86} & 0.84\varr{70} & 0.62\varr{146} & 0.74\varr{86} \\
& Gradient & 0.79\varr{19} & 0.51\varr{19} & 0.65\varr{12} & 0.66\varr{49} & 0.46\varr{85} & 0.61\varr{53} & 0.66\varr{49} & 0.46\varr{85} & 0.61\varr{53} \\
\midrule
\multirow{6}{*}{\rotatebox[origin=c]{90}{$t_2 = 100$}}& Bi-attn & 0.86\varr{71} & 0.62\varr{149} & 0.75\varr{87} & 0.86\varr{71} & 0.62\varr{149} & 0.75\varr{87} & 0.85\varr{71} & 0.62\varr{148} & 0.75\varr{87} \\
& InFlow & 0.78\varr{21} & 0.56\varr{124} & 0.70\varr{75} & 0.78\varr{21} & 0.56\varr{124} & 0.70\varr{75} & 0.78\varr{21} & 0.56\varr{124} & 0.70\varr{75} \\
& Grad-Rollout & 0.66\varr{91} & 0.51\varr{106} & 0.66\varr{66} & 0.66\varr{91} & 0.51\varr{106} & 0.66\varr{66} & 0.66\varr{91} & 0.51\varr{106} & 0.66\varr{66} \\
& T-attn & 0.84\varr{70} & 0.62\varr{146} & 0.74\varr{86} & 0.84\varr{70} & 0.62\varr{146} & 0.74\varr{86} & 0.84\varr{70} & 0.62\varr{146} & 0.74\varr{86} \\
& LRP & 0.66\varr{49} & 0.46\varr{85} & 0.61\varr{53} & 0.66\varr{49} & 0.46\varr{85} & 0.61\varr{53} & 0.66\varr{49} & 0.46\varr{85} & 0.61\varr{53} \\
& Gradient & 0.79\varr{19} & 0.51\varr{19} & 0.65\varr{12} & 0.79\varr{19} & 0.51\varr{19} & 0.65\varr{12} & 0.79\varr{19} & 0.51\varr{19} & 0.65\varr{12} \\
\midrule
\multirow{6}{*}{\rotatebox[origin=c]{90}{$t_2 = 500$}}& Bi-attn & 0.86\varr{71} & 0.62\varr{149} & 0.75\varr{87} & 0.86\varr{71} & 0.62\varr{149} & 0.75\varr{87} & 0.85\varr{71} & 0.62\varr{149} & 0.75\varr{87} \\
& InFlow & 0.80\varr{24} & 0.57\varr{126} & 0.70\varr{76} & 0.80\varr{24} & 0.57\varr{126} & 0.70\varr{76} & 0.80\varr{24} & 0.57\varr{126} & 0.70\varr{76} \\
& Grad-Rollout & 0.72\varr{110} & 0.54\varr{116} & 0.68\varr{71} & 0.72\varr{110} & 0.54\varr{116} & 0.68\varr{71} & 0.72\varr{110} & 0.54\varr{116} & 0.68\varr{71} \\
& T-attn & 0.84\varr{70} & 0.62\varr{146} & 0.74\varr{86} & 0.84\varr{70} & 0.62\varr{146} & 0.74\varr{86} & 0.84\varr{70} & 0.62\varr{146} & 0.74\varr{86} \\
& LRP & 0.68\varr{53} & 0.47\varr{86} & 0.61\varr{53} & 0.68\varr{53} & 0.47\varr{86} & 0.61\varr{53} & 0.68\varr{53} & 0.47\varr{86} & 0.61\varr{53} \\
& Gradient & 0.79\varr{19} & 0.51\varr{18} & 0.65\varr{11} & 0.79\varr{19} & 0.51\varr{18} & 0.65\varr{11} & 0.79\varr{19} & 0.51\varr{19} & 0.65\varr{11} \\

\bottomrule

\end{tabular}}
    \caption{\textit{Low variation between scaling parameters.} Across 6 different attribution methods considering convolutional and transformer based architectures, quantitative metrics measured using Region Attribution (RA), Intersection over Union (IoU), and F1 vary very slightly across various hyperparameter selections.}
    \label{table:vit_timm_32-abl}
\end{table*}

\subsection{Deletion test}

We report the results for the deletion test in Table ~\ref{tab:deletion}. In contrast to insertion tests a lower AUC is better. For CNNs, adding \ourmethod consistently reduces AUC (often $9$–$15\%$ on average for ResNet/WideResNet/DenseNet), i.e., removing the top-ranked pixels identified by \ourmethod drops the target probability faster and thus better targets class-critical evidence. The main exception is Grad-CAM (and ConvNeXT overall), where coarse, high-coverage maps can behave like near “one-shot” masks in deletion, leaving little room for refinement and sometimes worsening AUC when refinement becomes more selective. For ViTs, gains are limited and sometimes slightly negative on average: many transformer attributions are relatively diffuse, so deleting their top-ranked regions removes large image areas and can look strong under deletion, whereas \ourmethod tends to denoise/localize and therefore deletes less context early.

Deletion differs from insertion because confidence can fall not only when truly class-relevant evidence is removed, but also when unrelated yet supportive context (or general image structure) is destroyed; consequently, large or blurry masks may score well by broadly degrading the input rather than precisely isolating discriminative cues.

\begin{table*}
  \centering
  \caption{\textit{Improving transformer attributions on deletion test.}  Augmenting the base method with \ourmethod improves the AUC (lower is better) for insertion tests for convolutional architectures by $0$-$14\%$. \gradcam is a again an outlier since it almost deletes the image in one go. For ViTs modestly worsens the AUC, similar to \gradcam attribution methods for ViTs are often very diffuse and large, hence deleting a big part of the image will yield strong results. }\label{table:exp-insertion-supp}%
\begin{subtable}{0.55\linewidth}
  \centering
    \subcaption{CNN-based architectures}%
  \resizebox{0.9\linewidth}{!}{\begin{tabular}{l rrrr}
    \toprule
    Method & ResNet50  & WideResNet50-2 & DenseNet121 & ConvNeXT \\
    \midrule
        IG & 0.06\negvarr{25} & 0.06\negvarr{14} & 0.07\negvarr{22} & 0.10\negvarr{9} \\
        GBP & 0.04\negvarr{20} & 0.04\negvarr{20} & 0.04\negvarr{33} & 0.13\negvarr{13} \\
        IxG & 0.09\negvarr{18} & 0.08\negvarr{20} & 0.10\negvarr{17} & 0.13\negvarr{13} \\
        Guide-GC & 0.03\negvarr{25} & 0.04\textcolor{gray}{${+0\%}$} & 0.05\textcolor{gray}{${+0\%}$} & 0.07\textcolor{gray}{${+0\%}$} \\
        GradCam & 0.04\textcolor{gray}{${+0\%}$} & 0.05\textcolor{gray}{${+0\%}$} & 0.07\textcolor{gray}{${+0\%}$} & 0.08\textcolor{darkred}{${+33\%}$} \\
        LRP & 0.04\textcolor{gray}{${+0\%}$} & 0.04\textcolor{gray}{${+0\%}$} & 0.11\textcolor{gray}{${+0\%}$} & 0.00\textcolor{gray}{$+0\%$} \\
        \cmidrule[0.1pt](lr){2-5}
        Avg improvement & \negvarr{14.67} & \negvarr{9.00} & \negvarr{12.00} & \negvarr{0.33} \\
    \bottomrule
  \end{tabular}}
\end{subtable}\hfill%
\begin{subtable}{0.4\linewidth}
  \centering
  \small
    \subcaption{Transformer-based architectures}%
  \resizebox{\linewidth}{!}{\begin{tabular}{lrrr}
    \toprule
    Method & ViT-base-8 & ViT-base-16 & ViT-base-32 \\
    \midrule
        Bi-attn & 0.06\textcolor{gray}{${+0\%}$} & 0.05\textcolor{gray}{${+0\%}$} & 0.04\textcolor{gray}{${+0\%}$} \\
        T-attn & 0.07\textcolor{gray}{${+0\%}$} & 0.28\textcolor{darkred}{${+22\%}$} & 0.05\textcolor{darkred}{${+25\%}$} \\
        InFlow & 0.06\textcolor{gray}{${+0\%}$} & 0.28\textcolor{darkred}{${+22\%}$} & 0.04\textcolor{gray}{${+0\%}$} \\
        Gradient & 0.07\textcolor{gray}{${+0\%}$} & 0.06\textcolor{gray}{${+0\%}$} & 0.05\textcolor{darkred}{${+25\%}$} \\
        Grad-Rl & 0.07\textcolor{gray}{${+0\%}$} & 0.06\negvarr{14} & 0.04\negvarr{20} \\
        LRP & 0.00\textcolor{gray}{$+0\%$} & 0.00\textcolor{gray}{$+0\%$} & 0.00\textcolor{gray}{$+0\%$} \\
        \cmidrule[0.1pt](lr){2-4}
        Avg improvement & \textcolor{gray}{$0.00\%$} & \textcolor{darkred}{${+5.00\%}$} & \textcolor{darkred}{${+5.00\%}$} \\
    \bottomrule
  \end{tabular}}
\end{subtable}

  \label{tab:deletion}
\end{table*} 

\label{app:insdel}
\begin{figure}
    \centering
    \includegraphics[width=0.4\linewidth]{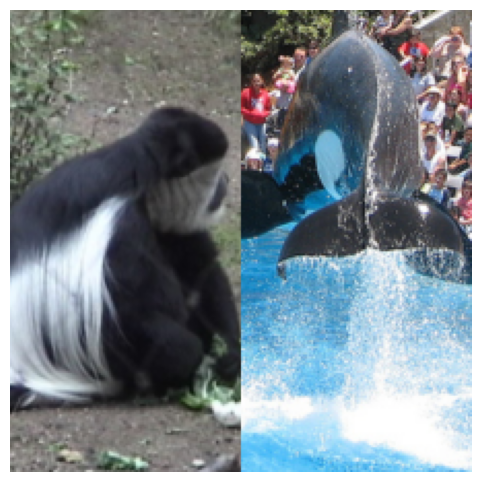}
    \caption{We show an example of how the combined images look. We can see that both objects are clearly visible and identifiable.}
    \label{fig:insdel-ex}
\end{figure}

\subsection{Sanity Checks}
We show the sanity check plots for these additional architectures in Figure \ref{fig:sanity-metrics-full-resnet}-\ref{fig:sanity-metrics-full-vit32}.

\subsection{Ablations}

\begin{figure*}
\resizebox{\textwidth}{!}{
	\begin{tikzpicture}
		\def\imgVZX{0}
		\def\imgVZY{0}
		\node (label_imgVZ) at (\imgVZX, \imgVZY) {\includegraphics[width=5cm, trim=0 0 0 {4em}, clip]{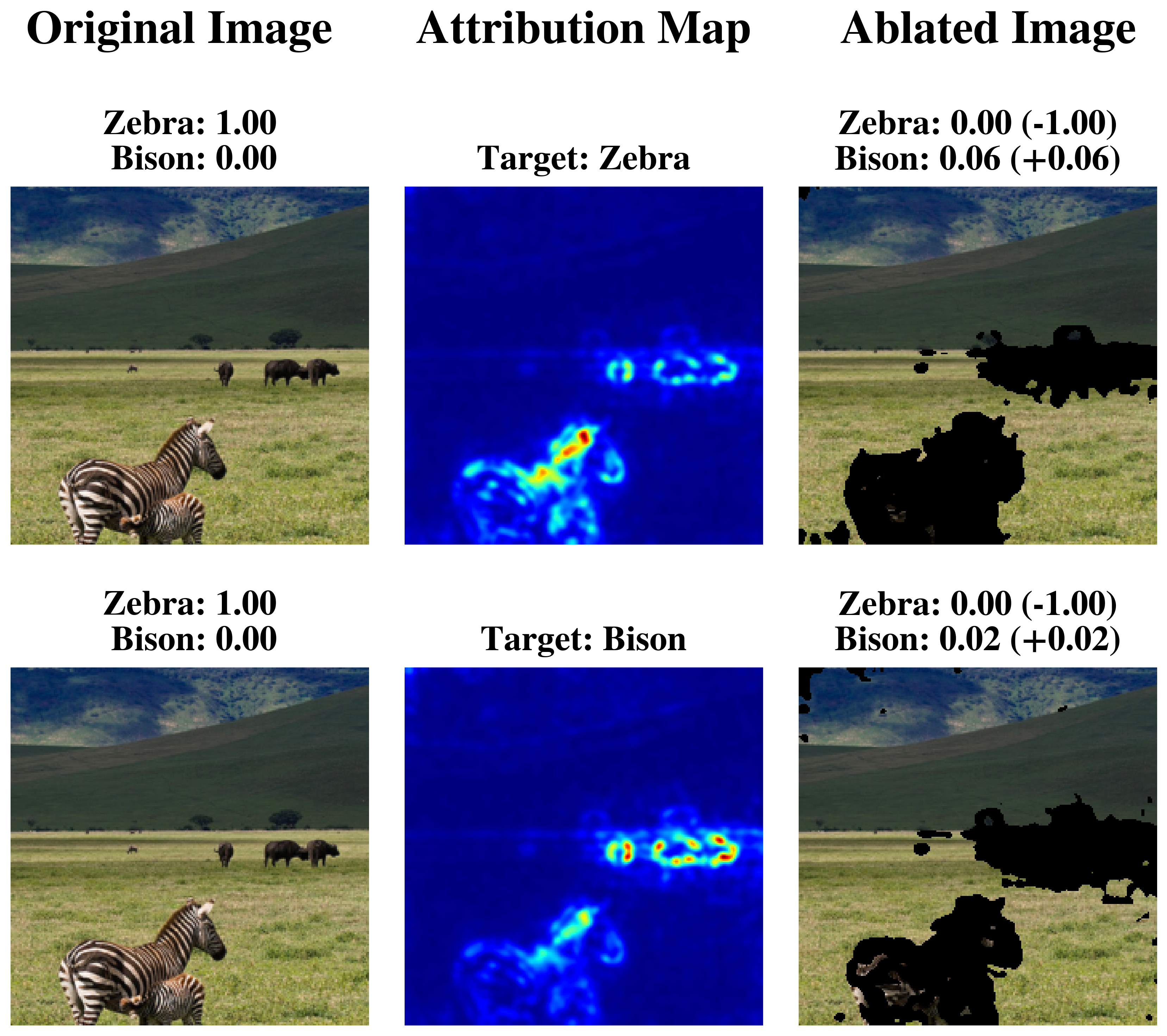}};

		\def\imgVPX{5.35}
		\def\imgVPY{0}
		\node (label_imgVP) at (\imgVPX, \imgVPY) {\includegraphics[width=5cm, trim=0 0 0 {4em}, clip]{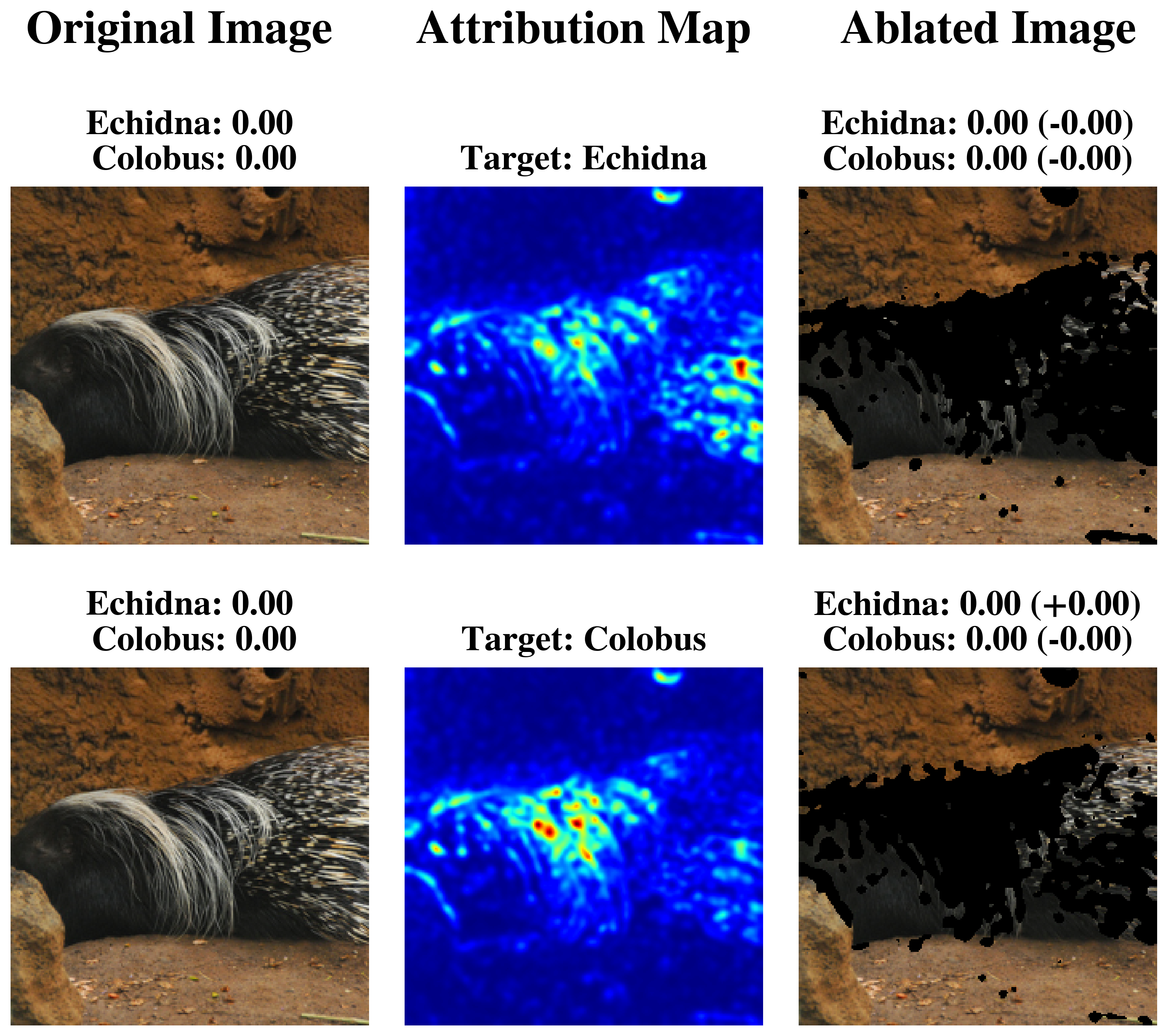}};
		
		\def\imgVCX{10.7}
		\def\imgVCY{0}
		\node (label_imgVC) at (\imgVCX, \imgVCY) {\includegraphics[width=5cm, trim=0 0 0 {4em}, clip]{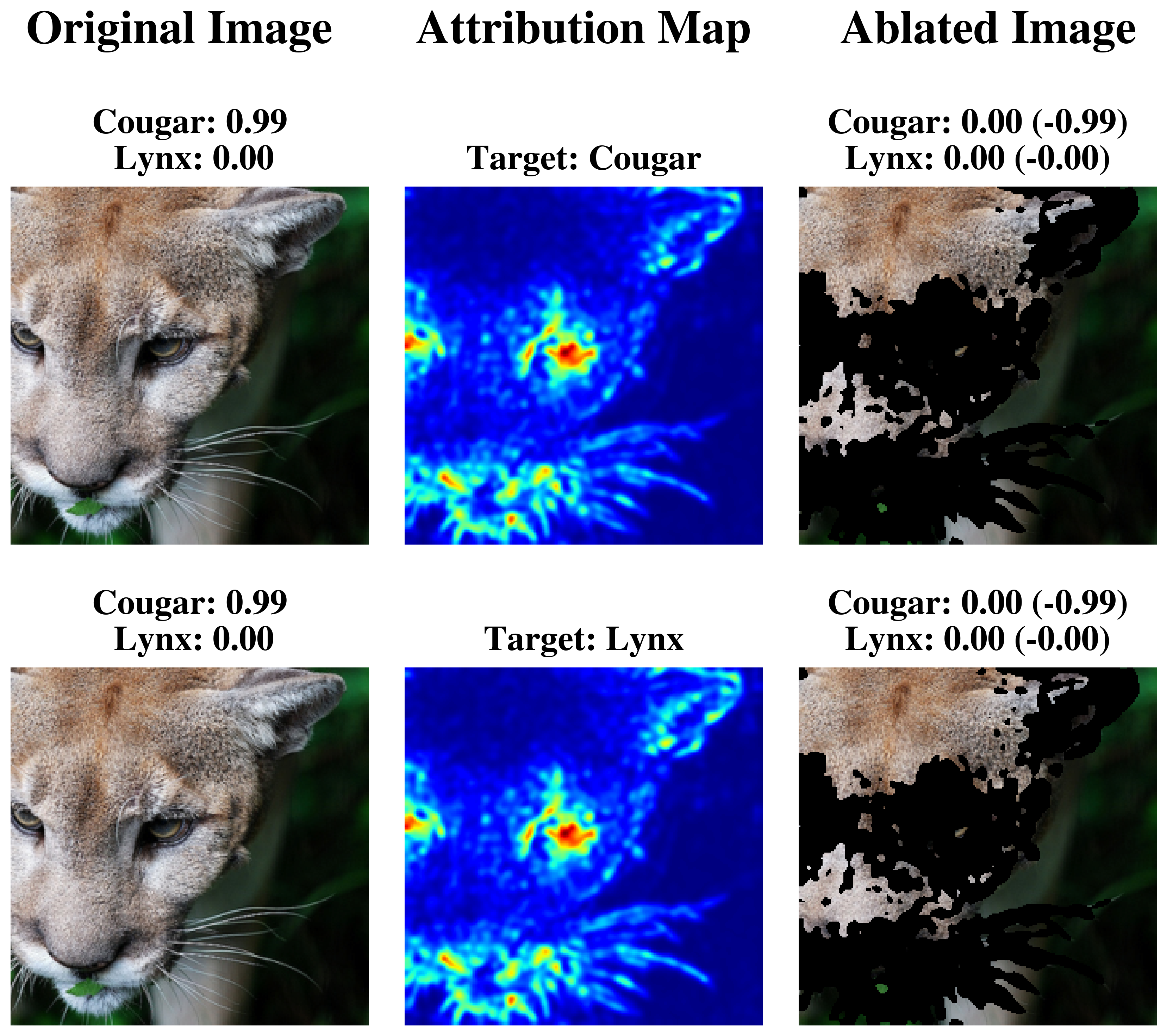}};

		\def\imgCZX{0}
		\def\imgCZY{-4.5}
		\node (label_imgCZ) at (\imgCZX, \imgCZY) {\includegraphics[width=5cm, trim=0 0 0 {5em}, clip]{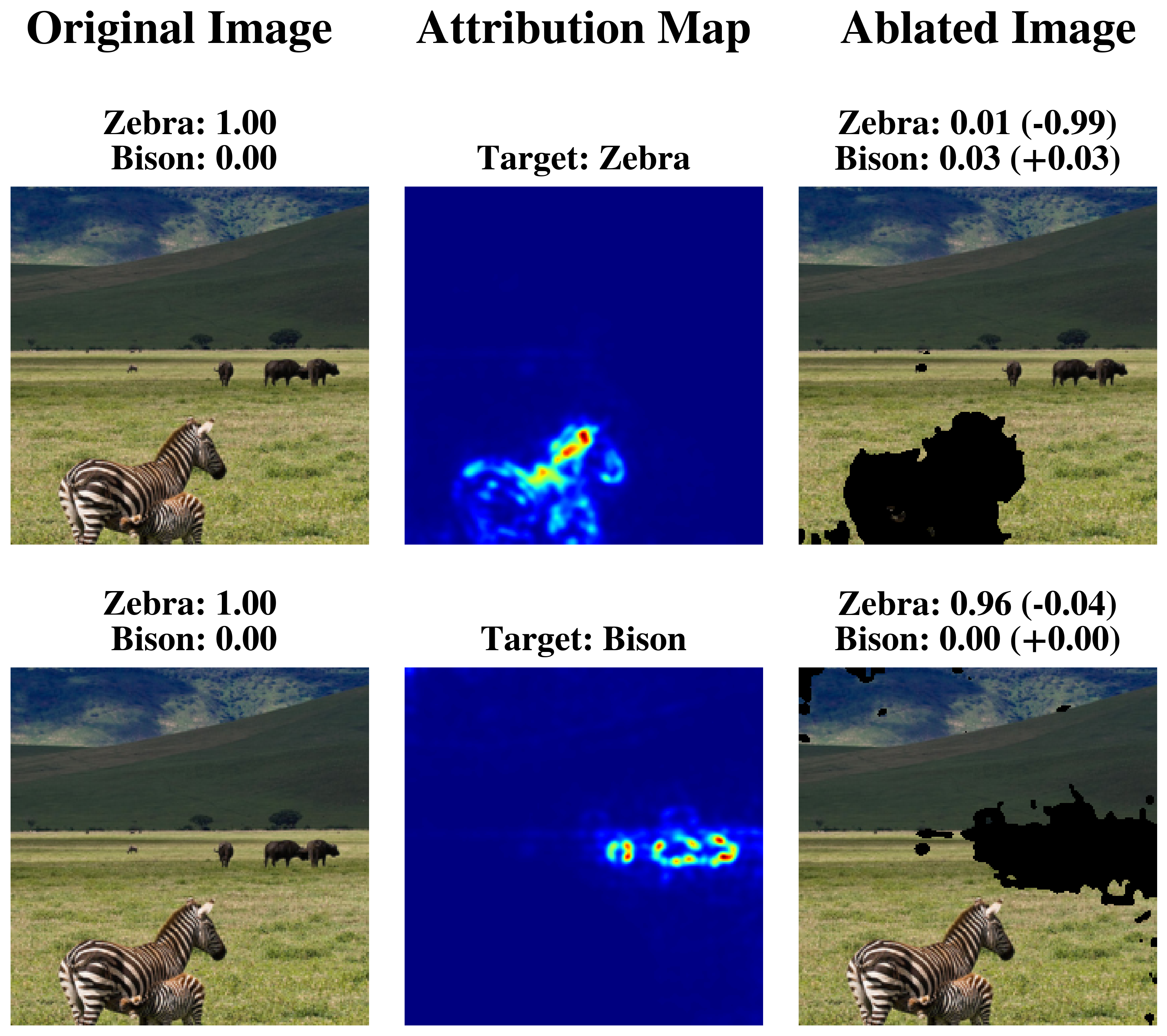}};

		\def\imgCPX{5.35}
		\def\imgCPY{-4.5}
		\node (label_imgCP) at (\imgCPX, \imgCPY) {\includegraphics[width=5cm, trim=0 0 0 {5em}, clip]{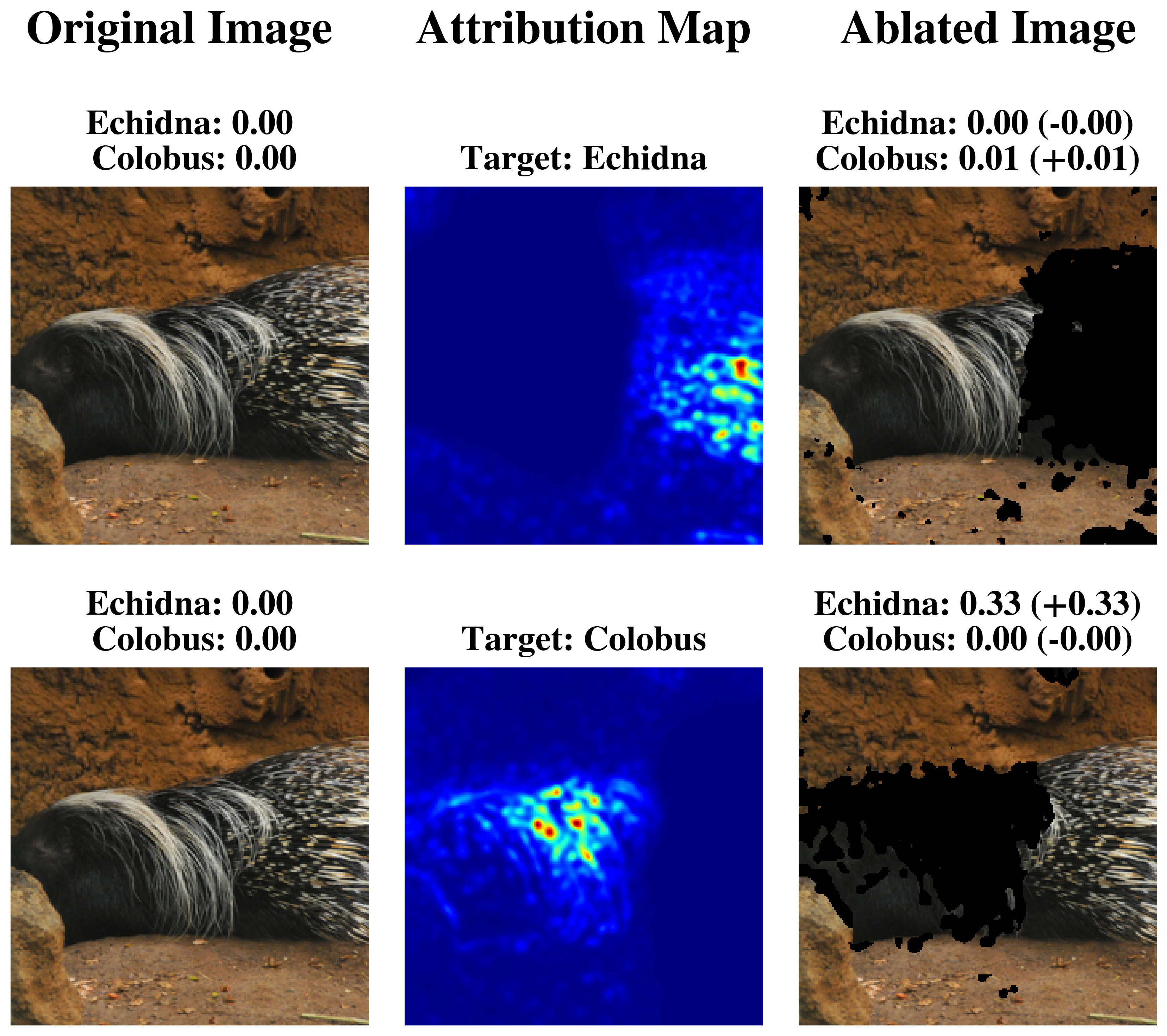}};
		
		\def\imgCCX{10.7}
		\def\imgCCY{-4.5}
		\node (label_imgCC) at (\imgCCX, \imgCCY) {\includegraphics[width=5cm, trim=0 0 0 {5em}, clip]{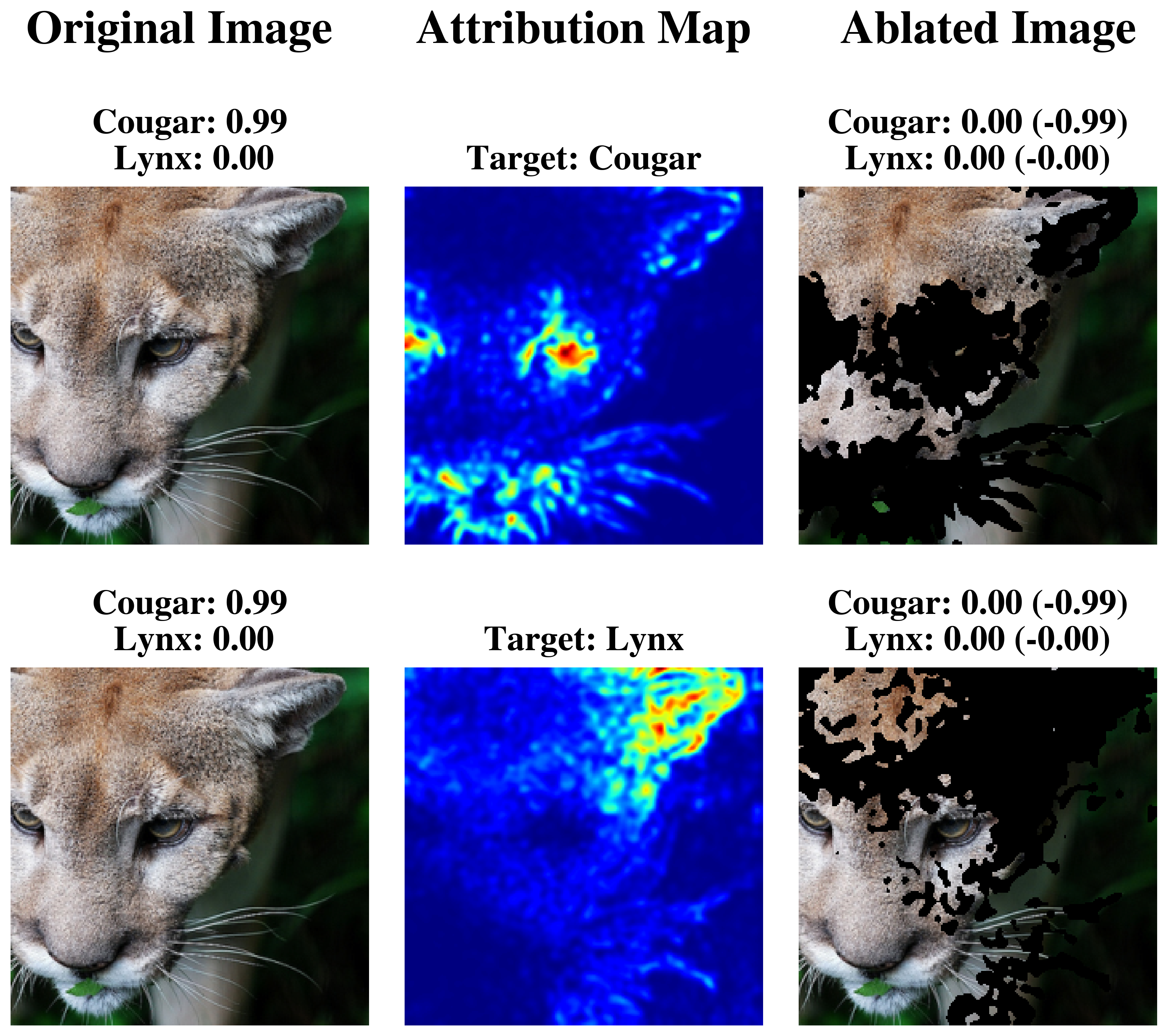}};

		\def\gbpX{-2.85}
		\def\gbpY{-0.25}
		\node [rotate=90] (label_gbp) at (\gbpX, \gbpY) {\small Guided Backprop};
		
		\def\gbpX{-2.85}
		\def\gbpY{-4.65}
		\node [rotate=90] (label_gbpc) at (\gbpX, \gbpY) {\small GBP + \ourmethod};		

		\node (label_oiz) at (-1.7, 2) {\notsotiny Original Image};		
		\node (label_amz) at (0, 2) {\notsotiny Attribution Map};		
		\node (label_abz) at (1.7, 2) {\notsotiny Ablated Image};		

		\node (label_oip) at (5.35-1.7, 2) {\notsotiny Original Image};		
		\node (label_amp) at (5.35+0, 2) {\notsotiny Attribution Map};		
		\node (label_abp) at (5.35+1.7, 2) {\notsotiny Ablated Image};		

		\node (label_oic) at (10.7-1.7, 2) {\notsotiny Original Image};		
		\node (label_amc) at (10.7+0, 2) {\notsotiny Attribution Map};		
		\node (label_abc) at (10.7+1.7, 2) {\notsotiny Ablated Image};
	\end{tikzpicture}}        
    \caption{\textit{Qualitative example of the ablation study.} For GBP (top) and GBP with \ourmethod (bottom) we provide examples from the insertion/deletion ablation. For each, we show the original image with class softmax scores for two classes, the attribution map for each of the classes, and the attribution-based intervention mask on each of the classes with resulting changes in class softmax scores.}
    \label{fig:ablation-app}
\end{figure*}

We show more examples of the ablation in Figure \ref{fig:ablation-app}.

\begin{figure*}[h!]
	\begin{tikzpicture}
    		\pgfplotscreateplotcyclelist{prcl-iccv-interleaved}{
			{dollarbill, mark=*, dotted}, {darkblue, mark=*, dotted},{mediumpurple, mark=*, dotted},
			{dollarbill, mark=*, solid},{darkblue, mark=*, solid},{mediumpurple, mark=*, solid},
			{mikadoyellow, mark=*, dotted}, {orange, mark=*, dotted},{pink, mark=*, dotted},
			{mikadoyellow, mark=*, solid},
			{orange, mark=*, solid},
            {pink, mark=*, solid},
		}
		\begin{groupplot}[
			pretty line,
			pretty ygrid,
			cycle list name = prcl-iccv-interleaved, 
			group style={group size=3 by 1, horizontal sep=5em},
			title style = {font=\footnotesize},
			width           = 0.33\textwidth,
			height          = 3cm,
			xlabel 			= {Layers Randomized (\%)},
			ylabel 			= {},		
			label style 	= {font = \scriptsize},
            xtick = {0,10,...,100},
            xticklabels = {0,,20,,40,,60,,80,,100},
			ymax            = 1.0,
			ymin            = 0.0,
			legend style    = {at = {(-2.1,-1.1)}, anchor = north west, nodes={scale=0.85}, draw=none, {/tikz/every even column/.append style={column sep=0.5cm}}},
			legend columns  = 3,
			]
			\nextgroupplot[title={Cosine}, mark options = {scale=0.5}, ylabel = {Similarity Score}]
			\foreach \ycontrast in {original,contrastive} {
				\foreach \ybase in {guided_backprop, integrated_gradients, input_x_gradient}{
					\addplot+[] table[x=x_axis, 	y={\ybase_\ycontrast_cosine}, col sep=comma] {data/rando/resnet50_inserted.csv};
			}}
        	\foreach \ycontrast in {original,contrastive} {
				\foreach \ybase in {gradcam, guided_grad_cam, lrp}{
					\addplot+[] table[x=x_axis, 		y={\ybase_\ycontrast_cosine}, col sep=comma] {data/rando/resnet50_inserted.csv};
			}}
			\nextgroupplot[title={Spearman}, mark options = {scale=0.5}]
			\foreach \ycontrast in {original,contrastive} {
				\foreach \ybase in {guided_backprop, integrated_gradients, input_x_gradient}{
					\addplot+[] table[x=x_axis, 		y={\ybase_\ycontrast_spearman}, col sep=comma] {data/rando/resnet50_inserted.csv};
			}}
        	\foreach \ycontrast in {original,contrastive} {
				\foreach \ybase in {gradcam, guided_grad_cam, lrp}{
					\addplot+[] table[x=x_axis, 	 	y={\ybase_\ycontrast_spearman}, col sep=comma] {data/rando/resnet50_inserted.csv};
			}}
			\nextgroupplot[title={Pearson}, mark options = {scale=0.5}]
			\foreach \ycontrast in {original,contrastive} {
				\foreach \ybase in {guided_backprop, integrated_gradients, input_x_gradient}{
					\addplot+[] table[x=x_axis, 		y={\ybase_\ycontrast_pearson}, col sep=comma] {data/rando/resnet50_inserted.csv};
			}}
			\foreach \ycontrast in {original,contrastive} {
				\foreach \ybase in {gradcam, guided_grad_cam, lrp}{
					\addplot+[] table[x=x_axis, 	 	y={\ybase_\ycontrast_pearson}, col sep=comma] {data/rando/resnet50_inserted.csv};
			}}
			\legend{Guided Backprop, Integrated Gradients, Input$\times$Gradients,
                    Guided Backprop + \ourmethod, Integrated Gradients + \ourmethod, Input$\times$Gradients + \ourmethod,
                GradCAM, Guided GradCAM, LRP,
				  GradCAM + \ourmethod, Guided GradCAM + \ourmethod, LRP + \ourmethod}				 			
		\end{groupplot}
	\end{tikzpicture}%
	\caption{ResNet50: \ourmethod improves all base methods under randomization [Lower is better]. For all methods and for varying level of randomization, we measure the similarity between the attention map for the unperturbed network and the randomized network. Dashed lines are base methods, solid lines when augmenting with \ourmethod, which improve the corresponding baseline method.}
	\label{fig:sanity-metrics-full-resnet}
\end{figure*}
\begin{figure*}[h!]
	\begin{tikzpicture}
    		\pgfplotscreateplotcyclelist{prcl-iccv-interleaved}{
			{dollarbill, mark=*, dotted}, {darkblue, mark=*, dotted},{mediumpurple, mark=*, dotted},
			{dollarbill, mark=*, solid},{darkblue, mark=*, solid},{mediumpurple, mark=*, solid},
			{mikadoyellow, mark=*, dotted}, {orange, mark=*, dotted},{pink, mark=*, dotted},
			{mikadoyellow, mark=*, solid},
			{orange, mark=*, solid},
            {pink, mark=*, solid},
		}
		\begin{groupplot}[
			pretty line,
			pretty ygrid,
			cycle list name = prcl-iccv-interleaved, 
			group style={group size=3 by 1, horizontal sep=5em},
			title style = {font=\footnotesize},
			width           = 0.33\textwidth,
			height          = 3cm,
			xlabel 			= {Layers Randomized (\%)},
			ylabel 			= {},		
			label style 	= {font = \scriptsize},
            xtick = {0,10,...,100},
            xticklabels = {0,,20,,40,,60,,80,,100},
			ymax            = 1.0,
			ymin            = 0.0,
			legend style    = {at = {(-2.1,-1.1)}, anchor = north west, nodes={scale=0.85}, draw=none, {/tikz/every even column/.append style={column sep=0.5cm}}},
			legend columns  = 3,
			]
			\nextgroupplot[title={Cosine}, mark options = {scale=0.5}, ylabel = {Similarity Score}]
			\foreach \ycontrast in {original,contrastive} {
				\foreach \ybase in {guided_backprop, integrated_gradients, input_x_gradient}{
					\addplot+[] table[x=x_axis, 	y={\ybase_\ycontrast_cosine}, col sep=comma] {data/rando/densenet121_inserted.csv};
			}}
        	\foreach \ycontrast in {original,contrastive} {
				\foreach \ybase in {gradcam, guided_grad_cam, lrp}{
					\addplot+[] table[x=x_axis, 		y={\ybase_\ycontrast_cosine}, col sep=comma] {data/rando/densenet121_inserted.csv};
			}}
			\nextgroupplot[title={Spearman}, mark options = {scale=0.5}]
			\foreach \ycontrast in {original,contrastive} {
				\foreach \ybase in {guided_backprop, integrated_gradients, input_x_gradient}{
					\addplot+[] table[x=x_axis, 		y={\ybase_\ycontrast_spearman}, col sep=comma] {data/rando/densenet121_inserted.csv};
			}}
        	\foreach \ycontrast in {original,contrastive} {
				\foreach \ybase in {gradcam, guided_grad_cam, lrp}{
					\addplot+[] table[x=x_axis, 	 	y={\ybase_\ycontrast_spearman}, col sep=comma] {data/rando/densenet121_inserted.csv};
			}}
			\nextgroupplot[title={Pearson}, mark options = {scale=0.5}]
			\foreach \ycontrast in {original,contrastive} {
				\foreach \ybase in {guided_backprop, integrated_gradients, input_x_gradient}{
					\addplot+[] table[x=x_axis, 		y={\ybase_\ycontrast_pearson}, col sep=comma] {data/rando/densenet121_inserted.csv};
			}}
			\foreach \ycontrast in {original,contrastive} {
				\foreach \ybase in {gradcam, guided_grad_cam, lrp}{
					\addplot+[] table[x=x_axis, 	 	y={\ybase_\ycontrast_pearson}, col sep=comma] {data/rando/densenet121_inserted.csv};
			}}
			\legend{Guided Backprop, Integrated Gradients, Input$\times$Gradients,
                    Guided Backprop + \ourmethod, Integrated Gradients + \ourmethod, Input$\times$Gradients + \ourmethod,
                GradCAM, Guided GradCAM, LRP,
				  GradCAM + \ourmethod, Guided GradCAM + \ourmethod, LRP + \ourmethod}				 			
		\end{groupplot}
	\end{tikzpicture}%
	\caption{DenseNet121: \ourmethod improves all base methods under randomization [Lower is better]. For all methods and for varying level of randomization, we measure the similarity between the attention map for the unperturbed network and the randomized network. Dashed lines are base methods, solid lines when augmenting with \ourmethod, which improve the corresponding baseline method.}
	\label{fig:sanity-metrics-full-densnet}
\end{figure*}
\begin{figure*}[h!]
	\begin{tikzpicture}
    		\pgfplotscreateplotcyclelist{prcl-iccv-interleaved}{
			{dollarbill, mark=*, dotted}, {darkblue, mark=*, dotted},{mediumpurple, mark=*, dotted},
			{dollarbill, mark=*, solid},{darkblue, mark=*, solid},{mediumpurple, mark=*, solid},
			{mikadoyellow, mark=*, dotted}, {orange, mark=*, dotted},{pink, mark=*, dotted},
			{mikadoyellow, mark=*, solid},
			{orange, mark=*, solid},
            {pink, mark=*, solid},
		}
		\begin{groupplot}[
			pretty line,
			pretty ygrid,
			cycle list name = prcl-iccv-interleaved, 
			group style={group size=3 by 1, horizontal sep=5em},
			title style = {font=\footnotesize},
			width           = 0.33\textwidth,
			height          = 3cm,
			xlabel 			= {Layers Randomized (\%)},
			ylabel 			= {},		
			label style 	= {font = \scriptsize},
            xtick = {0,10,...,100},
            xticklabels = {0,,20,,40,,60,,80,,100},
			ymax            = 1.0,
			ymin            = 0.0,
			legend style    = {at = {(-2.1,-1.1)}, anchor = north west, nodes={scale=0.85}, draw=none, {/tikz/every even column/.append style={column sep=0.5cm}}},
			legend columns  = 3,
			]
			\nextgroupplot[title={Cosine}, mark options = {scale=0.5}, ylabel = {Similarity Score}]
			\foreach \ycontrast in {original,contrastive} {
				\foreach \ybase in {guided_backprop, integrated_gradients, input_x_gradient}{
					\addplot+[] table[x=x_axis, 	y={\ybase_\ycontrast_cosine}, col sep=comma] {data/rando/wide_resnet502_inserted.csv};
			}}
        	\foreach \ycontrast in {original,contrastive} {
				\foreach \ybase in {gradcam, guided_grad_cam, lrp}{
					\addplot+[] table[x=x_axis, 		y={\ybase_\ycontrast_cosine}, col sep=comma] {data/rando/wide_resnet502_inserted.csv};
			}}
			\nextgroupplot[title={Spearman}, mark options = {scale=0.5}]
			\foreach \ycontrast in {original,contrastive} {
				\foreach \ybase in {guided_backprop, integrated_gradients, input_x_gradient}{
					\addplot+[] table[x=x_axis, 		y={\ybase_\ycontrast_spearman}, col sep=comma] {data/rando/wide_resnet502_inserted.csv};
			}}
        	\foreach \ycontrast in {original,contrastive} {
				\foreach \ybase in {gradcam, guided_grad_cam, lrp}{
					\addplot+[] table[x=x_axis, 	 	y={\ybase_\ycontrast_spearman}, col sep=comma] {data/rando/wide_resnet502_inserted.csv};
			}}
			\nextgroupplot[title={Pearson}, mark options = {scale=0.5}]
			\foreach \ycontrast in {original,contrastive} {
				\foreach \ybase in {guided_backprop, integrated_gradients, input_x_gradient}{
					\addplot+[] table[x=x_axis, 		y={\ybase_\ycontrast_pearson}, col sep=comma] {data/rando/wide_resnet502_inserted.csv};
			}}
			\foreach \ycontrast in {original,contrastive} {
				\foreach \ybase in {gradcam, guided_grad_cam, lrp}{
					\addplot+[] table[x=x_axis, 	 	y={\ybase_\ycontrast_pearson}, col sep=comma] {data/rando/wide_resnet502_inserted.csv};
			}}
			\legend{Guided Backprop, Integrated Gradients, Input$\times$Gradients,
                    Guided Backprop + \ourmethod, Integrated Gradients + \ourmethod, Input$\times$Gradients + \ourmethod,
                GradCAM, Guided GradCAM, LRP,
				  GradCAM + \ourmethod, Guided GradCAM + \ourmethod, LRP + \ourmethod}				 			
		\end{groupplot}
	\end{tikzpicture}%
	\caption{WRN50-2: \ourmethod improves all base methods under randomization [Lower is better]. For all methods and for varying level of randomization, we measure the similarity between the attention map for the unperturbed network and the randomized network. Dashed lines are base methods, solid lines when augmenting with \ourmethod, which improve the corresponding baseline method.}
	\label{fig:sanity-metrics-full-wrn}
\end{figure*}
                    
                    
\begin{figure*}[h!]
	\begin{tikzpicture}
    		\pgfplotscreateplotcyclelist{prcl-iccv-interleaved}{
			{dollarbill, mark=*, dotted}, {darkblue, mark=*, dotted},{mediumpurple, mark=*, dotted},
			{dollarbill, mark=*, solid},{darkblue, mark=*, solid},{mediumpurple, mark=*, solid},
			{mikadoyellow, mark=*, dotted}, {orange, mark=*, dotted},{pink, mark=*, dotted},
			{mikadoyellow, mark=*, solid},
			{orange, mark=*, solid},
            {pink, mark=*, solid},
		}
		\begin{groupplot}[
			pretty line,
			pretty ygrid,
			cycle list name = prcl-iccv-interleaved, 
			group style={group size=3 by 1, horizontal sep=5em},
			title style = {font=\footnotesize},
			width           = 0.33\textwidth,
			height          = 3cm,
			xlabel 			= {Layers Randomized (\%)},
			ylabel 			= {},		
			label style 	= {font = \scriptsize},
            xtick = {0,10,...,100},
            xticklabels = {0,,20,,40,,60,,80,,100},
			ymax            = 1.0,
			ymin            = 0.0,
			legend style    = {at = {(-2.1,-1.1)}, anchor = north west, nodes={scale=0.85}, draw=none, {/tikz/every even column/.append style={column sep=0.5cm}}},
			legend columns  = 3,
			]
			\nextgroupplot[title={Cosine}, mark options = {scale=0.5}, ylabel = {Similarity Score}]
			\foreach \ycontrast in {original,contrastive} {
				\foreach \ybase in {guided_backprop, integrated_gradients, input_x_gradient}{
					\addplot+[] table[x=x_axis, 	y={\ybase_\ycontrast_cosine}, col sep=comma] {data/rando/convnext_inserted.csv};
			}}
        	\foreach \ycontrast in {original,contrastive} {
				\foreach \ybase in {gradcam, guided_grad_cam}{
					\addplot+[] table[x=x_axis, 		y={\ybase_\ycontrast_cosine}, col sep=comma] {data/rando/convnext_inserted.csv};
			}}
			\nextgroupplot[title={Spearman}, mark options = {scale=0.5}]
			\foreach \ycontrast in {original,contrastive} {
				\foreach \ybase in {guided_backprop, integrated_gradients, input_x_gradient}{
					\addplot+[] table[x=x_axis, 		y={\ybase_\ycontrast_spearman}, col sep=comma] {data/rando/convnext_inserted.csv};
			}}
        	\foreach \ycontrast in {original,contrastive} {
				\foreach \ybase in {gradcam, guided_grad_cam}{
					\addplot+[] table[x=x_axis, 	 	y={\ybase_\ycontrast_spearman}, col sep=comma] {data/rando/convnext_inserted.csv};
			}}
			\nextgroupplot[title={Pearson}, mark options = {scale=0.5}]
			\foreach \ycontrast in {original,contrastive} {
				\foreach \ybase in {guided_backprop, integrated_gradients, input_x_gradient}{
					\addplot+[] table[x=x_axis, 		y={\ybase_\ycontrast_pearson}, col sep=comma] {data/rando/convnext_inserted.csv};
			}}
			\foreach \ycontrast in {original,contrastive} {
				\foreach \ybase in {gradcam, guided_grad_cam}{
					\addplot+[] table[x=x_axis, 	 	y={\ybase_\ycontrast_pearson}, col sep=comma] {data/rando/convnext_inserted.csv};
			}}
			\legend{Guided Backprop, Integrated Gradients, Input$\times$Gradients,
                    Guided Backprop + \ourmethod, Integrated Gradients + \ourmethod, Input$\times$Gradients + \ourmethod,
                GradCAM, Guided GradCAM, LRP,
				  GradCAM + \ourmethod, Guided GradCAM + \ourmethod, LRP + \ourmethod}				 			
		\end{groupplot}
	\end{tikzpicture}%
	\caption{ConvNext: \ourmethod improves all base methods under randomization [Lower is better]. For all methods and for varying level of randomization, we measure the similarity between the attention map for the unperturbed network and the randomized network. Dashed lines are base methods, solid lines when augmenting with \ourmethod, which improve the corresponding baseline method.}
	\label{fig:sanity-metrics-full-convnext}
\end{figure*}
\input{tex_figs/rando_vit}

\begin{figure*}
	\begin{tikzpicture}
		\path (0,0) -- (16,0);
		\def\igX{4.4}
		\def\igY{0}
		\node (label_ig) at (\igX, \igY) {\small Integrated Gradients};
		
		\def\gbpX{9.3}
		\def\gbpY{0}
		\node (label_gbp) at (\gbpX, \gbpY) {\small Guided Backprop};
		
		\def\ixgX{14.25}
		\def\ixgY{0}
		\node (label_ixg) at (\ixgX, \ixgY) {\small \ixg};
	\end{tikzpicture}
    \centering
    \includegraphics[width=\linewidth, trim=0 0 0 0, clip]{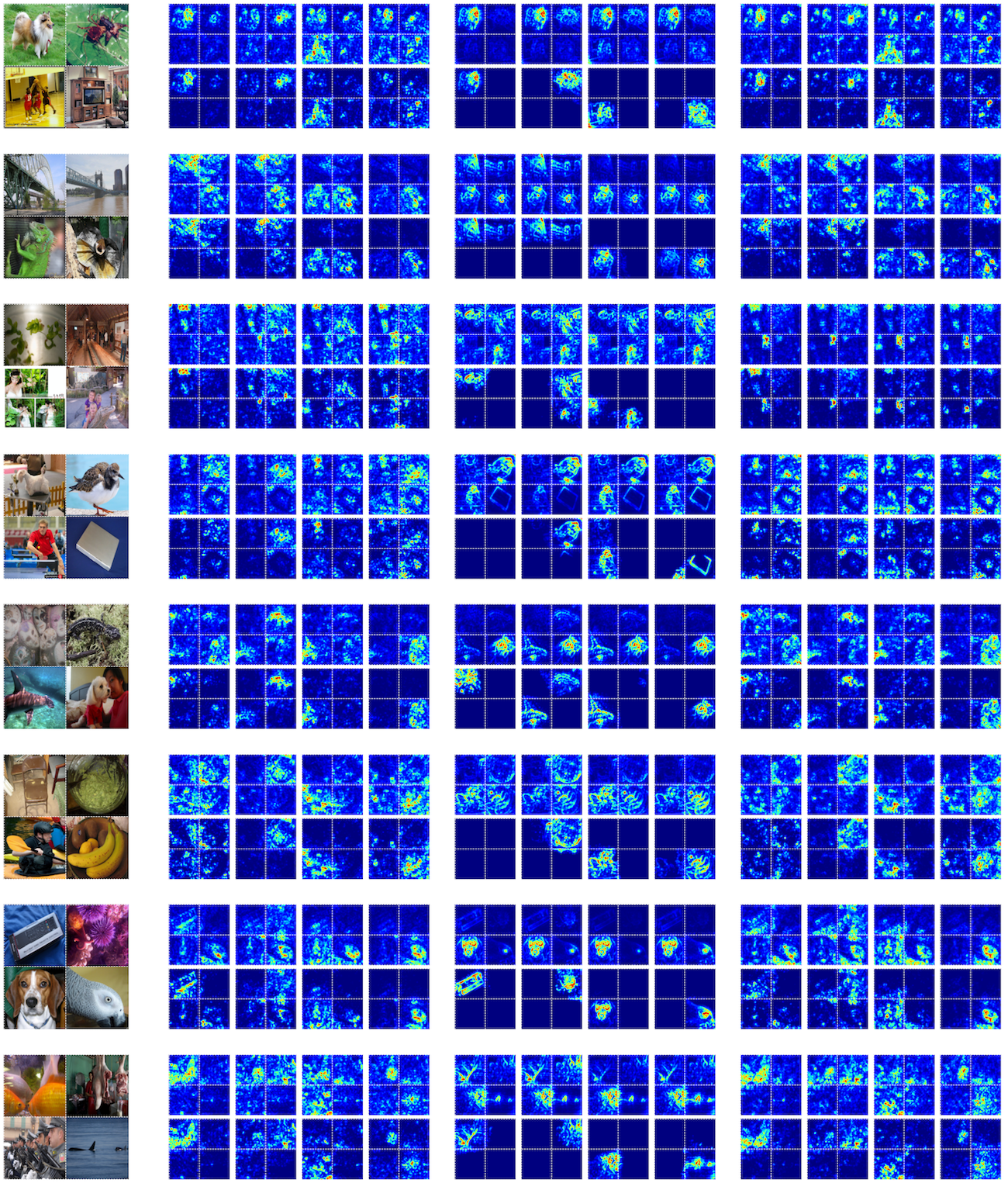} 
    \caption{\textbf{DenseNet121}: \textit{\ourmethod on the Grid Pointing Game.} We show examples from the grid pointing game for methods most affected by our framework (as columns: \igfull, \gbpfull, \ixgfull). Input Images are given on the left, for each we provide vanilla attribution methods (top row) and augmented with \ourmethod (bottom row).
    For each, we show the attribution for the four different classes in the grid as columns.}
        \label{app:densenet}
\end{figure*}

\begin{figure*}
	\begin{tikzpicture}
		\path (0,0) -- (16,0);
		\def\igX{4.2}
		\def\igY{0}
		\node (label_ig) at (\igX, \igY) {\small Integrated Gradients};
		
		\def\gbpX{9.3}
		\def\gbpY{0}
		\node (label_gbp) at (\gbpX, \gbpY) {\small Guided Backprop};
		
		\def\ixgX{14.25}
		\def\ixgY{0}
		\node (label_ixg) at (\ixgX, \ixgY) {\small \ixg};
	\end{tikzpicture}
    \centering
    \includegraphics[width=\linewidth, trim=0 0 0 0, clip]{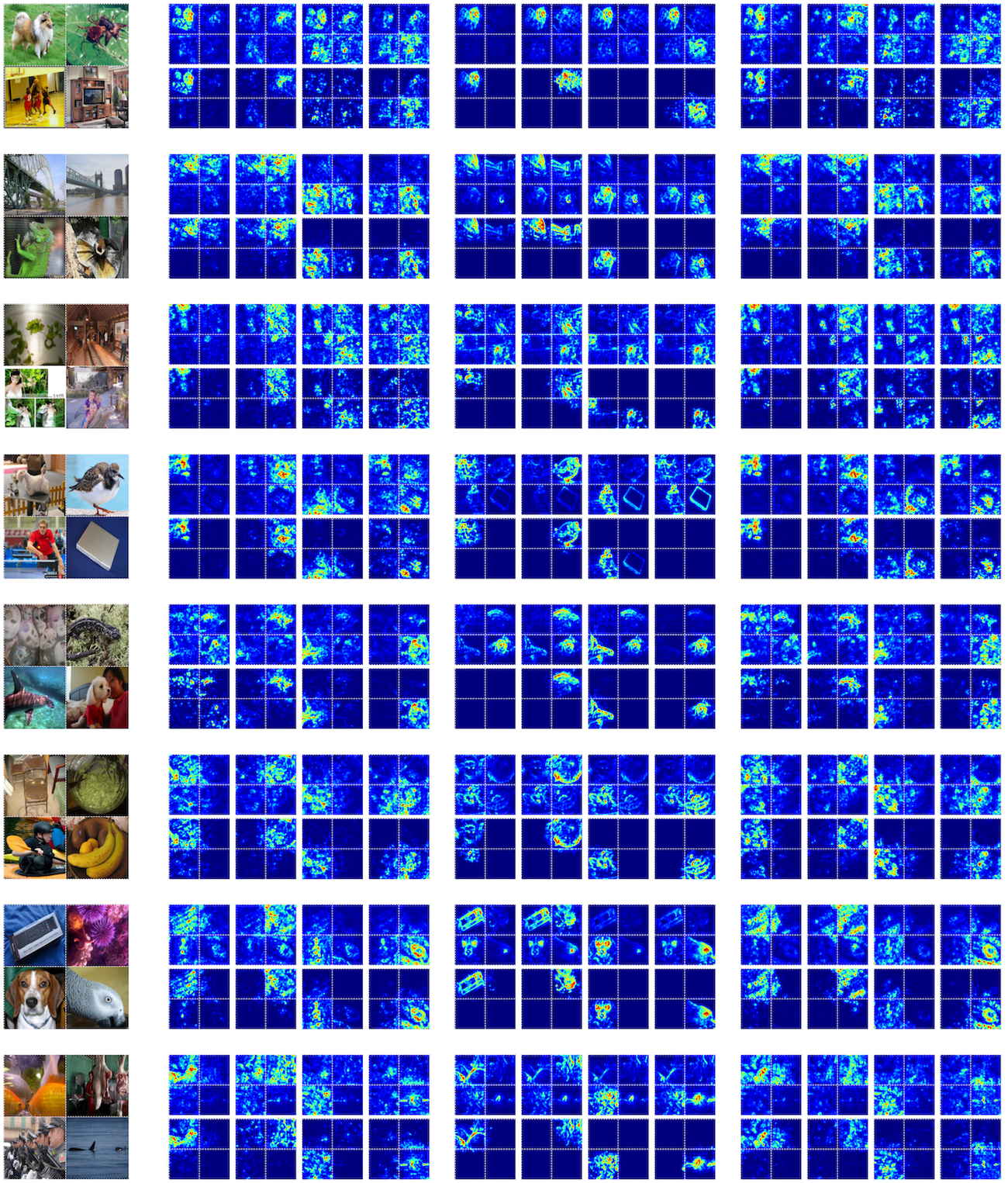} 
    \caption{\textbf{WideResNet50-2}: \textit{\ourmethod on the Grid Pointing Game.} We show examples from the grid pointing game for methods most affected by our framework (as columns: \igfull, \gbpfull, \ixgfull). Input Images are given on the left, for each we provide vanilla attribution methods (top row) and augmented with \ourmethod (bottom row).
    For each, we show the attribution for the four different classes in the grid as columns.}
        \label{app:wide}
\end{figure*}

\begin{figure*}
	\begin{tikzpicture}
		\path (0,0) -- (16,0);
		\def\igX{4.2}
		\def\igY{0}
		\node (label_ig) at (\igX, \igY) {\small Integrated Gradients};
		
		\def\gbpX{9.3}
		\def\gbpY{0}
		\node (label_gbp) at (\gbpX, \gbpY) {\small Guided Backprop};
		
		\def\ixgX{14.25}
		\def\ixgY{0}
		\node (label_ixg) at (\ixgX, \ixgY) {\small \ixg};
	\end{tikzpicture}
    \centering
    \includegraphics[width=\linewidth, trim=0 0 0 0, clip]{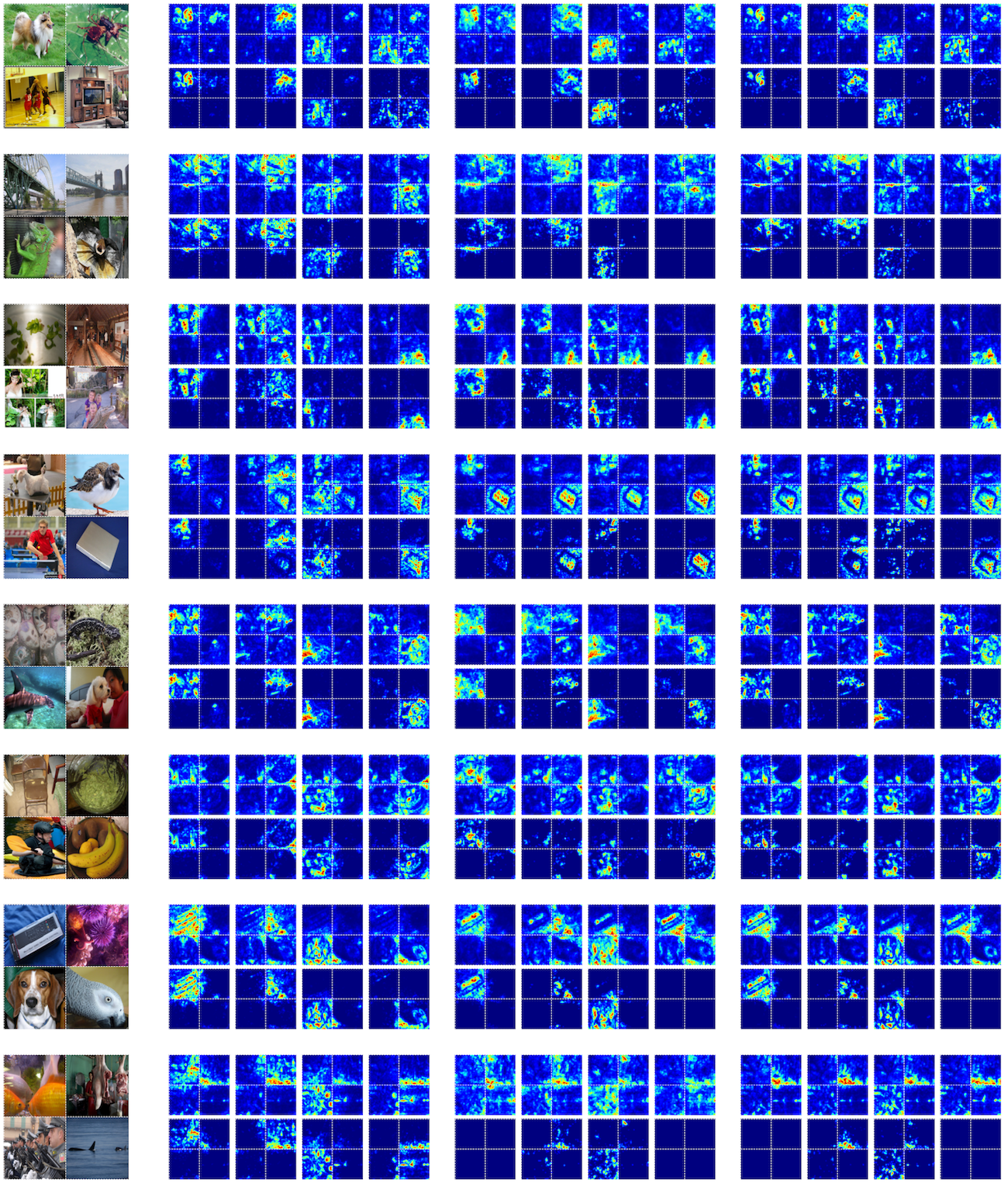} 
    \caption{\textbf{ConvNeXt}: \textit{\ourmethod on the Grid Pointing Game.} We show examples from the grid pointing game for methods most affected by our framework (as columns: \igfull, \gbpfull, \ixgfull). Input Images are given on the left, for each we provide vanilla attribution methods (top row) and augmented with \ourmethod (bottom row).
    For each, we show the attribution for the four different classes in the grid as columns.}
    \label{app:convnext}
\end{figure*}

\begin{figure*}
	\begin{tikzpicture}
		\path (0,0) -- (16,0);
		\def\igX{4.5}
		\def\igY{0}
		\node (label_ig) at (\igX, \igY) {\small Inflow};
		
		\def\gbpX{9.25}
		\def\gbpY{0}
		\node (label_gbp) at (\gbpX, \gbpY) {\small Gradient};
		
		\def\ixgX{14.15}
		\def\ixgY{0}
		\node (label_ixg) at (\ixgX, \ixgY) {\small Bi-attn};
	\end{tikzpicture}
    \centering
    \includegraphics[width=\linewidth, trim=0 0 0 0, clip]{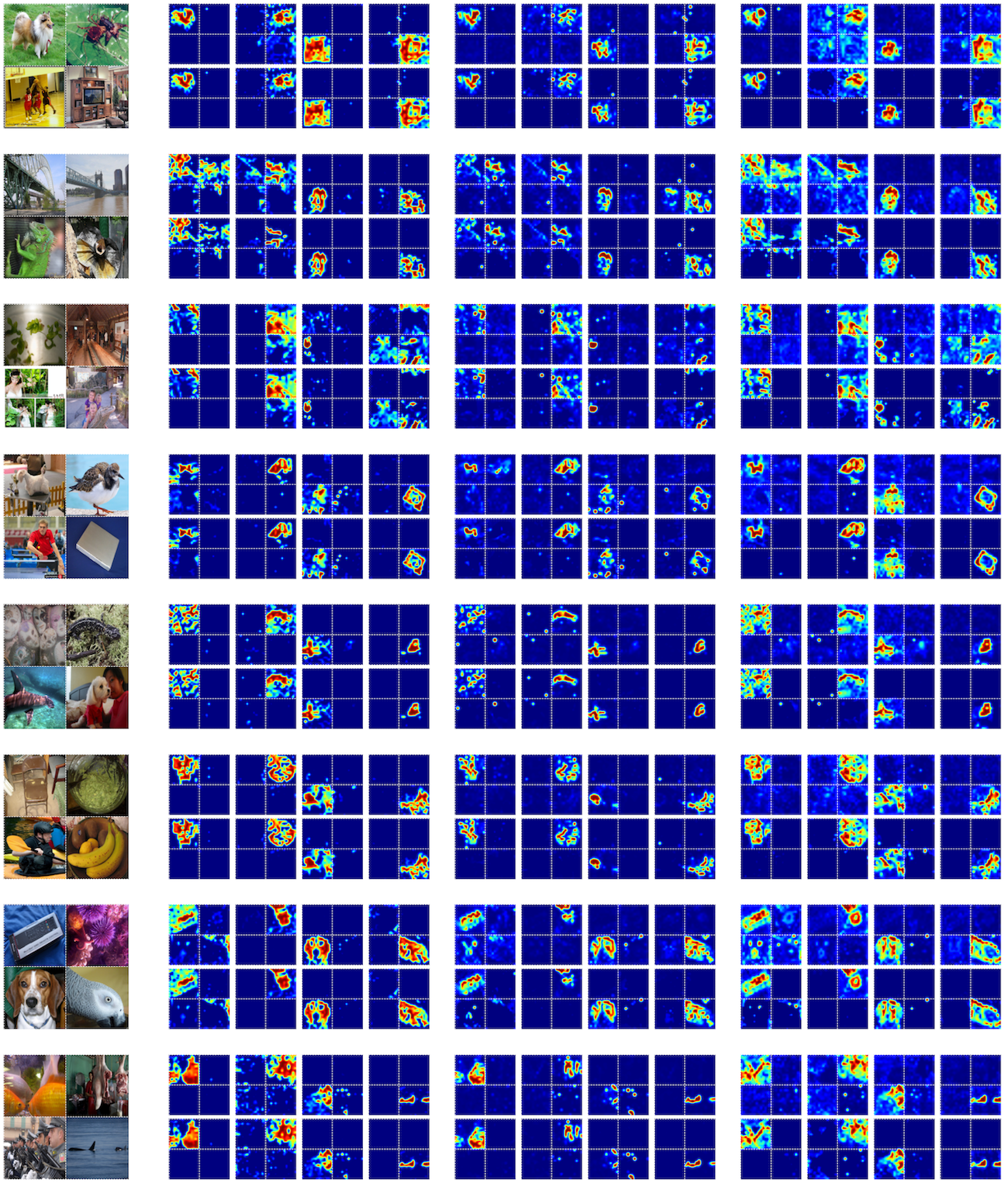} 
    \caption{\textbf{ViT-base-8}: \textit{\ourmethod on the Grid Pointing Game.} We show examples from the grid pointing game for methods most affected by our framework (as columns: \igfull, \gbpfull, \ixgfull). Input Images are given on the left, for each we provide vanilla attribution methods (top row) and augmented with \ourmethod (bottom row).
    For each, we show the attribution for the four different classes in the grid as columns.}
        \label{app:vit}
\end{figure*}

\begin{figure*}
	\begin{tikzpicture}
		\path (0,0) -- (16,0);
		\def\igX{4.5}
		\def\igY{0}
		\node (label_ig) at (\igX, \igY) {\small Inflow};
		
		\def\gbpX{9.25}
		\def\gbpY{0}
		\node (label_gbp) at (\gbpX, \gbpY) {\small Gradient};
		
		\def\ixgX{14.15}
		\def\ixgY{0}
		\node (label_ixg) at (\ixgX, \ixgY) {\small Bi-attn};
	\end{tikzpicture}
    \centering
    \includegraphics[width=\linewidth, trim=0 0 0 0, clip]{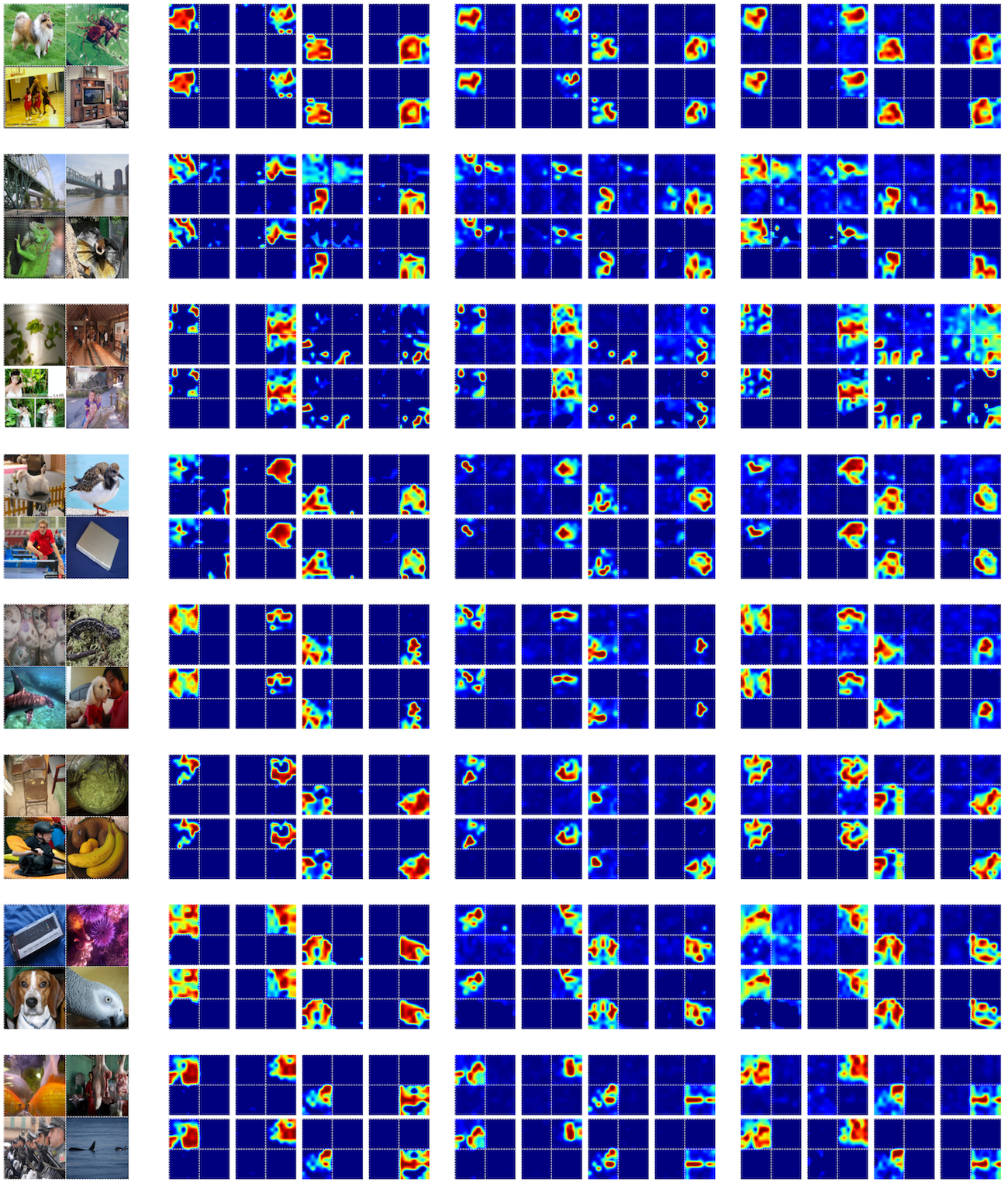} 
    \caption{\textbf{ViT-base-16}: \textit{\ourmethod on the Grid Pointing Game.} We show examples from the grid pointing game for methods most affected by our framework (as columns: \igfull, \gbpfull, \ixgfull). Input Images are given on the left, for each we provide vanilla attribution methods (top row) and augmented with \ourmethod (bottom row).
    For each, we show the attribution for the four different classes in the grid as columns.}
        \label{app:vit}
\end{figure*}

\begin{figure*}
	\begin{tikzpicture}
		\path (0,0) -- (16,0);
		\def\igX{4.5}
		\def\igY{0}
		\node (label_ig) at (\igX, \igY) {\small Inflow};
		
		\def\gbpX{9.25}
		\def\gbpY{0}
		\node (label_gbp) at (\gbpX, \gbpY) {\small Gradient};
		
		\def\ixgX{14.15}
		\def\ixgY{0}
		\node (label_ixg) at (\ixgX, \ixgY) {\small Bi-attn};
	\end{tikzpicture}
    \centering
    \includegraphics[width=\linewidth, trim=0 0 0 0, clip]{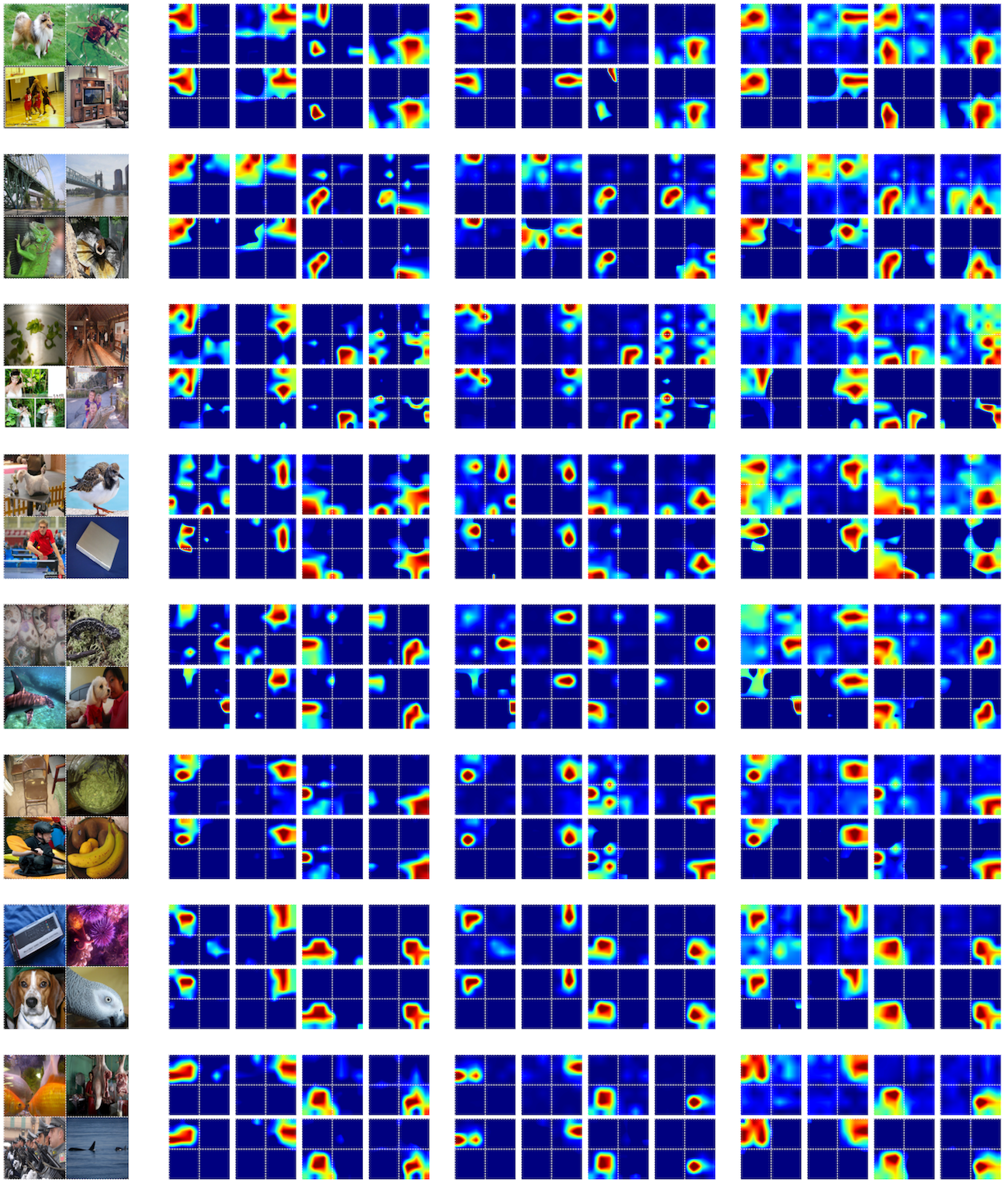} 
    \caption{\textbf{ViT-base-32}: \textit{\ourmethod on the Grid Pointing Game.} We show examples from the grid pointing game for methods most affected by our framework (as columns: \igfull, \gbpfull, \ixgfull). Input Images are given on the left, for each we provide vanilla attribution methods (top row) and augmented with \ourmethod (bottom row).
    For each, we show the attribution for the four different classes in the grid as columns.}
        \label{app:vit}
\end{figure*}

\end{document}